\documentclass[a4paper]{article}

\usepackage{amsmath, amssymb, amsthm}
\usepackage{mathtools}
\usepackage[english]{babel}
\usepackage[utf8]{inputenc}
\usepackage{graphicx}
\usepackage{subcaption}
\usepackage{cleveref}
\usepackage{float}
\usepackage[colorinlistoftodos]{todonotes}
\usepackage[numbers]{natbib}
\usepackage{geometry}
\usepackage{enumitem}
\usepackage[toc, page]{appendix}
\usepackage{booktabs}
\usepackage{multirow}
\usepackage{makecell}

\title{Alternative Strategies for Active Learning with Bayesian Bradley Terry Model}
\author{
    Saman Jayasinghe\textsuperscript{a}, Sean Walton\textsuperscript{a}, Alma Rahat\textsuperscript{b}, Andy Gray\textsuperscript{c}, Chris Wheadon\textsuperscript{d}\\[1em], 
    \small\textsuperscript{a}Swansea University, United Kingdom\\
    \small\textsuperscript{b}Loughborough University, United Kingdom\\
    \small\textsuperscript{c}Bath Spa University, United Kingdom\\
    \small\textsuperscript{d} No More Marking Ltd, United Kingdom
}

\begin{document}
\maketitle
\begin{abstract}
 Comparative judgement uses pairwise comparisons to rank items, but recovering accurate rankings from a limited number of judgements depends on both the inference method and the pairs selected. We developed a Bayesian Bradley-Terry model (BBTM) that uses Polya-Gamma data augmentation for efficient Gibbs sampling and quantifies uncertainty in latent abilities and ranks. We propose three active learning strategies based on the knowledge gradient (KG) principle. We define our acquisition functions using Kendall's tau distance, posterior rank entropy, and posterior separation. Approximate one-step updates to posterior moments reduce the computational cost of evaluating candidate pairs. We also introduce Posterior Separation Reliability (PSR), which measures the spread of posterior mean abilities relative to posterior uncertainty. It provides an internal indicator of ranking reliability without requiring a reference ranking. In synthetic experiments covering different item counts, comparison budgets, and noise levels, BBTM generally recovers ranking more accurately than preference based Bayesian comparative judgement (PBCJ). At low to moderate noise levels, pairwise entropy pair selection strategy performs best in ranking recovery. However, at the highest noise levels, the KG strategy based on posterior separation produces better point rankings. PSR increases alongside improvements in ranking accuracy as comparisons accumulate. Robustness experiments identify the proportion of informative outcomes needed to meet specified accuracy targets within a fixed comparison budget. In moderate noise setting, replacing 30\% of outcomes with random choices has little effect on ranking performance. These findings demonstrate the value of Bayesian inference and active learning pair selection recovering ranks from limited, noisy paired comparisons.               
\end{abstract}

\section{Introduction}

Measuring the reliability of responses for complex, open-ended assessments, such as extended writing essays, is a key challenge in educational assessments. Traditional marking procedures like rubrics have been inconsistent for these tasks. Standards can vary between assessors, and even a single assessor may show some inconsistency due to the higher cognitive workload of traditional marking. Comparative Judgement (CJ), therefore, emerges as an alternative to address those drawbacks in traditional marking~\cite{thurstone1927law}. In the CJ setting, assessors are given a pair of assessments to identify the best piece of work in the given pair rather than assign an absolute mark. CJ is recognised as a reliable, scalable approach in educational assessment since humans are better at relative judgment than at absolute judgment~\cite{pollitt2012method, jones2024comparative}. A series of pairwise comparisons is utilised to obtain the final rank of each item, obtained from the underlying quality score which are estimated from the comparison outcomes. The Bradley-Terry (BT) model~\cite{bradley1952rank} is the classical foundation for this inference, which links the outcome probabilities to their latent abilities.\\  

The BT model uses the maximum likelihood estimation type point estimation technique to estimate the latent abilities of items~\cite{bradley1952rank}. Efficient algorithms have been developed~\cite{shah2018simple, chen2015spectral} to derive a single ranking for point estimates; however, they do not provide an estimate of uncertainty around these estimates. It is crucial to know how certain these rankings are in the high-stakes context of educational assessments. Bayesian inference provides these uncertainty measures around estimates as it gives posterior distributions over items' latent abilities instead of point estimates. Since the BT likelihood has a form of logistic, it is hard to sample the posterior directly due to the non-conjugacy with the Gaussian prior. Polya-Gamma data augmentation~\cite{polson2013bayesian} converts this logistic likelihood into conditionally Gaussian form. This augmentation establishes conjugation and, more importantly it enables efficient Gibbs sampling. Recently, Bayesian BT has been used to compare multiple machine learning algorithms~\cite{wainer2023bayesian}, but to the best of our knowledge, this Bayesian BT with Polya-Gamma data augmentation has not been applied to CJ in an educational setting or to active learning for pair selection.\\

There are $\binom{N}{2}$ possible pairs for $N$ items. As $N$ increases, the possible number of pairs increases quadratically. It is not practical to compare all these pairs, because human judgments are expensive. Consequently, it is essential to identify the most informative pairs that provide the highest information about the ranking order, using AL methods. Adaptive comparative judgement (ACJ)~\cite{pollitt2012method} was the first such implementation in an educational CJ setting, and more recently, entropy-driven AL pair selection has been developed~\cite{gray2024bayesian, gray2025bayesian}. The Knowledge Gradient (KG) methods have been established as a highly effective acquisition function in areas such as optimal learning, ranking and selection (RS), sequential experimental design~\cite{frazier2008knowledge, frazier2009knowledge, mes2011hierarchical, groves2018sequential}. These KG methods have been adapted to the pairwise setting, but these implementations do not recover the ranking of all items; instead, they recover top-$k$ selection or best arm identification~\cite{groves2019top, li2025knowledge}. Further, these KG methods have not been tailored to the CJ framework and are difficult to scale for a higher number of items, since exact KG computation depends on Monte Carlo (MC) sampling.\\

The Scale Separation Reliability (SSR) measure is used to assess the reliability of CJ results~\cite{kinnear2025comparative, verhavert2018scale}. SSR is a point estimation based measure and it does not consider posterior uncertainty. Concerns have been emerging around this SSR measure in the CJ community~\cite{verhavert2019meta}. It is important to have a reliability measure that can be computed directly from the posterior distribution, without considering the ground truth. Under adaptive pair selection, estimates can be biased, because pair selection depends on outcomes. So, observations are not independent. \citet{vojnovic2020convergence} studied the convergence properties of BT model under a non-adaptive (i.i.d.) setting. Bernstein-von Mises (BvM) theorem and Bayesian posterior consistency exist for non-i.i.d. observations~\cite{ghosal2007convergence, van2000asymptotic} however, they have not been directly applied to the adaptive pairwise comparison setting. A proof that guarantees the convergence of Bayesian BT model under the adaptive pair selection is therefore still lacking. Since human judgments are expensive, recovering an optimal ranking entirely from human judgments would incur a high cost. Trade-off between human judgments and cheaper random coin toss outcomes has not been examined.\\

No existing Bayesian approach to CJ combines all of the following within a single framework: (i) theoretical guarantees of convergence under active learning pair selection; (ii) a computationally efficient KG pair selection that enables the full ranking recovery; (iii) a reliability measure that can be computed directly from the posterior, without access to the ground truth; (iv) a systematic study of the robustness of ranking recovery under imperfect/noisy judgments. We address these gaps through the introduction the Bayesian Bradley Terry Model (BBTM) - a model which integrates both theoretical foundations and practical pair selection.\\

Our main contributions can be listed as follows:
\begin{itemize}
    \item A hierarchical Bayesian version of the BT model with Polya-Gamma data augmentation which enables efficient Gibbs sampling. This model provides full posterior inference with uncertainty quantification on item strengths with sequential updating. The hierarchical structure yields better estimates for sparse or limited comparisons, due to shrinkage regularisation through borrowing strength~\citep[Chapter~12]{gelman2007data}.
    \item Theoretical guarantees for the convergence of BBTM under adaptive pair selection via posterior consistency and BvM theorem. The convergence conditions are met using $\varepsilon$-greedy exploration.
    \item KG approaches to the AL pair selections. We develop three new KG acquisition functions specially tailored for the CJ framework: expected rank disruption, expected rank entropy reduction, and expected improvement in reliability.    
    \item A computationally efficient KG based pair selection approach for the Bayesian BT CJ. We approximate one-step look-ahead values using Gaussian moment matching instead of nested Markov Chain Monte Carlo (MCMC) sampling. This enables scalable KG based pair selection under multiple acquisition functions.
    \item A new reliability measure, Posterior Separation Reliability (PSR) which is computed from the posterior distribution. The PSR measure is also compatible with the traditional SSR.
    \item A near optimal mixture design which combines informative human judgments and random outcomes strategically. This reduces the need for costly human judgments while producing accurate ranking.
\end{itemize}

The remainder of this paper is organised as follows. Section~\ref{BG_LW} describes the background and reviews the existing literatures on CJ and Bayesian CJ frameworks. Section~\ref{Method} presents the BBTM formulation, including the novel KG pair selection methods, fast KG via Gaussian moment updates, and the PSR measure. The Convergence proof of the BBTM is presented in Section~\ref{Conv_proof}. Section~\ref{Exp_Dis} discusses the experimental setup, the identification of the winning method, rank distribution recovery, PSR and the robustness of the model. Section~\ref{Conclusion} summarises the final conclusions.        

\newpage

\section{Background and Literature Review}\label{BG_LW}
\subsection{Paired comparisons and Comparative Judgment}
Paired comparison is a well-known and established approach for measuring latent qualities when absolute scoring is difficult. Thurstone's Law of Comparative Judgment~\cite{thurstone1927law} is the theoretical foundation of such models. Later, Luce~\cite{luce1959individual} developed probabilistic choice models and provided a general framework for modeling comparative decisions. These early contributions show that pairwise judgments can be systematically converted into a latent scale under suitable probabilistic assumptions. \\

\noindent Over time, paired comparison methods have been used in a wide range of domains. In sports analytics, models based on paired outcomes are used to infer player or team strengths from match results~\cite{elo1978rating}. In consumer research and sensory analysis, pairwise preferences are analysed to construct product rankings and understand decision behaviour~\cite{cattelan2012models}. In psychology and behaviour sciences, dominance structure and social hierarchies are often estimated from pairwise interactions (David, 1988). Wainer~\cite{wainer2023bayesian} has used paired comparisons to compare multiple Machine Learning algorithms on multiple data sets. Further, paired comparison has become central in Artificial Intelligence and Large Language model evaluation, where human judges compare model outputs to construct performance rankings~\cite{christiano2017deep}.\\

\noindent In the education sector, Comparative Judgment (CJ) has gained increasing attention as an alternative to traditional rubric-based marking. Instead of assigning absolute marks, teachers compare two pieces of student work and select the better piece. Pollitt~\cite{pollitt2012method} argues that CJ simplifies the assessment task and improves decision consistency, with Adaptive CJ (ACJ) further enhancing efficiency through dynamic pairing. But, Bramley~\cite{bramley2015investigating} warns that while ACJ is designed to select informative comparisons dynamically, the process produces biased statistics that artificially inflate the reliability scores. However, regardless of the domain, a common challenge remains: how to transform incomplete and noisy pairwise comparisons into a stable and reliable ranking.

\subsection{The Bradley Terry Model for Ranking Paired Comparisons}
Among the available probabilistic models for paired comparison data, the Bradley Terry Model (BTM)~\cite{bradley1952rank} is one of the most widely used. In this model, each item is associated with a latent strength parameter. The probability that item $i$ is preferred over item $j$ is modelled as a logistic function of the difference between their strengths. This formulation is simple, interpretable, and computationally convenient.\\

\noindent Estimation of the BTM is commonly performed using Maximum Likelihood methods, and efficient algorithms such as MM algorithm have been developed~\cite{hunter2004mm}. Several extensions have also been proposed, including models that handle ties~\cite{rao1967ties, davidson1970extending}, home advantage~\citep[Chapter 10]{agresti2013categorical}, order effect~\cite{davidson1977extending}, or incorporate object and subject specific covariates~\cite{cattelan2012models}.\\

\noindent Despite its popularity, the classical BTM has some limitations. Its parameter estimation relies heavily on the structure of the comparison network. If the comparison graph is sparse or poorly connected, latent strength estimates may become unstable. Moreover, the standard frequentist approach provides point estimations and asymptotic standard errors, but it does not provide full uncertainty distributions over strengths or ranks. Finally, in sequential and adaptive settings such as CJ, the classical BTM framework does not directly provide guidance on which pair should be compared next. \\

\noindent These limitations are particularly important in CJ, where judgments are collected sequentially, and decisions must be justified in terms of reliability and fairness. Hence, Bayesian extensions to the BTM are highly desirable.

\subsection{Bayesian Bradley Terry Model}
Prior distributions are placed on the latent strengths, and inference is performed through the posterior distribution given the observed pairwise outcomes in the Bayesian version of BTM. Conjugate priors, such as Gamma priors for the positive skill parameters, have been used, enabling a tractable posterior and efficient computation~\cite{caron2012efficient}. They use a data augmentation strategy that enables efficient Gibbs sampling for full Bayesian inference and a simple Expectation-Maximisation (EM) algorithm for Maximum a Posteriori (MAP) estimation. Their work shows that Bayesian inference in generalized BTM can be implemented in a computationally scalable way, even for large comparison datasets.\\

Wainer~\cite{wainer2023bayesian} proposes a Bayesian Bradley-Terry (BBT) framework for comparing multiple machine learning (ML) algorithms across multiple datasets. He demonstrates that the Bayesian approach provides posterior probabilities that one ML algorithm outperforms another and allows the definition of Regions of Practical Equivalence (ROPE). This enables uncertainty quantification in aggregated rankings while avoiding traditional hypothesis testing. In applied statistical modelling, ~\cite{seymour2022bayesian} introduce a Bayesian Spatial BTM for urban deprivation modelling by placing structured priors over latent strengths. These priors enable borrowing information across neighbouring regions and reduce the number of comparisons for stable inference. Further, ~\cite{krese2021bayesian} extend the BTM to allow time-varying latent strengths, using Gaussian process priors and MCMC sampling. These extensions allow the flexibility of the Bayesian framework to accommodate various structures.\\

Prior specification in the Bayesian BTM is crucial, as it directly affects invariance and interpretability properties. According to ~\cite{whelan2017prior}, a prior distribution for the BTM should satisfy these four desiderata: (i) Invariance under interchange of items, (ii). Invariance under interchange of winning and losing, (iii) Normalisability, (iv) Invariance under elimination of items. The analysis reveals that some commonly used priors, such as Jeffreys prior and certain conjugate priors, fail to satisfy one or more these desiderata. Furthermore, the paper identifies two families of prior distributions that satisfy all four: independent Gaussian distributions and the Type III Generalised Logistic distribution.\\        

\noindent Bayesian logistic models have been computationally feasible due to recent developments, particularly Polya-Gamma data augmentation~\cite{polson2013bayesian}. This enables efficient Gibbs sampling by introducing auxiliary variables that transform the logistic likelihood into a conditionally Gaussian form. It provides computationally stable and scalable posterior inference for the paired comparison models, particularly in hierarchical settings. Furthermore, Gibbs samplers show better mixing and require less tuning than Metropolis-Hastings methods.\\

\noindent The advantages of BBTM are threefold. (i) The uncertainty can be directly quantified through posterior distributions rather than approximated through asymptotic theory. This allows us to compute credible intervals for latent strengths and to derive probabilities of rank positions. (ii) Prior distributions enable regularisation, which is highly useful when the number of comparisons is low or when the comparison graph is sparse. (iii) Sequential data collection has been supported by the Bayesian updating process, and it is essential in active learning in CJ environment.\\

\noindent In our work, we implement a Bayesian BTM (BBTM) using Polya-Gamma data augmentation, which enables efficient Gibbs sampling. The model aggregates binomial outcomes for each item pair and updates the posterior distribution iteratively as new comparisons are observed. This model design is suitable for CJ in education, where decisions are sequential and uncertainty quantification is necessary.

\subsection{Bayesian Comparative Judgment in Education}
Rasch based scaling~\cite{rasch1993probabilistic} or frequentist BTM estimations have traditionally been used in CJ research in education~\cite{kinnear2025comparative}. While these approaches have shown practical usefulness~\cite{verhavert2019meta}\cite{pollitt2012method}, there has been growing interest in the Bayesian CJ framework in education~\cite{gray2024bayesian}. Uncertainty in latent abilities can be explicitly modelled in Bayesian modelling, and this uncertainty is carried over to the rank estimates.\\

The reliability of estimates is a central concern in educational assessment~\cite{kinnear2025comparative}\cite{verhavert2018scale}\cite{verhavert2019meta}. Decision makers require evidence that rankings are stable and defensible. Bayesian CJ offers posterior distributions over item strengths and ranks that provide additional interpretability for these estimates. This probabilistic perspective aligns well with the uncertain nature of human judgments.\\

Bayesian approaches to CJ are still a relatively new area, particularly when it comes to active learning and decision optimisation. Researchers have recently made considerable progress in combining Bayesian modelling with structured active learning strategies. ~\citet{gray2024bayesian} proposed a novel Bayesian approach to CJ (BCJ) that directly models pairwise preferences using Beta distributions rather than focusing on latent scores. They illustrate how to compute the entire predictive rank distribution for any item being assessed. Furthermore, they have developed a new active learning strategy to select next pair to evaluate based on entropy. By calculating predictive entropy of the preference densities, the algorithm actively selects the most informative pair to show the judges. They have shown that this will significantly improve the accuracy and efficiency over random or round robin selection methods.\\

~\citet{gray2025bayesian}, later extended the BCJ framework to handle multi-criteria assessments, bridging the gap between traditional CJ evaluations and detailed rubric-based grading. This approach models preferences across multiple independent learning outcomes (LOs) instead of relying on a single judgment. They used a weighted ensemble approach to aggregate these multi-dimensional comparisons into a reliable overall rank. Moreover, they have used multi-dimensional differential entropy to calculate the combined uncertainty across all LOs to intelligently select the next pair to evaluate under entropy driven pair selection method ~\citet{gray-BERA}.\\

Although BCJ introduces posterior rank distributions and entropy-based active learning pair selection, existing approaches primarily focus on local uncertainty reduction rather than explicitly optimising global ranking accuracy or reliability objectives.\\

\subsection{Scale Separation Reliability (SSR)}
In CJ, Scale Separation Reliability (SSR) is a measure used to evaluate the reliability of assessments~\cite{verhavert2018scale}\cite{jones2024comparative}. Assessors compare pairs of works in a CJ task, and their decisions are analysed using the BTM to create a ranked scale of quality. The CJ research community adopted the SSR metric directly from the Rasch literature~\cite{verhavert2018scale} because the CJ model shares strong mathematical similarities with Item Response Theory (IRT) and Rasch methods ~\cite{bramley2015investigating, pollitt2012method}. SSR reflects the same core idea found in Classical Test Theory (CTT); it attempts to measure the proportion of variance in observed scores that is truly attributable to the true scores of the representations being assessed~\cite{brennan2010generalizability}\\

SSR compares true score variance against the error present in the estimates. According to~\citet{bramley2015investigating}, the formula used to calculate SSR is:
\[
SSR = \frac{G^2}{(1+G^2)}
\]
$G$ is calculated as $G = \frac{\sigma_\beta}{RMMSE}$, where $\sigma_\beta$ represents the standard deviation of the true scores, and $RMSE$ is the root mean square error. This SSR formula can be algebraically simplified to directly show the ratio of true score variance $(\sigma^2_\beta)$ to observed score variance $(\sigma^2_\nu)$, and that will make the connection to CTT~\cite{verhavert2018scale}
\[
SSR = \frac{\sigma^2_\beta}{\sigma^2_\nu}
\]
The variance of the true score can be estimated from the variance of observed scores as follows:
\[
\sigma^2_\beta = \sigma^2_\nu - MSE
\]
Where MSE is the Mean Squared Error~\cite{andrich1982index}\\

SSR has been used as a measure of assessment consistency, even though it was a borrowed concept. However, ~\cite{verhavert2018scale} empirically validated it through a meta-analysis of 26 CJ assessments. They performed the study by splitting assessor groups into halves and found that SSR is a highly accurate estimate for split-half reliability. Additionally, they demonstrated that SSR can serve as a reliable indicator of inter-rater reliability. ~\cite{kinnear2025comparative} further validated these claims by conducting a meta-analysis of 101 CJ datasets, confirming that SSR acts as a reliable proxy for inter-rater agreement. Furthermore, their findings suggested that roughly 10 comparisons for every representation is generally sufficient to achieve reliable measurement scales.\\

According to ~\cite{verhavert2018scale}, caution should be exercised when copying reliability metrics from IRT to CJ, as there are different parameterisations in the measurement models. Furthermore, their results were mixed and inconclusive when they tested whether SSR could represent a strict correlation with the truth. They highlighted that this is likely due to conceptual differences in how IRT and CTT treat error variance. ~\citet{gray2024bayesian} pointed out that the relative uncertainty estimated by SSR highly depends on the underlying mathematical model, and this uncertainty is rarely presented to users intuitively. Additionally, they noted that due to the unavailability of true score variance before simulation, the calculation of SSR is highly problematic in practical scenarios. ~\citet{gray2024bayesian} also explained that SSR does not provide a real-time, natural performance metric; it fails to offer a dynamic stopping criterion. Hence, it is required to have a predetermined, fixed budget at the start of the simulation. While ~\citet{kinnear2025comparative} suggested that 10 judgments per script is roughly enough to get a reliable score in practice, this remains a static target. Since SSR is not a real-time monitoring measure, it cannot dynamically halt the process early even if true consensus is reached quickly.\\

These limitations suggest that reliability assessments in CJ would benefit from approaches that are more directly aligned with posterior rank uncertainty and decision-theoretic ranking objectives.

\subsection{Knowledge Gradient (KG) Methods}
In Ranking and Selection, efficiently allocating a limited sampling budget to identify the best alternatives is a central challenge. ~\citet{gupta1996bayesian} developed Bayesian look-ahead procedures to optimize these sampling allocations by evaluating the expected posterior risks. Recent researches show how samples were allocated to maximise the Expected Value of Information (EVI). For example, ~\citet{chick2007new} developed a new myopic sequential sampling procedure that avoided the asymptotic and probabilistic approximations used in earlier Bayesian probability models.\\

~\citet{frazier2008knowledge} formally introduced the KG methods based on the single look-ahead policy introduced by ~\citet{gupta1996bayesian}. They showed that the KG method myopically maximises the expected gain in information value one step at a time, whilst keeping a balance between computational methods and the need for sequential efficiency in information collection. \\

The KG method is mainly used in sequential information collection and offline learning to maximise the expected information gain~\cite{frazier2007knowledge}. It efficiently balances the trade-off between exploration and exploitation. It prefers to find alternatives in high variance regions (exploration) and alternatives close to the current maximum (exploitation), as these solutions are most likely to alter the final decision~\cite{frazier2007knowledge}. The KG policy is theoretically robust. ~\citet{frazier2008asymptotic} demonstrated that sequential sampling policies are asymptotically optimal.\\

The KG framework has been extended to handle more complex and realistic data structures. ~\citet{frazier2009knowledge} extended the KG framework to correlated normal beliefs, allowing the algorithm to exploit dependencies between alternatives and efficiently evaluate a large set of designs. Hierarchical Knowledge Gradient was introduced to accommodate multi-dimensional attributes~\cite{mes2011hierarchical}. Kullback-Leibler KG (KL-KG) was developed to minimise the divergence between the chosen and target offspring distributions rather than simply maximising the probability of correct selection~\cite{groves2018sequential}. It has been identified limitations of standard KG in some research settings. For instance, in problems involving Bernoulli outcomes and probability learning, the naive KG framework may be inefficient or require refinement~\cite{kaminski2015refined}.\\

The main idea of the KG methods is to evaluate how much the maximum of our current estimate would increase if a specific alternative/point were selected for evaluation. Since the outcome of the query is unknown at the time it is chosen, KG computes an expectation over all possible hypothetical outcomes, weighted by their predictive probabilities. The KG value for a candidate point $x$ is defined as 
\[
\mathbb{E}_{p(y \mid \mathcal{D})}
\left[
\max_{x' \in \mathcal{X}} 
\mathbb{E}\big[ f(x') \mid \mathcal{D}, (x,y) \big]
\right]
-
\max_{x' \in \mathcal{X}} 
\mathbb{E}\big[ f(x') \mid \mathcal{D} \big]
\]

Here, $y$ denotes the hypothetical observation at $x$, and $p(y\mid \mathcal{D})$ is the predictive distribution induced by the Gaussian Process (GP) model $\mathcal{f}$ given the current dataset $\mathcal{D}$.\\

Classical KG formulations assume that each measurement corresponds to observing a single alternative. However, in pairwise comparison experiments, two alternatives are involved in each sample, and the observation consists of a comparative outcome. Since each observation simultaneously informs beliefs about both items, this structure introduces additional complexity into the existing KG framework.~\citet{wu2026knowledge} developed an exact, analytical KG acquisition function for preferential Bayesian optimisation, which is a major advancement for traditional KG. They model preferences using a latent Gaussian Process (GP) utility function together with a probit likelihood. Since pairwise comparisons make the posterior distribution non-Gaussian, applying KG to preference learning is more challenging. However, they show that the one-step look-ahead posterior follows a skew normal distribution, which allows them to compute the look-ahead posterior mean in closed form.\\

The KG method has been extended to select the top-$k$ alternatives based on information collected from noisy, sequentially chosen pairwise comparisons~\cite{groves2019top}. They work with Borda scores that are built from pairwise comparison outcomes, instead of observing the true scores of each item. They adapt both KG and Optimal Computing Budget Allocation (OCBA) to the pairwise setting and compare their performance. They show that KG performs well when the goal is to identify the top performing set under a limited sampling budget. Furthermore, they prove that their approach is asymptotically optimal under certain conditions, and their method outperforms existing racing algorithms.\\

These developments in the KG framework are highly relevant to CJ. In CJ, each data point is a pairwise comparison between two student responses or pieces of work, and the aim is to produce a reliable ranking using a limited number of comparisons. This setup is very similar to a fixed-budget Bayesian information collection problem studied using KG~\cite{frazier2008knowledge}. Each comparison can be seen as a measurement decision, while the uncertainty about an item's true strength is captured by a posterior distribution over its latent strength.\\    

The KG framework has a strong theoretical foundation in value-of-information optimisation. However, it has not yet been fully integrated with hierarchical Bayesian BTM in the CJ context, particularly with respect to ranking reliability.\\   

Several important developments in the modelling and analysis of paired comparison data have been discussed above. The BTM offers a robust probabilistic framework for ranking based on pairwise outcomes, while Bayesian extensions introduce posterior uncertainty quantification and hierarchical regularisation. In the context of education assessment, BCJ extends this approach by deriving posterior rank distribution and enabling active learning pair selection strategies.\\

However, the following important limitations remain
\begin{itemize}
    \item Optimal pair selection under limited judgment budgets has only been partially addressed within existing Bayesian BTM models, even though those models provide posterior inference over latent strengths. Existing BCJ implementation primarily aims to reduce local uncertainty rather than explicitly maximising improvements in global ranking accuracy. 
    \item Reliability of ranking in CJ relies on SSR. Although it provides an interpretable measure of inter-item variability, it is sensitive to modelling assumptions and does not reflect posterior rank uncertainty or decision-relevant ranking performance.
    \item KG methods provide a framework for sequential information collection, and have been extended to pairwise and preference learning settings. However, these methods have not been incorporated into hierarchical Bayesian BTM within educational CJ. To our knowledge, there is no existing framework that explicitly connects KG-based active learning pair selection methods implemented within the Bayesian CJ context.      
\end{itemize}

These gaps motivate the development of a unified framework that integrates hierarchical Bayesian BTM, posterior-based reliability assessment, and KG-driven active pair selection. Such a framework aligns sequential decisions with explicit ranking and reliability objectives to improve rank recovery accuracy while efficiently utilising limited human judgments.    

\newpage

\section{Methodology}\label{Method}
\subsection{Bradley Terry Model}
The Bradley-Terry model (BTM) assigns probabilities to the outcomes of pairwise comparisons based on strength parameters. Let $[N] = {1,2,...,N}$ be the set of finite items, and each with a latent score $\theta_i \in \mathbb{R}, i=1,2,..., N $. Then the score vector can be written as $\theta = (\theta_1, \theta_2, ..., \theta_N)^\top$.\\
A comparison between $i$ and $j$ gives a binary outcome $y_{ij} \in \{0,1\}$, where \[
y_{ij}=\begin{cases}
1 & \text{if } i \succ j\\
0 & \text{if } j \succ i
\end{cases}
\]
The Bradley-Terry probability that $i$ is preferred over $j$ is
\begin{equation}
    P(y_{ij} = 1 \mid \theta) = \frac{e^{\theta_i}}{e^{\theta_i}+e^{\theta_j}} = \frac{1}{1+e^{-(\theta_i-\theta_j)}}=\sigma(\theta_i-\theta_j)
\end{equation}
At $t^{th}$ comparison, we can define $x_t = e_{i_t}-e_{j_t}$, so that $x_{t}^\top \theta = \theta_{i_t}-\theta_{i_t}$. The model depends only on score difference. Therefore, one location constraint is required for identifiability. So, we work on the centred subspace.
\[
\mathcal{C} = \{ \theta \in \mathbb{R}^N : 1^\top\theta=0\}
\]

\subsection{Bayesian Bradley Terry Model}
\subsubsection{Prior Selection}
In Bayesian statistics, the prior distribution is highly important for better posterior inferences.~\cite{whelan2017prior} suggests two prior distributions for the BTM, and we adopt the independent Gaussian prior $\theta_i\mid \sigma_{\theta}^2 \sim \mathcal{N}(0, \sigma_{\theta}^2)$ on the student's latent log abilities. Further, we apply the parameter expansion $\theta = \xi \eta$ and place an Inverse-Gamma hyperprior on $\sigma_{\eta}^2$. More details in this can be found in section ~\ref{PX}.   

\subsubsection{Likelihood for Paired Comparisons}
The outcome $y_{t}$ follows a Bernoulli distribution in the $t^{th}$ comparison between $i$ and $j$
\[
y_t\mid \theta \;\overset{\text{ind}}{\sim}\; Bernoulli(\sigma(x_t^\top \theta)), \qquad t=1,2,...,K
\]
where $K$ is the total number of comparisons.\\
Since these outcomes are independent, we can write the cumulative likelihood as follows:
\[
\mathcal{L}_K(\theta) 
= \prod_{t=1}^K 
\left[ \sigma(x_t^\top \theta)^{y_t}
\left(1 - \sigma(x_t^\top \theta)\right)^{1 - y_t} \right]
\]
Then we can write cumulative log likelihood as follows:
\begin{equation}
    \ell_K(\theta) = \sum_{t=1}^K \left[ y_t \, x_t^\top \theta - \log\!\left(1 + e^{x_t^\top \theta}\right) \right]
\end{equation}
The design matrix $X \in \mathbb{R}^{K \times N}$ can be written by stacking these contrasts as rows $x_t^\top$, and the outcome vector $y\in \{ 0,1\}^K$. 

\subsubsection{Polya Gamma Data Augmentation}
To make posterior inference more efficient, we use Polya-Gamma data augmentation for Bernoulli Logistic Bradley Terry likelihood~\cite{polson2013bayesian}. For each pairwise comparison, we model the observed outcome as:
\[
y_t \mid \theta \sim Bernoulli(\sigma(x_t^\top \theta))
\]
Where $\sigma(\cdot)$ is the logistic function. Following Polson et al. (2013), we introduce an auxiliary latent variable $\omega_t$
\[
\omega_t\mid \theta \sim PG(1, x_t^\top \theta)
\]
This choice makes the likelihood conditionally Gaussian and therefore easier to work with. Then we define $\kappa_t = y_t - 1/2$, the augmented likelihood for $\omega_t$ can be written as follows:
\[
p(y_t \mid \theta, \omega_t) \propto \exp\!\left(- \frac{\omega_t}{2} (x_t^\top \theta)^2 + \kappa_t x_t^\top \theta \right)
\]
Given the Polya-Gamma variables, the model simplifies to a Gaussian regression in the log-ability parameters. This ensures straightforward conjugate Gibbs updates in the parameter expanded hierarchical model.

\subsubsection{Parameter Expansion}\label{PX}
\noindent Since commonly used inverse-gamma priors on variance parameters are not recommended due to their undesirable behaviour near zero~\cite{gelman2006prior}, we placed a weakly informative half-t prior on the global scale of the log strength parameters. However, this prior is not conjugate within the resulting hierarchical model, so we considered the parameter expansion suggested~\cite{gelman2006prior}. This strategy represents the half-t prior as a scale mixture of conjugate distributions, enabling efficient Gibbs sampling.

\noindent Parameter expansion introduces:
\[\theta = \xi \eta\]

\noindent with:
\[
\eta_i \sim \mathcal{N}(0, \sigma_\eta^2), \quad
\xi \sim \mathcal{N}(0, s_\xi^2), \quad
\sigma_\eta^2 \sim \text{Inv-Gamma}(a_0, b_0).
\]
\noindent This induces a conditionally Gaussian prior $\theta_k \mid \xi, \sigma_{\eta}^2 \sim \mathcal{N}(0, \xi^2 \sigma_{\eta}^2)$ on the log strength parameters. The effective scale $\sigma_\theta = \lvert \xi \rvert \sigma_\eta$ is marginally distributed as a half-$t$ distribution. 

\subsubsection{Hierarchical structure of the model}
For comparisons $t = 1,2,...,K$, the full hierarchical structure of BBTM can be written as:
\[
\begin{aligned}
y_t \mid \theta &\sim \mathrm{Bernoulli}\!\left(\sigma(x_t^\top \theta)\right), \\
\omega_t \mid \theta &\sim \mathrm{PG}\!\left(1, x_t^\top \theta\right), \\
\theta &= \xi \eta, \\
\eta \mid \sigma_\eta^2 
&\sim \mathcal{N}_{\mathcal{C}}\!\left(0, \sigma_\eta^2 I_c\right), 
\quad \mathbf{1}^\top \eta = 0, \\
\xi &\sim \mathcal{N}(m_\xi, s_\xi^2), \\
\sigma_\eta^2 &\sim \mathrm{Inv\text{-}Gamma}(a_0, b_0).
\end{aligned}
\]

The joint posterior is
\[
\begin{aligned}
p(\eta, \xi, \sigma_\eta^2, \omega \mid y) \;\propto\;
\underbrace{\prod_{t=1}^{K}\exp\!\left\{\kappa_t\, \xi\, x_t^\top \eta\;-\;
\frac{1}{2}\,\omega_t \bigl(\xi\, x_t^\top \eta\bigr)^2\right\}
\, p_{\mathrm{PG}}(\omega_t \mid 1, 0)
}_{\text{augmented likelihood at } \theta = \xi \eta}
\;\cdot\;\underbrace{p(\eta \mid \sigma_\eta^2)\, p(\xi)\, p(\sigma_\eta^2)
}_{\text{prior + hyperprior}}.
\end{aligned}
\]

\subsubsection{Gibbs sampling}
BBTM employs a four block Gibbs sampler that alternate between the following updates. Let $\kappa_t = y_t-\frac{1}{2}$, $\Omega_K = \text{diag}(\omega_t)$, $A_K=X^\top \Omega_K X$, and $r_K=X^\top\kappa$
\begin{enumerate}
    \item \textbf{Polya Gamma latent variables}. Draw $\omega_t\mid \cdot \sim PG(1, \xi x_t^\top \eta)$ for $t=1,2,...,K$, transforming the logistic likelihood into conditionally Gaussian form.
    \item \textbf{Direction $\eta$ on $\mathcal{C}$}. Draw $\eta \mid \cdot \sim \mathcal{N}_{\mathcal{C}}(\mu_n, \Lambda_{\eta}^{-1})$ with $\Lambda_{\eta} = \xi^2 A_K + \sigma_{\eta}^{-2}I_{\mathcal{C}}$ and $\mu_{\eta} = \Lambda_{\eta}^{-1} \xi r_K $ with $1^\top \eta = 0$
    \item \textbf{Global scale $\xi$}. Draw $\xi \mid \cdot \sim \mathcal{N}(\mu_{\xi}, \nu_{\xi})$ with $\nu_{\xi}^{-1} = \eta^\top A_K \eta + s_{\eta}^{-2}$ and $\mu_{\xi} = \nu_{\xi}(\eta^\top r_K + m_{\xi}/s_{\xi}^2)$
    \item \textbf{Variance $\sigma_{\eta}^2$}. 
    Draw $\sigma_{\eta}^2 \mid \cdot \sim \mathrm{Inv\text{-}Gamma}(a_0 + N/2,\; b_0 + \tfrac{1}{2}\lVert \eta \rVert^2)$.
    \item \textbf{Recover strengths}. Set $\theta = \xi \eta \in \mathcal{C}$
\end{enumerate}
This whole process yields an efficient, fully Gibbs sampler with good mixing properties even under sparse comparison data.

\subsubsection{Sigmoid-Normal Approximation}
\noindent When computing Knowledge Gradient (KG)~\cite{frazier2008knowledge} metrics, the expected utility of each item comparison must be evaluated under a hypothetical future outcome. This would require running separate MCMC samples for each possible outcome of each unique pair, which becomes computationally infeasible as the number of items grows. For example, when the number of items is 10, there are 45 unique pairs, and it should run 90 MCMC samplers, as it considers both outcomes of each pair. When the number of items increases to 20, the number of unique pairs increases to 190, requiring 380 MCMC runs. So, the number of required MCMC runs grows quadratically with the number of items.\\

\noindent To make this more computationally efficient, we replace the expensive MCMC with a fast analytical approximation. To evaluate the predictive probability of a comparison outcome, we must compute the expectation of the logistic likelihood over the Gaussian posterior distribution of the skill difference. This requires solving the convolution of a logistic sigmoid $\sigma(z)$ with a Gaussian $\mathcal{N}(\mu, \sigma^2)$, which is analytically intractable.\\

\noindent We employ Mackay's Probit Approximation~\cite{bishop2006pattern} to solve this. This method approximates the logistic sigmoid with a scaled Probit function, $\Phi(\lambda z)$, where the scaling factor $\lambda = \sqrt{\pi/8}$ is derived by matching the slopes of the functions at the origin. This allows for a closed-form analytic update where the steepness of the sigmoid is moderated by the uncertainty $\sigma^2$. The approximate expected win probability is given by:
\[
Pr(i \succ j) \approx \sigma\!\left(\frac{\mu_{ij}}{\sqrt{1+\frac{\pi}{8}\,\sigma_{ij}^2}}\right)
\]
Where $\mu_{ij}$ is the mean difference and $\sigma^2_{ij}$ is the variance of that difference.

\subsection{Novel Pair Selection Methods}
It is expensive to obtain comparisons in the real world for rankings. We have implemented a sophisticated active learning strategy known as Knowledge Gradient (KG)~\cite{frazier2008knowledge}\cite{frazier2009knowledge} to intelligently select the most informative pairwise comparisons. In the following section, we will outline the general mathematical framework of the KG and describe its application to our probabilistic ranking model. Further, we detail three distinct utility functions that are used to define “Knowledge”: Kendall’s Tau distance~\cite{kendall1938new}, Entropy of the posterior rank distribution, and separation of the posterior distributions. The objective of this implementation is to maximise the rate of learning and converge to an accurate ranking with a minimum number of pairwise comparisons.\\

\noindent The selection of which pairs of items to compare next to most efficiently improve our understanding of the overall ranking is the central problem with the ranking of paired comparisons. A random selection method might miss valuable information as it can pick the less informative pairs, for a given fixed budget $N*K$ of total comparisons. An entropy driven pair selection method will select the pair that has the highest uncertainty. However, it might be less informative in improving the final rank order of all items. For instance, let’s suppose that, after several rounds of comparisons of 5 items $(A, B, C, D, E)$ and we are very confident that Item $A$ and Item $B$ are the two best, but we are extremely uncertain who is better. They have the highest pairwise entropy, i.e., $\Pr(A \succ B)=0.5$. Further, we have some information about Item $C$ and $D$ and believe that Item $C$ is moderately certain than Item $D$, i.e., $\Pr(C \succ D)=0.7$. We have learned that Item E loses to all Items, and we are very much certain that $\Pr(A \succ E)=0.9$ and $\Pr(B \succ E)=0.9$. But we have never compared Item $C$ and Item $E$ as we do not have their data. However, the model’s prior belief assumes they also have the highest pairwise entropy. In the entropy selection method, it looks at all possible pairs and it might choose pair $(C, E)$ over pair $(A, B)$. This is less informative as Item $C$ is more certain outcome in this pair from the transitivity principle. If the algorithm had chosen pair $(A, B)$ instead, the outcome would have been highly informative. The KG metrics are specifically designed to avoid these kinds of scenarios. KG metrics would have simulated both pairs and concluded that pair $(A, B)$ would lead to a higher improvement in the final rank order, making it the better choice.

\subsubsection{Mathematical Framework}
In this work, we have implemented adaptive strategies for selecting pairs in a pairwise comparison-based ranking problem, using a Bayesian Bradley-Terry model with Polya-Gamma data augmentation. To decide which pair to compare next, we introduce the following three active learning metrics. The goal is to maximise information gain as possible by choosing the pair that, depending on its outcome, would either disrupt the current rank order, or reduce uncertainty, or separation of the posterior distributions.

\begin{enumerate}
    \item \textbf{KG-Tau}\\
    The KG-Tau matrix is formulated to find the comparison that maximises the expected change in our model's existing rank order. We measure this change using the normalised Kendall's Tau distance between the current rank order and after adding a new comparison to the existing comparisons. A larger expected distance indicates a more informative comparison that leads to faster learning.\\
    
    Let's consider a latent variable $\pi_{i}$ as $\beta_{i}=log(\pi_{i})$. Then from the equation 2, we can map the difference $\theta_{i,j} = \sigma(\beta_i-\beta_j)$ to $[0,1]$.Then the outcome function, $\mathcal{O}(x)$ can be defined as follows,
    \[
    \mathcal{O}(x) =
    \begin{cases}
        0, & 0 < x < 0.5\\
        1, & 0.5 < x < 1
    \end{cases}  
    \]

    The density of $x$,$f_{\delta_{ij}}(x)$, is induced by the posterior distribution of $\delta_{ij}$. Then we can calculate the expected tau distance using the following equation:
    \begin{equation}
        \mathbb{E}(\tau) = \int_{0}^{1}\tau(\mathcal{R}_{c}, \mathcal{R}_{n}(x))f_{\delta_{ij}}(x) dx
    \end{equation}

    \begin{equation}
        \mathbb{E}(\tau) = \int_{0}^{0.5}\tau(\mathcal{R}_{c}, \mathcal{R}_{n}(x))f_{\delta_{ij}}(x) dx + \int_{0.5}^{1}\tau(\mathcal{R}_{c}, \mathcal{R}_{n}(x))f_{\delta_{ij}}(x) dx
    \end{equation}

    Since $\tau(\mathcal{R}_{c}, \mathcal{R}_{n}(x)$ is constant in each interval $(\kappa_{0} \text { for } 0 < x <0.5, \; \kappa_{1} \text{ for } 0.5 < x < 1)$, we can simplify equation 16 as follows:
    \begin{equation}
        \mathbb{E}(\tau) = \kappa_{0}\int_0^{0.5}f_{\delta_{ij}}(x) dx + \kappa_{1}\int_{0.5}^1f_{\delta_{ij}}(x) dx
    \end{equation}

    Where:
    \begin{itemize}
        \item $\int_{0}^{0.5}f_{\delta_{ij}}(x) dx = Pr(x<0.5) = Pr(\delta_{ij}<0) = 1 - Pr(i\succ j)$
        \item $\int_{0.5}^{1}f_{\delta_{ij}}(x) dx = Pr(x \geq 0.5) = Pr(\delta_{ij}>0) = Pr(i\succ j)$
    \end{itemize}

    Hence, the integral form reduces to 
    \begin{equation}
        \mathbb{E}(\tau) = \kappa_{0}(1 - Pr(i \succ j)) + \kappa_{1}Pr(i \succ j)
    \end{equation}
     The outcome probability can be calculated as follows
   \begin{equation}
    p = \Pr(i \succ j) = \Pr(\lambda_i > \lambda_j \mid \mathcal{D}) = \Phi\left(\frac{\delta_{\mu}}{\sqrt{\delta_{\sigma}^2}}\right)
    \end{equation}
    where $\Phi$ is the standard normal CDF and   
    \begin{align*}
        \delta_{\mu} = \lambda_{mean}[i] - \lambda_{mean}[j] \\
        \delta_{\sigma}^2 = \Sigma_{[i,i]}+\Sigma_{[j,j]}-2\Sigma_{[i,j]} \\
    \end{align*}

    Select the pair that maximises the expected tau. 

    Kendall's tau distance measures ranking disagreement via discordant pairs, and a high expected tau indicates the pair that makes the highest disruption to the current rank order, providing high information value.
    
    \item \textbf{KG-Entropy}\\
    Entropy measures the uncertainty in the posterior rank probabilities. For each item $k$, the rank probability distribution $r_{k} = \mathrm{rank\_dist}[k,:]$. 
    \begin{equation}
        H(r_{k}) = - \sum_{l=0}^{N-1} r_{k,l}\log_{2}(r_{k,l})
    \end{equation}
    Then the average entropy
    \begin{equation}
        H_{avg} = \frac{1}{N}\sum_k H(r_{k})
    \end{equation}
    Then the expected entropy is calculated as follows:    
    \begin{equation}
    \mathbb{E}({\Delta H}) = Pr(\mathcal{O}(x)=0 \mid i \succ j)\Delta H(\mathcal{R}_{c}, \mathcal{R}_{n}(0)) + Pr(\mathcal{O}(x)=1 \mid i \succ j)\Delta H(\mathcal{R}_{c}, \mathcal{R}_{n}(1)) 
    \end{equation}
   
    We can think of the entropy difference as the difference between the current entropy and the entropy after adding new data. So, we can write it as follows:
    \begin{equation}
        \Delta H (\mathcal{R}_{c}, \mathcal{R}_{n}(C) = H(\mathcal{R}_{c}) - H(\mathcal{R}_{n}(C))
    \end{equation}    
    
    Since entropy reduces with the addition of new data, it is always \\ $H(\mathcal{R}_{c}) > H(\mathcal{R}_{n}(C)$. \\ 

     Let's assume $p_{1} = (1 - Pr(i \succ j)) \text { and } p_{0} = Pr(i \succ j)$, then we can write equation 10 as follows:
     \begin{equation}
    \mathbb{E}({\Delta H}) = p_{0} \Delta H(\mathcal{R}_{c}, \mathcal{R}_{n}(0)) + p_{1}\Delta H(\mathcal{R}_{c}, \mathcal{R}_{n}(1))
    \end{equation}

    Then we can rewrite equation 12 as follows   
    \begin{align*}
    \mathbb{E}({\Delta H}) = p_{0}[H(\mathcal{R}_{c}) - H(\mathcal{R}_{n}(0))] + p_{1}[H(\mathcal{R}_{c}) - H(\mathcal{R}_{n}(1))] \\
    \end{align*}
    \begin{equation}
    \mathbb{E}({\Delta H}) = (p_{0}+p_{1})H(\mathcal{R}_{c})-[p_{0}H(\mathcal{R}_{n}(0))+p_{1}H(\mathcal{R}_{n}(1))]    
    \end{equation}
     
    In equation 13 $(p_{0}+p_{1})H(\mathcal{R}_{c})$ is a constant. So, we have to minimise the $[p_{0}H(\mathcal{R}_{n}(0))+p_{1}H(\mathcal{R}_{n}(1))]$ to get the largest reduction in expected entropy.
   
    \item \textbf{KG-SSR}\\
    The primary objective of a comparative ranking model is to distinguish items with higher confidence. Due to the overlap of posterior distributions and the creation of noise that masks the true ranking order of items, these ranking tasks will become more difficult.\\    
    Most of the time, standard active learning methods focus on local pairwise uncertainty or rank stability. Hence, these methods may not directly optimise the global clarity of the rankings. So, we propose a new Knowledge Gradient acquisition function $(KG-SSR)$ that selects the comparisons maximising the expected increase in the Scale Separation Reliability (SSR).\\    
    The acquisition function explicitly targets the global separability of the items and drives the system towards a state where items are distinct, and the rankings are unambiguous.\\
    Let the current state at time $t$ be $S_{t}$, parametrised by a mean vector $\mu^{(t)}$ and a covariance matrix $\Sigma^{(t)}$ for $N$ items. Further, we extract the variance vector for the items as $\nu = diag(\Sigma^{(t)})$. Then we define the utility function $SSR$ to quantify the ratio as follows: 
    \[
    SSR(S_{(t)}) = \frac{var(\mu_{i}) - \frac{1}{N} \sum_{i=1}^{N} \sigma_{i}^2}{var(\mu_{i})}
    \]
    KG-SSR selects the pair $(i,j)$ that maximises the expected gain in this SSR metric. We look one step ahead into the future:
    \[
    KG_{SSR}(S_{t}, S_{t+1}\mid (i,j)) = \mathbb{E}_{y_{ij}}[\mathrm{SSR}(S_{t+1})] - \mathrm{SSR}(S_t)
    \]
    Expanding the expectation over the two possible outcomes (item $i$ wins or item $j$ wins)
    \[
    KG_{SSR}(S_{t}, S_{t+1}\mid (i,j)) = [Pr(i\succ j)\cdot SSR(S_{t+1}^{(1)}) + Pr(j\succ i)\cdot SSR(S_{t+1}^{(0)}) ] - SSR(S_{t})
    \]
  
\end{enumerate}

\subsection{Fast Knowledge Gradient via One-step Moment updates for BBTM}
KG methods were developed in ranking and selection for sequential measurement selection and were later extended to Bayesian optimisation~\cite{frazier2008knowledge}. Since human judgments are costly and the aim is to select comparisons that yield the optimal rank order within a limited budget, KG methods can be highly effective in CJ. A direct KG implementation can be computationally expensive and time-consuming, as it requires evaluating each item pair, which would require repeated posterior inference under hypothetical outcomes. To address this, we implemented a one-step look-ahead moment approximation that allows analytical moment updates and replaces nested posterior sampling, which makes KG implementation feasible in practice.\\

To the best of our knowledge, this is the first integration of KG with Bayesian Bradley-Terry in a CJ environment that avoids nested MCMC sampling via analytical one-step moment updates. \\

\noindent\textbf{The Predictive Distribution and MacKay's Approximation}\\
Consider a hypothetical comparison between item $i$ and $j$. Let $\delta_{ij}=\lambda_i-\lambda_j$ denote the latent skill difference. Under our current Gaussian posterior belief, $\lambda \sim \mathcal{N}(\mu, \Sigma)$, the marginal distribution of the difference is a univariate Gaussian, $\delta_{ij} \sim \mathcal{N}(m_{ij}, \nu_{ij})$, where:
\[
m_{ij} = \mu_i-\mu_j
\quad
\nu_{ij}=\Sigma_{ii}+\Sigma_{jj}-2\Sigma{ij}
\]
The BTM likelihood of item $i$ winning is given by the logistic sigmoid function, $\sigma(\delta_{ij})$. We need to marginalise over the latent skill difference to compute the predictive probability of this outcome:
\[
p(y = +1 \mid m_{ij}, v_{ij}) 
= \int \sigma(\delta_{ij}) \mathcal{N}(\delta_{ij} \mid m_{ij}, v_{ij}) \, d\delta_{ij}
\]
We use MacKay's probit approximation~\cite{bishop2006pattern}, as the convolution of a Gaussian and a logistic sigmoid is analytically intractable. The predictive probability can be expressed in closed form by approximating the logistic distribution with a normal distribution with variance $8/\pi$. 
\[
Pr(y\mid m_{ij}, \nu_{ij})\approx \Phi\left(\frac{y\cdot m_{ij}}{\sqrt{\nu_{ij}+\frac{8}{\pi}}}\right)
\]
where $\Phi(\cdot)$ is the CDF of the standard normal distribution, and $y\in \{+1, -1\}$ shows the winner.\\

\noindent\textbf{Analytic posterior updates}\\
After the outcome $y$ from hypothetical comparison, the intermediate true posterior is proportional to $\sigma(y\delta_{ij})\mathcal{N}(\lambda\mid \mu, \Sigma)$. This is no longer strictly Gaussian. However, ADF converts this back to Gaussian, $\mathcal{N}(\mu_{new}, \Sigma_{new})$, by computing the mean and covariance of the intermediate distribution.\\
Let the total assumed standard deviation $s=\sqrt{\nu_{ij}+8/\pi}$, and normalised margin $u = \frac{y\cdot m_{ij}}{s}$ of the outcome. Then, define a truncation factor based on the inverse Mills Ratio to get the moment match updates as follows:
\[
\beta(u) = \frac{\phi(u)}{\Phi(u)}
\]
Here $\phi(\cdot)$ and $\Phi(\cdot)$ represent PDF and CDF of the standard normal distribution, respectively.
We can calculate closed-form updates for the mean vector and covariance matrix as follows:
\[
\mu_{new} = \mu + y \frac{\beta(u)}{s}(\Sigma_{\cdot, i}, \Sigma_{\cdot, j})
\]
\[
\Sigma_{new} = \Sigma - \frac{\beta(u)(\beta(u)+u)}{s^2}(\Sigma_{\cdot, i}, \Sigma_{\cdot, j})(\Sigma_{\cdot, i}, \Sigma_{\cdot, j})^T
\]
Here, $(\Sigma_{\cdot, i}, \Sigma_{\cdot, j})$ is the column vector of the covariance of every item in the dataset with the specific pairwise difference.\\

We can calculate analytical posterior $(\mu_{new}^{(y)}, \Sigma_{new}^{(y)})$ for any hypothetical comparison quickly.\\

\noindent\textbf{KG computation}\\
For a utility $U(\mu, \Sigma)$, the KG value for a hypothetical pair $(i,j)$ is the expected on-step gain.
\[
KG_U(i,j) = \mathbb{E}_{y\mid\mathcal{D}_t}[U(\mu_t^{(y)},\Sigma_t^{(y)})] - U(\mu_t,\Sigma_t)
\]
\[
KG_U(i,j) = Pr(i \succ j)[U(\mu_t^{(+1)},\Sigma_t^{(+1)})] + Pr(j \succ i)[U(\mu_t^{(-1)},\Sigma_t^{(-1)})] - U(\mu_t,\Sigma_t)
\]
The next comparison is selected as
\[
(i^*, j^*)= \arg\max_{(i,j)\in D_t}KG_U(i,j)
\]

\subsection{Model Robustness Check}
\noindent To validate the robustness of the active learning model, we have conducted a controlled experiment to quantify its performance degradation in the presence of noisy data. The objective was to identify the minimum proportion of true skill-based outcomes required to recover an accurate ranking.\\

\noindent We introduced a parameter $\alpha$ (threshold), which controls the signal-to-noise ratio of the comparison outcomes. This parameter represents the probability that a given comparison result reflects the true latent ability of items rather than a random guess. The experiment was conducted across a range of threshold levels $\alpha \in \{0.0, 0.1, 0.2, \ldots, 1.0\}$, where
\begin{itemize}
    \item $\alpha = 0.0$: All outcomes are determined by a random coin toss, independent of item skills
    \item $\alpha = 1.0$: All outcomes are determined by the item's true skill differences.
\end{itemize}
For every pair $(i,j)$ selected by the pair selection method, the outcome $y_{ij}$ was generated using the following decision rule:
\begin{enumerate}
    \item Draw a random variable $u \sim \mathcal{U}(0,1)$.
    \item If $u \le \alpha$, the outcome is generated by the true skills of the item. The winner is determined by comparing samples drawn from the item's true skill distributions. 
    \item if $u > \alpha$, the outcome is generated randomly. The winner is selected with equal probability, providing no information about the true skills. 
\end{enumerate}
\noindent The main goal of this experiment is to assess the rank recovery potential under imperfect conditions. By varying $\alpha$, we aim to:
\begin{enumerate}
    \item Quantify Tolerance: Determine the threshold where model performance stabilises, indicating the minimum proportion of skill-based outcomes required to achieve near optimal ranking. 
    \item Validate Efficiency: Depicts that the model does not require perfect data to converge. For example, if the ranking accuracy remains stable at $\alpha = 0.7$, it shows that the model can successfully recover the true ranks even with 30\% coin toss outcomes.
    \item Establish Reliability: Confirm that the active learning method focuses on the signal within the comparisons rather than over-fitting to the noise.
\end{enumerate}

\noindent For each noise level, we conducted 50 independent runs to ensure statistical reliability. The Normalised Kendall's Tau distance was used to measure the rank accuracy between the model's posterior ranking and the true rankings.

\subsection{Posterior Separation Reliability (PSR) for BBTM}
It is essential that estimated latent strengths are sufficiently well separated to recover a reliable ranking order in CJ. In traditional CJ, SSR is used for this purpose, which is based on point estimates and associated error terms~\cite{verhavert2018scale}. However, within a Bayesian version of BTM framework, posterior inference provides both the estimated strengths and their uncertainty. Therefore, it is straightforward and internally consistent to define a reliability measure directly from the posterior distributions.\\   
Conceptually, PSR measures the extent to which the spread in posterior mean strengths across items exceeds the average posterior uncertainty in those items, thereby quantifying the degree of effective separation achieved by the model. We can define PSR as follows:

\[
PSR = 1 - \frac{\frac{1}{N}\sum_{i=1}^{N}\sigma_i^2}{\operatorname{Var}(\{\mu_i\}_{i=1}^{N})}
\]
Where $\mu_{i} = \mathbb{E}(\theta_{i}\mid \mathcal{D})$, $\sigma_{i}^2 = Var(\theta_{i}\mid \mathcal{D})$, $Var(\cdot)$ denotes the variance operator, and $N$ = Number of items.

The proposed PSR has the maximum value of 1 when there is no posterior uncertainty in the latent strengths. However, unlike other reliability measures, PSR has no finite lower bound. PSR values close to 1 indicate strong posterior separation, values closer to 0 mean posterior strength separation and uncertainty are similar, and negative values demonstrate that posterior uncertainty exceeds the spread of the posterior means.\\

\newpage

\section{Convergence of the Bayesian Bradley Terry Model}\label{Conv_proof}
We give a full convergence proof for the BBTM on the centred identifiable subspace. We allow the comparison schedule to be predictable. This means that the pair $(i_t, j_t)$ compared at time $t$ may be fixed in advance, chosen randomly without reference to the outcome, or adapted based on previous comparisons and their outcomes. Under these conditions, we cover many possible comparison schemes, including no repeating pairs, random, round robin method, and certain active learning methods, such as entropy and novel knowledge gradient methods. The only condition is that they must satisfy our assumptions about how information grows over time. If the long-run frequency of each pair is positive, the marginal posterior is consistent at the true centred score vector. We show that the posterior is locally tight when the normalised Fisher information matrix is assumed to have a limit. This is because of the consistency of the marginal posterior at the true centred score, and the local strong concavity of the logistic likelihood. This allows us to prove a Bernstein-von Mises theorem on the rescaled posterior, where the asymptotic covariance is given by the inverse limiting information matrix. Consequently, ranking consistency then follows when the true scores are strictly separated. We do not need a separate proposition for any specific active learning policy. Such a proposition is only needed if someone wants to prove that a particular method, for example KG rule, satisfies the information growth assumptions we used.\\

\subsection{Assumptions}
\begin{enumerate}
    \item \textbf{Finite number of items}. The total number of items $N < \infty$ is fixed. The true latent score vector $\theta^* \in \mathcal{C}$ is fixed and unknown.
    \item \textbf{The scores are strictly separable}. No two scores are equal.This means that $|\theta_i^*-\theta_j^*| \neq 0,\quad \forall i \neq j$. This ensures that the true ranking $\pi^*$ is well-defined.
    \item \textbf{Adaptive Bradley-Terry sampling}. Let $\mathcal{F}_t = \sigma\bigl((i_1,j_1,y_1),...,(i_t,j_t,y_t)\bigr)$ be the comparison history filtration. At time $t$, the selected pair $(i_t,j_t)$ is $\mathcal{F}_{t-1}$ measurable, i.e., chosen based on the past comparison outcomes.
    \[
    y_t|\mathcal{F}_{t-1}, (i_t,j_t) \sim \text{Bernoulli}(\sigma(x_t^\top\theta^*)) 
    \]
    Then, $y_t-\sigma(x_t^\top\theta^*)$ is a martingale difference.
    \item \textbf{Polya-Gamma data augmentation}. The Gibbs sampler is designed, which uses exact Polya-Gamma data augmentation ~\cite{polson2013bayesian}. The auxiliary variable $\omega_t|\theta, y_t \sim \textbf{PG}(1, x_t^\top \theta)$ are sampled exactly, and the full conditional $(\theta | y, \omega)$ is a multivariate Gaussian computed in closed form on $\mathcal{C}$. This is an exact data augmentation, not an approximation.
    \item \textbf{Information growth}. Three types of information growth conditions are defined here.
        \begin{enumerate}[label=\arabic{enumi}\alph*.] 
            \item \textbf{Consistency-level information growth}. The cumulative Fisher information on $\mathcal{C}$ is
            \[
             \mathcal{J}_K(\theta^*) = \sum_{t=1}^K\sigma(x_t^\top\theta^*)\bigl(1-\sigma(x_t^\top\theta^*))\bigr) \cdot x_tx_t^T
            \]
            A minimal consistency level information condition is $\lambda_{min, \mathcal{C}}(\mathcal{J_K(\theta^*)}) \to \infty$ as $K \to \infty$. This means information accumulates in every identifiable direction.
            \item \textbf{Positive pair frequency growth}. There exists $p_{min} >0$ s.t., for every pair $(i,j)$
            \[
            \liminf_{K \to \infty} \frac{k_{ij}(K)}{K} \ge p_{\min} > 0
            \]
            where $k_{ij}(K)$ is the number of times pair $(i,j)$ has been compared in total comparisons $K$. This implies 5a and is sufficient for the posterior consistency through the KL argument.
            \item \textbf{Rate level information limit}. The normalised Fisher information converges on $\mathcal{C}$
            \[
            \frac{1}{K}\mathcal{J}_K(\boldsymbol{\theta}^*) \to \bar{\mathcal{I}}(\boldsymbol{\theta}^*) \succ 0
            \]
            This condition is required for BvM theorem and the $1/\sqrt{K}$ convergence rate of posterior distribution. Notice that 5b does not imply 5c. The pair frequencies $\frac{k_{ij}(K)}{K}$ may be bounded away from zero for all $(i,j)$ but not necessarily converge to any limit. 
        \end{enumerate}
    \item \textbf{Proper Gaussian prior on $\mathcal{C}$}. Each $\theta_i$ drawn independently from $\theta_i \sim \mathcal{N}(0, \sigma^2)$, with $0<\sigma<\infty$. The prior has full support on $\mathcal{C}$, and the variance $\sigma^2$ is not assigned a hyperprior.\\
    
    \textbf{Remark } For theoretical simplicity, this argument is presented under a fixed prior variance. In our implementation, however, we use a weakly-informative half-$t$ prior on the global scale, represented through its conjugate scale-mixture form with parameter expansion to improve MCMC mixing, as discussed in Section~3.2.4. This does not change the convergence results, because the proofs require only that the prior has full support and is locally positive in a neighbourhood of $\theta^*$.       
    \item \textbf{Identifiability constraint}. The location non-identifiability of the BTM can be fixed by setting $1^\top \theta = 0$, $1^\top \theta^* = 0$.    
\end{enumerate}

\subsection{Foundation Lemmas}
\textbf{Lemma 1.} \textbf{Exact Gaussian full conditional under PG data augmentation}. \textit{The full conditional posterior of $\theta \in \mathcal{C}$ given comparison outcomes $y = (y_1,y_2,...,y_K)$ and the auxiliary variables $\omega = (\omega_1,\omega_2,..., \omega_K)^T$ is the multivariate Gaussian under assumptions \textbf{3}, \textbf{4}, \textbf{6} and \textbf{7} \[
\theta \mid y_K, \omega \sim \mathcal{N}_{\mathcal{c}}(\mu^*(\omega), \Lambda^*(\omega)^{-1})
\] where \[
\Lambda^*(\omega) = X^\top \text{diag}(\omega)X + \Lambda_0,
\]
\[
\mu^*(\omega) = \Lambda^*(\omega)^{-1}X^\top\kappa, \qquad \kappa_t=y_t-\frac{1}{2}
\] and $\Lambda_0=\sigma^{-2}_0 I_{\mathcal{C}}$ is the fixed prior precision on ${\mathcal{C}}$}.\\

\noindent \textit{Proof.} The Polya-Gamma identity ~\cite{polson2013bayesian} states that, for any $\psi \in \mathbb{R}$ and $y\in \{0, 1\}$
\begin{equation}
    \sigma(\psi)^y\{1-\sigma(\psi)\}^{1-y} = \frac{e^{\kappa\psi}}{2}\int_0^\infty
e^{-\omega \psi^2 / 2}
\, p(\omega)\, d\omega
\end{equation}
where $\kappa=y-\frac{1}{2}$ and $p(\omega)$ is the density of PG(1,0).
BT logistic likelihood is not conjugate with the Gaussian prior. However, Polya-Gamma data augmentation turns the logistic likelihood conditionally Gaussian, enabling efficient Gibbs sampling~\cite{polson2013bayesian}. Equivalently, the conditional form can be written as follows.
\begin{equation}\label{Identity}
    \sigma(\psi)^y\{1-\sigma(\psi)\}^{1-y} \propto_{\theta}\exp\left(\kappa\psi - \frac{\omega \psi^2}{2}\right)
\end{equation}
where $\omega\mid \psi \sim PG(1, \psi)$.
The linear predictor of BTM is $\psi_t = x_t^\top\theta = \theta_{i_t}-\theta_{j_t}$ and substituting this into equation \ref{Identity} gives
\begin{equation}
    \sigma(x_t^\top\theta)^{y_t}\{1-\sigma(x_t^\top\theta)\}^{1-y_t} \propto_{\theta}\exp\left(\kappa_t x_t^\top\theta - \frac{\omega_t (x_t^\top\theta)^2}{2}\right)
\end{equation}
This is an exact form of a Gaussian kernel. By Assumption 3, conditional on history $\mathcal{F}_{t-1}$ and selected pair $(i_t, j_t), y_t$ values are conditionally independent. So, the joint augmented likelihood for all $K$ comparisons as follows:
\begin{equation}\label{full_lh}
    \prod_{t=1}^K\sigma(x_t^\top\theta)^{y_t}\{1-\sigma(x_t^\top\theta)\}^{1-y_t} \propto_{\theta}\exp\left(\sum_{t=1}^K\kappa_t x_t^\top\theta - \frac{1}{2} \sum_{t=1}^{K}{\omega_t (x_t^\top\theta)^2}\right)
\end{equation}
Converting the linear and quadratic terms of equation ~\ref{full_lh} into matrix form, equation ~\ref{full_lh} becomes:
\begin{equation}
     \prod_{t=1}^K\sigma(x_t^\top\theta)^{y_t}\{1-\sigma(x_t^\top\theta)\}^{1-y_t} \propto_{\theta}\exp\left(\theta^{\top} X^{\top} \kappa - \frac{1}{2} \theta^{\top} X^{\top} \operatorname{diag}(\omega) X \theta\right)
\end{equation}
The Gaussian prior $\theta\sim\mathcal{N}_{\mathcal{C}}(0, \sigma^2_0 I_{\mathcal{C}})$ has density 
\[
\pi_0({\theta}) \propto \exp \left(-\frac{1}{2} \theta^\top \Lambda_0\theta\right), \qquad \Lambda_0 = \sigma_0^{-2}I_{\mathcal{C}}
\]
The posterior distribution can be written as follows, according to the Bayes' rule
\[
\pi(\theta\mid y_K, \omega) \propto p(y_K \mid \theta, \omega) \cdot \pi_0({\theta})
\]
Therefore, the full conditional posterior is
\begin{equation}\label{Posterior}
    \pi(\theta\mid y_k, \omega) \propto \exp\left(\theta^{\top} X^{\top} \kappa - \frac{1}{2} \theta^{\top} (X^{\top} \operatorname{diag}(\omega) X +\Lambda_0)\theta\right)
\end{equation}
The canonical form of a standard Gaussian is
\[
\mathcal{N}(\mu, \Lambda^{-1}) \propto \exp \left(\theta^\top\Lambda\mu - \frac{1}{2} \theta^\top\Lambda\theta \right)
\]
Equation ~\ref{Posterior} matches this canonical form with $\mu^*(\omega) = \Lambda^*(\omega)^{-1}X^\top\kappa$ and $\Lambda^*(\omega) = X^\top \operatorname{diag}(\omega)X + \Lambda_0$, hence the full conditional posterior is well defined on $\mathcal{C}$.
\\

\noindent \textbf{Remark }It should be noted that Lemma 1 is an exact statement about the conditional posterior $p(\theta\mid y,\omega)$, i.e. the posterior given the PG auxiliary variables. This does not, however, imply that the marginal posterior $p(\theta\mid y) = \int p(\boldsymbol{\theta}\mid y, \omega)p(\omega \mid y)d\omega$ is Gaussian in finite samples. In fact, the marginal posterior is a scale mixture of Gaussians, which is generally non-Gaussian for finite $K$. The asymptotic Gaussianity of the marginal posterior is a separate phenomenon, driven by increasing information, and we establish this via the BvM theorem in Lemma 8.\\

\noindent \textbf{Lemma 2.} \textbf{Conditional posterior precision grows linearly on $\mathcal{C}$}. \textit{Under assumptions \textbf{3, 4, 5b, 6} and \textbf{7} for PG variables drawn from their full conditional distribution, we have \[ \lambda_{\min, \mathcal{C}}(\Lambda^*(\omega))=\Omega(K)\] almost surely. It follows that, \[\left\| \Lambda^{*}(\omega)^{-1} \right\|_{C} = O(K^{-1})\] almost surely. Additionally, the conditional posterior variances of the identifiable centred directions and pairwise contrasts are of order $O(K^{-1})$.}\\

\noindent \textit{Proof. } The conditional posterior precision can be written as follows:
\[
\Lambda^*(\omega) = \Lambda_{data}(\omega) + \Lambda_0
\]
where $\Lambda_0 = \sigma_0^{-2}I_{\mathcal{C}}\succeq 0$ is the prior precision. Hence, it suffices to show 
\[ \lambda_{\min, \mathcal{C}}(\Lambda_{data}(\omega)) =\Omega(K) \quad \text{a.s.}\]
For all comparisons of pair $(i,j)$, design vectors $x_t = e_{i_t}-e_{j_t}$ are same. Then \[
\Lambda_{data}(\omega) = \sum_{i<j}\left(\sum_{t:(i_t,j_t)=(i,j)}\omega_t\right)(e_i-e_j)(e_i-e_j)^\top
\]
Define the aggregated edge weight: \[w_{ij}(K) := \sum_{t:(i_t,j_t)=(i,j)}\omega_t\]
Then \[
\Lambda_{data}(\omega) = \sum_{i<j}w_{ij}(K)(e_i-e_j)(e_i-e_j)^\top
\]
This is exactly the weighted graph Laplacian for the complete graph $K_N$ with edge weights $w_{ij}(K)$. 

For a fixed pair $(i,j)$, the linear predictor $ x_t^\top \theta = \theta_i-\theta_j$ is same. Therefore, $\omega_t$ values conditionally i.i.d. given current $\theta$
\[
\omega_t\mid \theta \sim \text{PG}(1, \theta_i-\theta_j) \quad \forall \: t:(i_t, j_t)=(i,j)
\]
The expected value of $\omega_t \mid \theta$ ~\cite{polson2013bayesian}
\[
\mathbb{E}[\omega\mid\theta] = \frac{\tanh\{(\theta_i-\theta_j)/2\}}{2(\theta_i-\theta_j)} =: c_{ij}(\theta)
\]
Note that $c_{ij} = 1/4$ when $\theta_i=\theta_j$ and $c_{ij}(\theta)>0$ for any finite $\theta_i-\theta_j$. Further, $c_{ij}$ is continuous and $c_{ij} \to 0$ when $|\theta_i-\theta_j| \to \infty$. But for bounded $\theta$, $c_{ij}$ is bounded away from zero.
By assumption 5b
\[
k_{ij}(K) := |\{t \leq K:(i_t,j_t)=(i,j)\}| \geq p_{min}K \quad \text{eventually.}
\]
So $k_{ij}(K) \to \infty$ a.s. The SLLN for conditionally i.i.d. PG draws~\cite{hall2014martingale}, then gives
\begin{equation}\label{PG_draw}
    \frac{1}{k_{ij}(K)}\sum_{t:(i_t,j_t)=(i,j)} \omega_t \to c_{ij}(\theta)>0 \quad \text{a.s.}
\end{equation}
Combining ~\ref{PG_draw} with assumption 5b,
\begin{equation}
    w_{ij}(K) = k_{ij}(K) \cdot \frac{1}{k_{ij}(K)}\sum_t\omega_t \geq p_{min}K \cdot \frac{c_{ij}(\theta)}{2}  \quad a.s.
\end{equation}
for all $K$ sufficiently large. Set $w_*(K) := min_{i<j}w_{ij}(K)$. Then $w_*(K) = \Omega(K)$ a.s.\\
For any $\nu \in \mathcal{C}$
\[
\nu^\top \Lambda_{data}(\omega)\nu = \sum_{i<j}w_{ij}(K)(\nu_i-\nu_j)^2 \geq w_*(K)\sum_{i<j}(\nu_i-\nu_j)^2
\]
We know that $\nu\in\mathcal{C}$ and hence $1^\top\nu = 0$
\[
\sum_{i<j} (\nu_i - \nu_j)^2 = \frac{1}{2} \sum_{i,j} (\nu_i - \nu_j)^2 = N \|\nu\|^2 - \left( \sum_i \nu_i \right)^2 = N \|\nu\|^2
\]
Therefore,
\[
\lambda_{\min,C}\{\Lambda_{\mathrm{data}}(\omega)\} \geq N \cdot w_{*}(K) = \Omega(K) \quad \text{a.s.}
\]
Addition of prior precision $\Lambda_0 \succeq 0$ does not change the lower bound, and hence $\lambda_{\min,C}\{\Lambda^*(\omega)\} = \Omega(K) $ a.s. 

$||\Lambda^*(\omega)^{-1} ||_{\mathcal{C}} = O(K^{-1})$ is almost surely by spectral inversion. For any pair $(i,j)$, the vector $e_i-e_j$ lies in $\mathcal{C}$, so
\[
\operatorname{Var}(\theta_i - \theta_j \mid y_K, \omega) = (e_i - e_j)^{\top} \Lambda^{*} (\omega)^{-1} (e_i - e_j) \leq 2 \|\Lambda^{*}(\omega)^{-1}\|_{C} = O(K^{-1}).
\]
\\
\textbf{Definition: }The pairwise KL divergence for each pair $(i,j)$ can be defined as follows
\[
KL_{ij}(\theta) = KL\{ P_{\theta^*}^{ij} \:||\: P_{\theta}^{ij}\}
\]
where $P_{\theta}^{ij}$ is the Bernoulli distribution with the probability of success $\sigma(\theta_i-\theta_j)$. Then we can write KL divergence explicitly 
\begin{equation}\label{KLD_pairs}
    KL_{ij}(\theta) = \sigma(\theta_i^* - \theta_j^*) \log \frac{\sigma(\theta_i^* - \theta_j^*)} {\sigma(\theta_i - \theta_j)} + \sigma(\theta_j^* - \theta_i^*) \log \frac{\sigma(\theta_j^* - \theta_i^*)} {\sigma(\theta_j - \theta_i)}
\end{equation}

The population average KL: \[ KL_{\mathcal{P}}(\theta) = \frac{1}{|\mathcal{P}|} \sum_{i<j} KL_{ij}(\theta) \]
where $\mathcal{P} = \binom{N}{2}$.\\
The empirical design weighted KL: 
\begin{equation}\label{Emp_KL}
    KL_K(\theta) = \frac{1}{K} \sum_{t=1}^K KL_{i_tj_t} (\theta) = \sum_{i<j} \frac{k_{ij} (K)}{K}KL_{ij}(\theta)
\end{equation}
 where $K$ is the total number of comparisons considered.\\

\noindent \textbf{Lemma 3.} \textbf{KL coercivity and empirical separation}. \textit{ Under Assumptions 3, 5b, 6, and 7
\begin{enumerate}[label=(\roman*)]
    \item The sublevel sets \[S_M = \{ \theta \in \mathcal{C} : KL_P(\theta) \leq M \}\] are compact for every $M < \infty$.
    \item For every $\varepsilon > 0$, \[c(\varepsilon) := \inf_{\theta \in \mathcal{C}:\lVert \theta - \theta^{*} \rVert\geq\varepsilon} KL_P(\theta)>0\]
    \item For all sufficiently large K, \[\inf_{\lVert \theta - \theta^{*} \rVert \ge \varepsilon} KL_{K}(\theta) \ge c'(\varepsilon) > 0\]
    \item For every compact $B\subset \mathcal{C}$, \[\sup_{\theta \in B} \left|\frac{1}{K} \left\{\ell_{K}(\theta) - \ell_{K}(\theta^{*}) \right\} + KL_{K}(\theta) \right| \to 0\] in $P_{\theta^*}$-probability.
\end{enumerate}
}

\noindent \textit{Proof. } $\mathrm{KL}_P(\theta)$ depends on $\theta$ through $\sigma(\theta_i-\theta_j)$. Since $\sigma(\cdot)$ is a continuous function, $KL_P (\theta)$ is also continuous on $\mathcal{C}$. Therefore, the sublevel set $S_M = \mathrm{KL}_P^{-1}((-\infty, M])$ is closed, being the preimage of a closed set under a continuous function.\\

Suppose there is a sequence such that $\theta^{(n)} \in \mathcal{C}$ and $\lVert \theta^{(n)} \rVert \to \infty$. Since $\theta^{(n)} \in \mathcal{C}$, $\sum_i \theta_i^{(n)} = 0$. If all pairwise differences $\lvert\theta_i^{(n)}-\theta_j^{(n)}\rvert$ are bounded, then all coordinates must be bounded under the centred condition. But this contradicts $\lVert \theta^{(n)} \rVert \to \infty$. Hence, at least one pair $(i,j)$ diverges along a subsequence, i.e. $\lvert\theta_i^{(n)}-\theta_j^{(n)}\rvert \to \infty$. If $ \theta_i^{(n)}-\theta_j^{(n)} \to +\infty$, then $\sigma(\theta_i^{(n)}-\theta_j^{(n)}) \to 1$, but the true probability $\sigma(\theta_i^*-\theta_j^*) \in (0,1)$. According to pairwise KL equation (~\ref{KLD_pairs}) $KL_{ij}(\theta) \to \infty$ and then $\mathrm{KL}_P(\theta) \to \infty$. So, $\mathrm{KL}_P$ is coercive, and there is no sequence such that $\lVert \theta \rVert \to \infty$ in the set $\mathrm{KL}_P(\theta) \leq M$. Because of this, $S_M$ is bounded. Since $\mathcal{C}$ is finite-dimensional, the Heine--Borel theorem implies $S_M$ is compact, being closed and bounded.\\

By the information equality for a single pair $\mathrm{KL}_{ij}(\theta) = 0 $ iff $\sigma(\theta_i-\theta_j) = \sigma(\theta_i^*-\theta_j^*)$. So, $\theta_i-\theta_j = \theta_i^*-\theta_j^* $ as $\sigma(\cdot)$ is a strictly increasing function. Therefore, $\mathrm{KL}_P(\theta) = 0$ means all pairwise differences match, i.e., 
$\theta_i - \theta_j = \theta_i^* - \theta_j^*$ for all pairs. Equivalently, 
$(\theta_i - \theta_i^*) - (\theta_j - \theta_j^*) = 0$ for all $(i,j)$, so 
$\theta_i - \theta_i^*$ is the same for all $i$; that is, $\theta - \theta^* = c\mathbf{1}$ 
for some constant $c$. Since both $\theta, \theta^* \in \mathcal{C}$, we have 
$\mathbf{1}^\top(\theta - \theta^*) = cN = 0$, forcing $c = 0$. Hence $\theta = \theta^*$.\\

Define the set $A_{\varepsilon} = \{\theta \in \mathcal{C}: \lVert\theta-\theta^*\rVert \geq \varepsilon\}$. Note that $\theta^* \notin A_{\varepsilon}$ since $\lVert\theta^*-\theta^*\rVert = 0 < \varepsilon$. Consider $A_{\varepsilon}\cap S_M$ for any finite $M > 0$. Since $S_M$ is compact  (established above) and $A_{\varepsilon}$ is closed, the intersection $A_{\varepsilon}\cap S_M$ is compact. As $\mathrm{KL}_P$ is continuous, the 
Weierstrass extreme value theorem guarantees that $\mathrm{KL}_P$ reaches its 
infimum on $A_{\varepsilon}\cap S_M$ at some point $\theta_0$. Since 
$\theta_0 \in A_{\varepsilon}$, we have $\theta_0 \neq \theta^*$, so 
$\mathrm{KL}_P(\theta_0) > 0$.\\

For points outside $S_M$, i.e., $\theta \in A_{\varepsilon} \setminus S_M$, we have 
$\mathrm{KL}_P(\theta) > M$. Choosing $M > \mathrm{KL}_P(\theta_0)$, no such point can 
fall below $\theta_0$. Hence, the global infimum over $A_{\varepsilon}$ equals 
$\mathrm{KL}_P(\theta_0) > 0$, so $c(\varepsilon) = \mathrm{KL}_P(\theta_0) > 0$. \\

By Assumption 5b, $\liminf_{K\to\infty} k_{ij}(K)/K \geq p_{\min} > 0$ for every 
pair. For all sufficiently large $K$, every pair satisfies
\[
\frac{k_{ij}(K)}{K} \ge \frac{p_{\min}}{2}.
\]
By equation~\ref{Emp_KL}, $\mathrm{KL}_K$ expands with frequency weights as
\[ 
\mathrm{KL}_K(\theta) = \sum_{i<j} \frac{k_{ij}(K)}{K}\,\mathrm{KL}_{ij}(\theta).
\]
Since $\frac{k_{ij}(K)}{K} \ge \frac{p_{\min}}{2}$ and $\mathrm{KL}_{ij}(\theta) \ge 0$,
\[
\mathrm{KL}_K(\theta) \ge \frac{p_{\min}}{2} \sum_{i<j} \mathrm{KL}_{ij}(\theta) = \frac{p_{\min}|\mathcal{P}|}{2}\,\mathrm{KL}_P(\theta).
\]
By Part (ii), $\inf_{\lVert \theta - \theta^{*} \rVert\geq\varepsilon} \mathrm{KL}_P(\theta) 
= c(\varepsilon) > 0$. Taking the infimum over 
$\lVert \theta - \theta^{*} \rVert\geq\varepsilon$ on both sides,
\[
\inf_{\lVert \theta - \theta^{*} \rVert \geq \varepsilon} \mathrm{KL}_{K}(\theta) 
\;\geq\; \frac{p_{\min}\lvert \mathcal{P} \rvert}{2} 
\inf_{\lVert \theta - \theta^{*} \rVert \geq \varepsilon} \mathrm{KL}_{P}(\theta) 
\;=\; \frac{p_{\min}\lvert \mathcal{P} \rvert}{2}\, c(\varepsilon).
\]
Hence $c'(\varepsilon) := \frac{p_{\min} |\mathcal{P}|}{2}\, c(\varepsilon) > 0$.\\

Define a martingale difference for each fixed $\theta$ as follows;
\begin{equation}\label{mds}
    Z_t(\theta) = \log\frac{p_{\theta}(y_t \mid i_t, j_t)}{p_{\theta^*}(y_t \mid i_t, j_t)} + \mathrm{KL}_{i_t, j_t}(\theta)
\end{equation}

The pair $(i_t, j_t)$ is $\mathcal{F_{t-1}}$ measurable, and $y_t\mid\mathcal{F_{t-1}} \sim \mathrm{Bernoulli}(\sigma(\theta_{i_t}^*-\theta_{j_t}^*))$ according to assumption 3. The conditional expectation of the log-likelihood ratio is the negative KL.
\[
\mathbb{E}_{\theta^*}\left[ \log\frac{p_{\theta}(y_t \mid i_t, j_t)}{p_{\theta^*}(y_t \mid i_t, j_t)} \mid \mathcal{F}_{t-1}\right] = \mathrm{-KL}_{i_t, j_t}(\theta)
\]
Consequently, $\mathbb{E}_{\theta^*}[Z_t(\theta) \mid \mathcal{F}_{t-1}] = 0$ and hence, $Z_t(\theta)$ is a martingale difference sequence with respect to $\mathcal{F}_{t}$. 

Averaging over $t=1, \ldots, K$,
\[
\frac{1}{K}\sum_{t=1}^{K}Z_t(\theta) = \frac{1}{K}\left[\sum_{t=1}^{K} \log\frac{p_{\theta}(y_t \mid i_t, j_t)}{p_{\theta^*}(y_t \mid i_t, j_t)} \right] + \frac{1}{K}\sum_{t=1}^{K}\mathrm{KL}_{i_t, j_t} (\theta) = \frac{1}{K}\{\ell_K(\theta) - \ell_K(\theta^*)\} + \mathrm{KL}_K(\theta)
\]

$x_t^\top\theta = \theta_{i_t}-\theta_{j_t}$ is bounded since $\|x_t\| \leq \sqrt{2}$ for $\theta \in B$ and when $B$ is compact. As a result of this, $Z_t(\theta)$ is uniformly bounded. For each fixed $\theta$, the martingale SLLN~\cite{hall2014martingale} then gives 
\[
\frac{1}{K}\sum_{t=1}^{K}Z_t(\theta) \to 0 \quad \text{in } P_{\theta^*}\text{-probability}.
\]

The score gradient is:
\[
\nabla_\theta \log p_\theta(y_t \mid i_t, j_t) = x_t(y_t - \sigma(x_t^\top\theta))
\]
The log-likelihood is Lipschitz continuous with constant $\leq\sqrt{2}$ as $\|x_t\|\leq\sqrt{2}$ and $|(y_t - \sigma(x_t^\top\theta))| \leq 1$. On the compact set $B$, the KL is also Lipschitz continuous. Therefore, $\theta \mapsto \frac{1}{K}\sum_t Z_t(\theta)$ 
is Lipschitz with some constant $L$, uniform in $K$. \\

For any $\delta>0$, there is a finite $\delta$-net, $\{\theta^{(1)}, \ldots, \theta^{(L)}\}$ covering $B$ as $B$ is compact. Choosing a net point $\theta^{(l)}$ with $\| \theta-\theta^{(l)}\| < \delta$ for any $\theta \in B$, Lipschitz continuity gives 
\[
\left|\frac{1}{K}\sum_t Z_t(\theta)\right| 
\leq \left|\frac{1}{K}\sum_t Z_t(\theta^{(l)})\right| + L\delta
\]
By taking the supremum over $\theta \in B$,
\[
\sup_{\theta\in B}\left|\frac{1}{K}\sum_t Z_t(\theta)\right| 
\leq \max_{1\leq l\leq L}\left|\frac{1}{K}\sum_t Z_t(\theta^{(l)})\right| + L\delta
\]
Hence, by pointwise convergence, the finite maximum converges to zero as there are finitely many net points. Letting $\delta \to 0$, we will have
\[
\sup_{\theta\in B}\left|\frac{1}{K}\{\ell_K(\theta) - \ell_K(\theta^*)\} 
+ \mathrm{KL}_K(\theta)\right| \to 0
\]
in $P_{\theta^*}$-probability. \\

\noindent \textbf{Lemma 4.} \textbf{Posterior consistency}. \textit{Under assumptions 1-4, 5b, 6, and 7, for every $\varepsilon >0,$\[
\Pi(\|\theta - \theta^*\| > \varepsilon \mid y_K) \to 0
\] in $P_{\theta^*}$ probability}\\

\noindent\textit{Proof. } Let us define $A_{\varepsilon} = \{\theta \in \mathcal{C}: \| \theta - \theta^* \| \geq \varepsilon\}$. By Bayes' rule, the posterior probability is follows:
\begin{equation}\label{post_prob}
    \Pi(A_{\varepsilon} \mid y_K ) = \frac{\int_{A_\varepsilon}\exp\!\left\{\ell_K(\theta)-\ell_K(\theta^{*})\right\}\, d\Pi_0(\theta)}{\int_{\mathcal{C}}\exp\!\left\{\ell_K(\theta)-\ell_K(\theta^{*})\right\}\, d\Pi_0(\theta)}
\end{equation}

For any fixed $M<\infty$, on the compact set $A_{\varepsilon}\cap S_M$, Lemma 3 part (iii) and part (iv) give the following, respectively: 
\[1. \quad \mathrm{KL}_K(\theta) \geq c'(\varepsilon) \quad \text{on} \: A_{\varepsilon} \]
\[2. \quad \frac{1}{K} \{ \ell_K(\theta) - \ell_K(\theta^*) \} \approx -\mathrm{KL}_K(\theta) + o(1) \quad \text{uniformly } \]
By combining these results with the probability that tends to 1, we have 
\[
\sup_{\theta \in A_{\varepsilon} \cap S_M} \frac{1}{K}\{ \ell_K(\theta) - \ell_K(\theta^*)\} \leq - \frac{c'(\varepsilon)}{2}
\]
Consequently, \[ \int_{A_\varepsilon \cap S_M}\exp\!\left\{\ell_K(\theta)-\ell_K(\theta^{*})\right\}\, d\Pi_0(\theta) \leq \exp \{- K c'(\varepsilon)/2 \}\]

Now consider a small ball $B_r(\theta^*) \subset \mathcal{C}$. Since our Gaussian prior has full support, $\Pi_0(B_r(\theta^*)) >0$. According to the result of part iv of Lemma 3, for $r$ small enough, $\frac{1}{K}\{\ell_K(\theta) - \ell_K(\theta^*)\} \geq -c'(\varepsilon)/4$ on $B_r(\theta^*)$ with probability tending to 1. Hence, \[
\int_{\mathcal{C}}\exp\{\ell_K(\theta) - \ell_K(\theta^*)\}\,d\Pi_0 
\geq \Pi_0(B_r(\theta^*))\exp\{-Kc'(\varepsilon)/4\}
\]
By combining these bounds, the posterior probability in equation ~\ref{post_prob} satisfies
\[
\Pi(A_\varepsilon \cap S_M \mid y_K) \leq \Pi_0(B_r(\theta^*))^{-1}\exp\{-Kc'(\varepsilon)/4\} \to 0
\]

According to part i of Lemma 3, the KL sublevel sets $S_M$ compact and exhaust $\mathcal{C}$ as $M \to \infty$. Further, the proper Gaussian prior has full support, and KL is large outside $S_M$ by coercivity. As a result, the likelihood ratio is exponentially suppressed. Therefore, by the standard finite dimensional Schwartz sieve argument~\cite{ghosh2003bayesian}, $\Pi(A_\varepsilon \setminus S_M \mid y_K) \to 0$ for $M$ large. By choosing $M$ large and then letting $K \to \infty$, we have \[
\Pi(A_\varepsilon \mid y_K) = \Pi(A_\varepsilon \cap S_M \mid y_K) 
+ \Pi(A_\varepsilon \setminus S_M \mid y_K) \to 0
\] in $P_{\theta^*}$ probability. \\

\noindent \textbf{Lemma 5.} \textbf{Fisher information on $\mathcal{C}$.} \textit{For one comparison of pair $(i,j)$, the Fisher information contribution is \[
I_{ij}(\theta) = \sigma(\theta_i-\theta_j)\{ 1-\sigma(\theta_i-\theta_j) \}(e_i-e_j)(e_i-e_j)^\top.
\]
The cumulative Fisher information is 
\[
\mathcal{J}_K(\theta^*) = \sum_{t=1}^K \sigma(x_t^\top\theta^*)\{ 1 - \sigma(x_t^\top\theta^*)\}x_t x_t^\top
\]
Under assumption 5c, $\frac{1}{K}\mathcal{J}_K(\theta^*) \to \bar{I}(\theta^*) \succ 0$ on $\mathcal{C}$.
}\\

\noindent\textit{Proof. } The log likelihood for a single comparison is \[
\ell_{ij}(\theta) = y(\theta_i-\theta_j) - \log\{ 1+\exp(\theta_i-\theta_j)\}
\]
Let $\psi = \theta_i-\theta_j$, then the first and second derivatives of the log-likelihood are $\frac{\partial\ell_{ij}}{\partial\psi} = y - \sigma(\psi)$ and $\frac{\partial^2\ell_{ij}}{\partial\psi^2} = -\sigma(\psi)\{1-\sigma(\psi)\}$ respectively.\\

The Fisher information is the negative Hessian as the second derivative does not depend on $y$. Since $\psi = (e_i-e_j)^\top \theta$, mapping to $\theta$ space gives
\[
I_{ij}(\theta) = \sigma(\theta_i-\theta_j)\{ 1-\sigma(\theta_i-\theta_j) \}(e_i-e_j)(e_i-e_j)^\top
\]
Summing these contributions along the realised design, the cumulative Fisher information is \[
\mathcal{J}_K(\theta^*) = \sum_{t=1}^K \sigma(x_t^\top\theta^*)\{1 - \sigma(x_t^\top\theta^*)\}x_tx_t^\top
\]
Under assumption 5c, the convergence and positive definiteness of the normalised information matrix,
\[\frac{1}{K}\mathcal{J}_K(\theta^*) \to \bar{I}(\theta^*) \succ 0 \quad \text{on} \quad \mathcal{C}\]
are exactly the assumption itself.\\

\noindent \textbf{Lemma 6.} \textbf{Adaptive local asymptotic normality (LAN).} \textit{Under assumptions 1-7, for every fixed $h \in \mathcal{C}$
\[
\ell_K\left(\theta^* + \frac{h}{\sqrt{K}}\right) - \ell_K(\theta^*) = h^\top\Delta_K - \frac{1}{2}h^\top\bar{I}(\theta^*)h + o_{P_{\theta^*}}(1)
\] where $\Delta_K = \frac{1}{\sqrt{K}} \sum_{t=1}^K x_t \{y_t - \sigma(x_t^\top\theta^*)\}$. Moreover, \[
\Delta_K \Rightarrow \mathcal{N}_{\mathcal{C}}(0, \bar{I}(\theta^*))
\]}\\

\noindent\textit{Proof. } For a fixed $h \in \mathcal{C}$, a Taylor expansion around $\theta^*$ gives:
\[\ell_K\left(\theta^* + \frac{h}{\sqrt{K}}\right) - \ell_K(\theta^*) = \frac{1}{\sqrt{K}}h^\top\nabla\ell_K(\theta^*) + \frac{1}{2K}h^\top\nabla^2\ell_K(\tilde\theta_K)h \]
where $\tilde\theta_K$ lies between $\theta^*$ and $\theta^* + h/\sqrt{K}$.\\

Define the score term $\Delta_K = \frac{1}{\sqrt{K}}\nabla\ell_K(\theta^*) = \frac{1}{\sqrt{K}}\sum_{t=1}^{K}x_t\{y_t - \sigma(x_t^\top\theta^*)\}$. Further, the linear term is $h^\top\Delta_K$.\\

The Hessian is \[
\nabla^2\ell_K(\theta) = -\sum_{t=1}^{K}\sigma(x_t^\top\theta)\{1 - \sigma(x_t^\top\theta)\}x_t x_t^\top
\]
Since the second derivative of the logistic function is continuous and $\theta^* + h/\sqrt{K} \to \theta^*$, assumption 5c gives \[
-\frac{1}{K}\nabla^2\ell_K(\tilde\theta_K) \to \bar{I}(\theta^*) \quad \text{on } \mathcal{C}
\]. Consequently, the quadratic term converges to $-\frac{1}{2}h^\top\bar{I}(\theta^*)h$. Combining these results, \[
\ell_K\left(\theta^* + \frac{h}{\sqrt{K}}\right) - \ell_K(\theta^*) 
= h^\top\Delta_K - \frac{1}{2}h^\top\bar{I}(\theta^*)h + o_{P_{\theta^*}}(1)
\]\\

The summands $\frac{1}{\sqrt{K}}x_t\{y_t - \sigma(x_t^\top\theta^*)\}$ are martingale differences with respect to $\{\mathcal{F}_t\}$. According to Assumption 3, the pair $(i_t, j_t)$ is predictable and $y_t - \sigma(x_t^\top\theta^*)$ has mean zero given 
$\mathcal{F}_{t-1}$. Since $\|x_t\| \leq \sqrt{2}$ and $|y_t - \sigma(\cdot)| \leq 1$, 
the summands are uniformly bounded. The predictable quadratic variation is
\[
\frac{1}{K}\sum_{t=1}^{K}x_t x_t^\top\,\sigma(x_t^\top\theta^*)\{1 - \sigma(x_t^\top\theta^*)\} 
\to \bar{I}(\theta^*)
\]
by assumption 5c, and the Lindeberg condition~\cite{durrett2019probability} is automatic from boundedness. Therefore, by the martingale central limit theorem,
\[
\Delta_K \Rightarrow \mathcal{N}_{\mathcal{C}}(0, \bar{I}(\theta^*)).
\]
This proves the adaptive LAN.\\

\noindent \textbf{Lemma 7.} \textbf{Posterior local tightness.} \textit{Under assumptions 1-7, the posterior is locally tight at the $\frac{1}{\sqrt{K}}$ scale: \[ \sqrt{K}(\theta-\theta^*) \mid y_K = O_{P_{\theta^*}}(1)\] in the posterior sense. Equivalently, for every $\eta > 0$, there exists $M<\infty$ s.t. \[
\limsup_{K\to \infty} P_{\theta^*}\left[ \Pi\{ \sqrt{K} \|\theta-\theta^*\| >M \mid y_K\} > \eta \right] < \eta.
\]}

\noindent\textit{Proof. } According to Lemma 4, the posterior is consistent and hence for any small $r>0$ \[
\Pi\{ \|\theta-\theta^*\| > r \mid y_K\} \to 0 \quad \text{in} \; P_{\theta^*} \text{-probability}
\]
So, it is sufficient to control the posterior inside a local ball $B_r(\theta^*) = \{\theta \in \mathcal{C} : \|\theta-\theta^*\| \leq r \}$. 
By considering $u = \theta - \theta^*$, define the score as $S_K = \nabla\ell_K(\theta^*) = \sum_{t=1}^K x_t\{y_t-\sigma(x_t^\top \theta^*)\}$. Under assumption 3, $S_K$ is a martingale sum and the predictable quadratic variation is of order $K$. Hence, \[ \frac{1}{\sqrt{K}} S_K = O_{P_{\theta^*}}(1)\].

By Taylor expansion for $\theta = \theta^* + u \in B_r(\theta^*)$, we have \[
\ell_K(\theta^*+u) - \ell_K(\theta^*) = u^\top S_K - \frac{1}{2}u^\top\mathcal{J}_K(\tilde{\theta})u
\]
where $\tilde{\theta}$ lies between $\theta^*$ and $\theta^*+u$, and $\mathcal{J}_K(\theta) = \sum_t \sigma(x_t^\top \theta)\{1-\sigma(x_t^\top \theta)\} x_t x_t^\top$. Further $\frac{1}{K}\mathcal{J}_K(\theta^*) \to \bar{I}(\theta^*) \succ 0$ under assumption 5c. For small $r$, the same positive curvature holds uniformly on $B_r(\theta^*)$, as the logistic weight is continuous. With probability tending to one, there is a constant $c>0$ \[
u^\top \mathcal{J}_K(\tilde{\theta})u \geq c K \| u\|^2, \quad \forall \; \| u\| \leq r 
\]  
On this high probability event, \[
\ell_K(\theta^* + u) - \ell_K(\theta^*) \leq u^\top S_K - \frac{cK}{2}\|u\|^2
\]

Setting $h = \sqrt{K}u$ and $ \Delta_K = K^{-1/2}S_K $, we will have \[
\ell_K(\theta^* + h/\sqrt{K}) - \ell_K(\theta^*) \leq h^\top\Delta_K - \frac{c}{2}\|h\|^2
\]

Since $\Delta_K = O_{P_{\theta^*}}(1)$, the negative quadratic term dominates the 
linear term for large $\|h\|$. The prior density is continuous and bounded near 
$\theta^*$. Therefore, after the change of variables $h = \sqrt{K}(\theta - \theta^*)$, 
the posterior mass of $\{\|h\| > M\}$ is bounded by a Gaussian-type tail
\[
\int_{\|h\| > M}\exp\left\{h^\top\Delta_K - \frac{c}{2}\|h\|^2\right\}dh,
\]
up to a factor bounded in probability. This tail can be made arbitrarily small for 
large $M$. By combining this local bound with consistency, the local posterior 
tightness follows,
\[
K^{1/2}(\theta - \theta^*) \mid y_K = O_{P_{\theta^*}}(1)
\]

\noindent \textbf{Lemma 8.}\textbf{Adaptive Bernstein von Mises theorem} \textit{Under assumptions 1-7, \[
\sup_{B\in\mathcal{B}(\mathcal{C})}|\Pi\{ \sqrt{K}(\theta-\hat{\theta}_K)\in B \mid y_K\} - \mathcal{N}_{\mathcal{C}}\big(0, \bar{I}(\theta^*)^{-1} \big)(B)| \to 0
\] in $P_{\theta^*}$-probability, where $\hat{\theta}_K$ is the MLE on $\mathcal{C}$. Consequently, \[
\sqrt{K}(\theta-\hat{\theta}_K)\mid y_K = O_{P_{\theta^*}}(1)
\] in the posterior tightness sense.}\\

\noindent\textit{Proof. } The posterior is consistent at $\theta^*$ according to Lemma 4, and the likelihood satisfies LAN with limiting information $\bar{I}(\theta^*) \succ 0$ from Lemma 6. Lemma 7 ensures the posterior mass concentrates on bounded sets $\{\| h \| \leq M\}$ by rescaling $h = \sqrt{K}(\theta-\theta^*)$. Hence, it is enough to analyse the posterior on bounded local sets, with $M$ allowed to increase afterwards.\\

We have shown that the LAN expansion is uniform on a bounded local set in Lemma 6. Since the prior density is continuous and positive near $\theta^*$, for bounded $h$\[
\frac{\pi_0(\theta^*+h/\sqrt{K})}{\pi_0(\theta^*)} \to 1
\]
As a result of this, the posterior density in $h$ is asymptotically proportional to \[
\exp\left\{h^\top\Delta_K - \frac{1}{2}h^\top \bar{I}(\theta^*)h\right\}
\] on bounded sets.\\

A Gaussian local limit with covariance $\bar{I}(\theta^*)^{-1}$ and centre $\bar{I}(\theta^*)^{-1}\Delta_K$ is given by completing the square. Since the MLE satisfies \[
\sqrt{K}(\hat\theta_K - \theta^*) = \bar{I}(\theta^*)^{-1}\Delta_K + o_{P_{\theta^*}}(1)
\], recentring the local posterior at $\hat\theta_K$ gives $\mathcal{N}_{\mathcal{C}}(0, \bar{I}(\theta^*)^{-1})$.\\

Because of the local tightness, the mass outside of the bounded local sets is negligible. Consequently, the Gaussian approximation extends from bounded sets to total variation over $\mathcal{C}$. This is the LAN-to-BvM argument of~\cite{van2000asymptotic}, with the i.i.d. score CLT replaced here by the martingale CLT of Lemma 6. Therefore, \[
\sup_{B \in \mathcal{B}(\mathcal{C})}\left|\Pi\{\sqrt{K}(\theta - \hat\theta_K) \in B \mid  y_K\} - \mathcal{N}_{\mathcal{C}}(0, \bar{I}(\theta^*)^{-1})(B)\right| \to 0
\] in $P_{\theta^*}$-probability. 

\subsection{Main theorem}
\noindent\textbf{Theorem }\textbf{Adaptive posterior consistency and posterior convergence rate.} Assume assumptions 1-4, 5b, 6, and 7. Then for every $\varepsilon>0,$ \[
\Pi\left(\|\theta-\theta^*\|>\varepsilon \mid y_K\right) \to 0
\] in $P_{\theta^*}$-probability.\\
If, additionally, assumption 5c holds, then the marginal posterior satisfies BvM convergence \[
\sup_{B \in \mathcal{B}(\mathcal{C})}\left|\Pi\{\sqrt{K}(\theta - \hat\theta_K) \in B \mid  y_K\} - \mathcal{N}_{\mathcal{C}}(0, \bar{I}(\theta^*)^{-1})(B)\right| \to 0
\]
in $P_{\theta^*}$-probability. Hence, the posterior contracts around $\hat{\theta}_K$ at the rate of $\frac{1}{\sqrt{K}}$.\\

\noindent\textit{Proof. } Fix $\varepsilon>0$. The selected pair $(i_t, j_t)$ is predictable, and score increments are martingale differences under assumption 3. Further, assumption 5b ensures that every pair is sampled with positive long-run frequency. So, Lemma 3 then gives the following:
\[1. \quad \inf_{\|\theta-\theta^*\|\geq\varepsilon} \mathrm{KL}_K(\theta)\geq c'(\varepsilon) >0 \; \text{for all sufficiently large} \; K.\]
\[2. \quad \text{For every compact} \; B\subset \mathcal{C}, \; \sup_{\theta \in B} \left|\frac{1}{K} \left\{\ell_{K}(\theta) - \ell_{K}(\theta^{*}) \right\} + \mathrm{KL}_{K}(\theta) \right| \to 0\ \]
Consequently, the likelihood ratio is exponentially suppressed away from $\theta^*$. Lemma 4 ensures \[
\Pi(\|\theta-\theta^*\| > \varepsilon\mid y_K) \to 0
\] in $P_{\theta^*}$-probability as the proper Gaussian prior gives positive mass near $\theta^*$. Therefore, the posterior consistency is achieved.\\
If, additionally, assumption 5c holds, the normalised Fisher information is $\frac{1}{K} \mathcal{J}_K(\theta^*) \to \bar{I}(\theta^*) \succ 0$. By Lemma 6, the adaptive likelihood satisfies the LAN expansion,\[
\ell_K\left(\theta^* + \frac{h}{\sqrt{K}}\right) - \ell_K(\theta^*) = h^\top\Delta_K - \frac{1}{2}h^\top\bar{I}(\theta^*)h + o_{P_{\theta^*}}(1)
\], with $\Delta_K\Rightarrow \mathcal{N}_{\mathcal{C}}(0, \bar{I}(\theta^*))$. Combining LAN, Posterior consistency, prior positivity, and local tightness, Lemma 8 provides BvM convergence, \[
\sup_{B \in \mathcal{B}(\mathcal{C})}\left|\Pi\{\sqrt{K}(\theta - \hat\theta_K) \in B \mid  y_K\} - \mathcal{N}_{\mathcal{C}}(0, \bar{I}(\theta^*)^{-1})(B)\right| \to 0.
\]
Because the limiting Gaussian is tight, for every sequence $M_K \to \infty$,
\[
\Pi\left(\|\theta - \hat\theta_K\| > \frac{M_K}{\sqrt{K}} \,\Big|\, y_K\right) \to 0.
\]
Thus the posterior contracts around $\hat{\theta}_K$ at rate $\frac{1}{\sqrt{K}}$.

\subsection{Corollary}
\noindent\textbf{Corollary 1. }\textbf{Posterior and posterior mean ranking consistency}. Assume assumptions 1-4, 5b, 6, and 7. Let $\pi(\theta)$ be the ranking induced by a posterior draw $\theta$. Then \[
\Pi\{\pi(\theta) = \pi^* \mid y_K\} \to 1
\] in $P_{\theta^*}$-probability.
If, additionally, assumption 5c holds and the main theorem applies, then the posterior mean ranking \[
\hat{\pi} = \pi(m_K), \quad m_K = \mathbb{E}(\theta \mid y_K)
\] is frequentist consistent:\[
P_{\theta^*}(\hat{\pi} = \pi^*) \to 1
\]

\noindent\textit{Proof. } By assumption 2, $\delta_{\min} =\min_{i<j}|\theta_i^*-\theta_j^*| > 0$. Choose $\eta \in (0, \delta_{min}/2)$. If $\|\theta - \theta^*\|_{\infty} < \eta$, then for every pair $(i,j)$ \[
|(\theta_i-\theta_j)-(\theta_i^*-\theta_j^*)| \leq |\theta_i - \theta_i^*|\;+\; |\theta_j-\theta_j^*| <2\eta < \delta_{min}
\]
Therefore every pairwise sign is correct, so $\pi(\theta) = \pi^*$. Hence,\[
\Pi\{\pi(\theta) = \pi^* \mid y_K\} \geq \Pi\{\|\theta - \theta^*\|_{\infty} < \eta \mid y_K\}
\]
Since $\mathcal{C}$ is finite-dimensional, all norms are equivalent. Because of that, the Euclidean consistency of the main theorem extends to $\ell_{\infty}$ consistency. Therefore $\Pi\{\|\theta - \theta^*\|_\infty < \eta \mid y_K\} \to 1$, and combined with the inequality above \[
\Pi\{\pi(\theta) = \pi^* \mid y_K\} \to 1
\] and which proves the posterior ranking consistency.\\
For the posterior-mean ranking, the main theorem gives
\[
m_K = \hat\theta_K + o_{P_{\theta^*}}(K^{-1/2}),
\]
and $\hat\theta_K \to \theta^*$ in $P_{\theta^*}$-probability. Hence 
$m_K \to \theta^*$ in $P_{\theta^*}$-probability, so
\[
P_{\theta^*}\{\|m_K - \theta^*\|_\infty < \eta\} \to 1.
\]
On this event, $m_K$ has the same ranking as $\theta^*$. Therefore,
\[
P_{\theta^*}(\hat\pi_K = \pi^*) \to 1.
\]

\newpage

\section{Experiments and Discussions}\label{Exp_Dis}
\subsection{Experimental Setup}
We tested our framework on synthetic ranking problems. Since we know the ground-truth ranking order, we were able to measure the recovery accuracy directly. Each problem starts with $N$ item abilities, drawn from truncated normal distributions with a between-item spread of $\tau = 15$. We chose a truncated normal instead of a uniform draw because real abilities gather around the population mean. We used Thurstone's model~\cite{thurstone1927law} to generate the comparison outcomes: the abilities of two items are independently sampled from two Gaussian distributions with noise (standard deviation $\sigma$), and the item with the higher sample value becomes the winner of that comparison. Here $\sigma$ acts as within-item measurement noise, while $\tau$ decides 
how hard the problem is to recover the correct ranking.\\

We simulated the experiment with different numbers of items, $N = \{5, 10, 15, 20, 25\}$, and different numbers of comparisons per item, $K = \{5, 10, 20, 30\}$. The total number of comparisons per experiment was therefore $N \times K$, which forms the total budget. This is the commonly used budget allocation mechanism in CJ research; according to \citep{jones2024comparative}, $K = 10$ is the default convention for CJ experiments. We additionally controlled the within item noise, $\sigma = \{0.0001, 1.25, 2.5, 5.0, 10.0\}$, to bring the experiment closer to a real-world scenario. We start from a near-deterministic regime where the stronger item always wins, and increase the noise up to the between item 
spread $\tau$, which creates uncertainty about the outcome, i.e., on some days a weaker 
student might perform better than a stronger one.\\

We compare our BBTM with the existing BCJ model of ~\citep{gray2024bayesian}, since both are Bayesian approaches to CJ and therefore directly comparable. Each model is tested with six 
pair selection methods, $S = \{\text{random, no-repeat, entropy, KG-Tau, KG-Entropy, KG- SSR}\}$. The knowledge-gradient methods KG-Tau, KG-Entropy, and KG-SSR are newly designed 
active learning rules, discussed in Section 3.3. Every combination was run 50 times from scratch, and the results were recorded. Within a seed, all methods saw the same generated problems, so their results form matched pairs. We then tested the significance of each model and method with the Wilcoxon signed-rank test, adjusted with a Bonferroni correction across the models and methods at $\alpha = 0.05$. Accuracy is recorded as the normalised Kendall's tau distance from the true ranking.\\

We implemented Thurstone's model~\cite{thurstone1927law} for outcome generation to compare the performance of BBTM and PBCJ. However, to prove the convergence of BBTM in accordance with Assumption 3, we used a Bradley Terry like outcome generator based on the sigmoid function. Like Thurstone's model, this is also a sampling based generator, but the win probability comes from the sigmoid rather than from comparing two Gaussian draws. The win probability is obtained by passing the ability difference through the sigmoid function. We then draw a random number between 0 and 1 and compare it with the win probability: if the win probability exceeds the random number, the first item wins; otherwise the second item wins.\\

We then checked the convergence of BBTM for different $N$, $K$, and $\sigma$ under random and non-repeating pair selection, since these selectors already satisfy Assumptions 5a and 5b. Greedy pair-selection methods such as entropy and KG do not always satisfy Assumption 5b. The BBTM convergence theorem may therefore fail to apply to these greedy methods. To address this, we incorporated an exploration phase into the greedy methods and checked whether convergence is restored.\\

\subsection{Model convergence check}
\label{model_conv}
\begin{figure}[H]
    \centering
    \includegraphics[width=\textwidth]{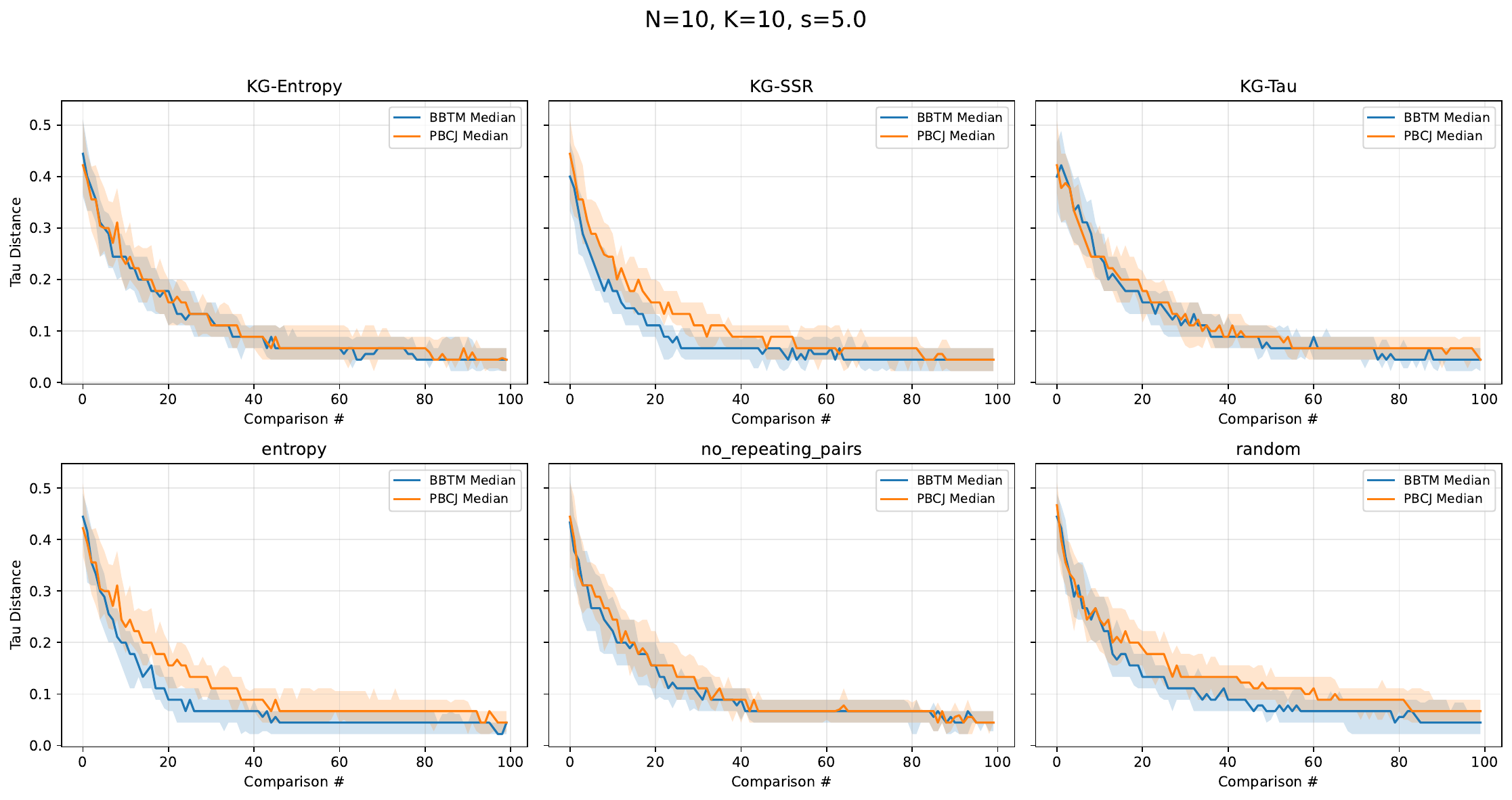}
    \caption{Convergence of BBTM and PBCJ across six pair selections, for $\sigma=5.0$, $N=10$, $K=10$. Each subplot shows the median tau distance across $50$ seeds against the number of comparisons. The shaded region indicates the interquartile range. Results for $\sigma=1.25$ and $\sigma=10.0$ are given in Appendix~\ref{app:conv_plots}.}
    \label{fig:conv_sigma_5}
\end{figure}

The convergence behaviour of BBTM and PBCJ can be seen in Figure~\ref{fig:conv_sigma_5} for $\sigma=5.0$, $N=10$ items, $K=10$, and each configuration was run 50 times. Both models converge as the number of comparisons increases. BBTM converges faster than PBCJ under KG-SSR, KG-Tau, entropy, and random selection, while the two models perform comparably under KG-Entropy and no-repeating-pairs. At lower noise level ($\sigma=1.25$; Appendix~\ref{app:conv_plots}), BBTM converges faster across all selection methods. However, at higher noise level ($\sigma=10.0$; Appendix~\ref{app:conv_plots}), BBTM converges faster under KG-SSR, entropy, and random selection while the other three selection methods comparable. We believe this faster convergence of BBTM arises because of its hierarchical structure. This structure enables information sharing across items through borrowing strength.

\subsection{Identifying the winning method}
\label{win_methd}
We evaluate both methods and six different pair selections on the same 50 seeds with different $N, K$ \& $\sigma$. Then we consider the final normalised Kendall's Tau value for each seed in each combination, and we use these 50 tau values to determine which method is best. We used the same 50 seeds when comparing two combinations, and therefore, those tau values behave as matched pairs. Because of this, we used the Wilcoxon Signed Rank test to exploit these pairings and this is the appropriate paired test for matched observations.\\

Consider the difference of each seed $s$ as $d_s = \tau_s^{(x)}-\tau_s^{(y)}$ for two combinations $(x,y)$. The signed rank test then ranks the $|d_s|$ values, takes the sum of the signed ranks, and gives a one-tailed $p$-value. This tests the hypothesis that combination $x$ is worse than combination $y$ using the tau value, i.e., a higher tau value combination is worse as we consider normalised Kendall's tau distance. We then count how many combinations each one is significantly worse than: \[
V(i) = \sum_{x \neq y}1\left[ p\text{-value}(x > y) \leq \alpha_{\text{adj}} \right]
\]   
Here $\alpha_{\text{adj}} = \alpha / (m-1)$ is the Bonferroni adjusted level, $m$ is the number of combinations compared in the cell, and $\alpha = 0.05$. A combination with $V(i) = 0$ is one that no other combination significantly outperforms, marking it the best for that configuration. Sweeping this count across the full $(N, K, \sigma)$ grid shows how the winning method shifts with problem size, budget, and noise.\\

\begin{figure}[H]
    \centering
    \includegraphics[width=\textwidth]{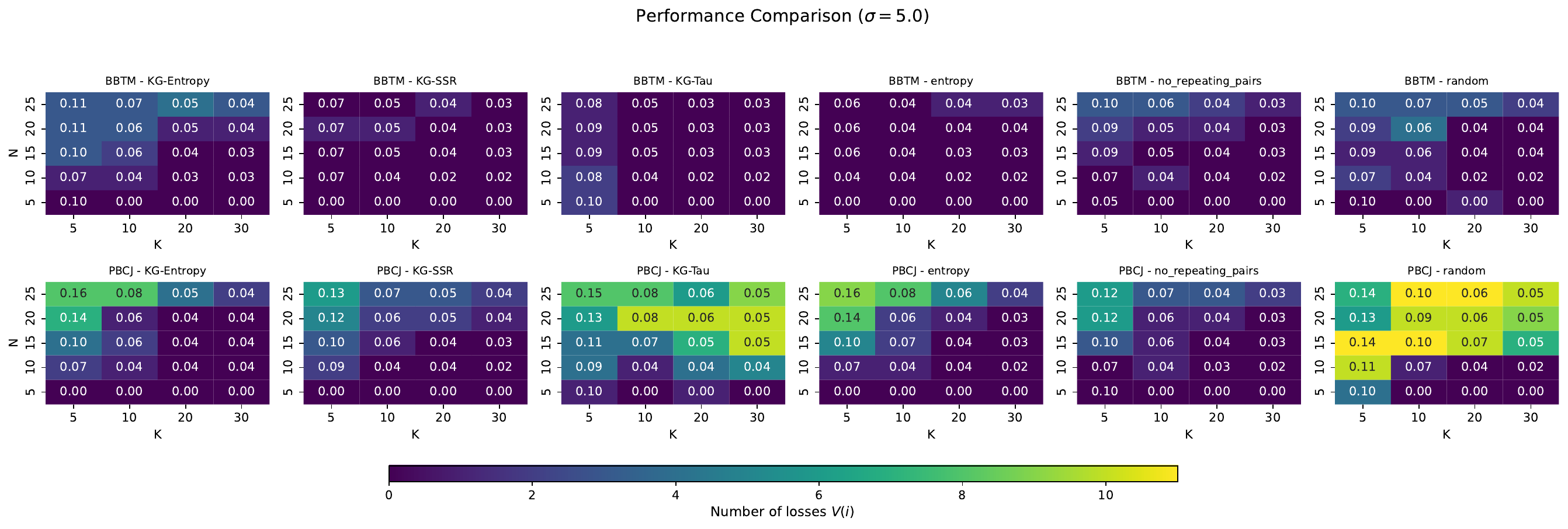}
    \caption{Performance comparison at $\sigma = 5.0$. Each panel corresponds to one combination of a ranking model (BBTM, top row; PBCJ, bottom row) and a pair-selection method (columns), evaluated over the $N \times K$ grid. The colour scale shows the number of losses $V(i)$, that is, the number of competing combinations that significantly outperform the given one under the Wilcoxon signed-rank test, with the darkest cells indicating a combination not beaten by any other ($V(i) = 0$). The number in white indicates the median final normalised Kendall's $\tau$ distance over 50 repeats for that configuration, with lower values indicating better ranking recovery. Heatmaps for the remaining noise levels are provided in Appendix~\ref{app:heatmap-results}.}
    \label{fig:heatmap-sigma-50}
\end{figure}

We first compare the overall performance of BBTM and PBCJ. Figure~\ref{fig:heatmap-sigma-50} shows a representative pattern for $\sigma = 5.0$; heatmaps for the remaining noise levels are given in Appendix~\ref{app:heatmap-results}. BBTM clearly has fewer losses than PBCJ across all the noise levels. Both models show the same $N \times K$ gradient, i.e., hardest at the high $N$ and low $K$ (top left) while easiest at low $N$ and high $K$ (bottom right). Note that this gradient is consistent for all the noise levels (Appendix~\ref{app:heatmap-results}). However, we will focus only on BBTM in subsequent analysis as it performs better than PBCJ at every noise level.\\

We now identify which pair selection strategy perform best under BBTM. We focus on the results at $\sigma=5.0$, which is closer to a real world scenario. We consider following measures to reach a decision: The mean $\tau$ distance, the number of cells with $V(i) = 0$, and the total number of losses. According to these measures, entropy pair selection under BBTM outperforms all other pair selection strategies. KG-SSR follows closely, then KG-Tau. These validate the active learning premise: informed selection driven by posterior uncertainty clearly outperforms uninformed random selection. Interestingly, although plain maximum entropy selection is the best strategy, its KG variant (KG-Entropy) is the worst; a likely explanation is that the one-step look-ahead over exploits the entropy signal, whereas plain maximum entropy selection explores more effectively. Whether this ordering remains stable as noise grows is the question we turn to next.\\

The preceding analysis was based on $\sigma=5.0$. We now analyse how the selection strategies behave as noise grows, from $\sigma=0.0001$ (near deterministic) to $\sigma=10.0$ (high noise). It can be seen that for low and moderate noise levels $\sigma \leq 5.0$, entropy pair selection consistently outperforms all other pair selection strategies. However, as the noise increases to $\sigma=10.0$, the picture shifts. Here KG-SSR overtakes entropy on all metrics discussed above. Note that entropy's lead narrows steadily with noise, until the crossover occurs at $\sigma=10.0$. This shift is understandable. At low noise levels, comparison outcomes are informative, and entropy's uncertainty sampling is efficient because it selects genuinely uncertain pairs. At higher noise levels, entropy might select uninformative pairs because outcomes are unreliable, and it chases noise inflated uncertainty. But on the other hand, KG-SSR focuses on posterior separation, specifically targeting the pairs that improve the rank separability. Therefore, KG-SSR is more robust in these scenarios as it is driven by the signal-to-noise ratio rather than by pure uncertainty. This makes it important to sweep the noise levels rather than fix a single one, since the best strategy depends on the difficulty of the problem.\\     

\subsection{Empirical validation of the Convergence Theorem}
In order to validate the BBTM convergence theorem, the comparison outcomes must follow the likelihood assumed by the model (Assumption 3). Hence, we generate the comparison outcomes using a Bradley-Terry generator for this analysis instead of the Thurstone-style generator. Since we assume the same level of uncertainty $\sigma$ at the item level, this matches Thurstone's case V~\citep{thurstone1927law}. According to Thurstone's case V, the win probability depends on the scale difference divided by the uncertainty, i.e., $(\theta_i - \theta_j)/(\sqrt{2}\,\sigma)$. This uses a normal (probit) link, but we need its logistic counterpart to preserve the Bradley-Terry likelihood: \[
P(i \succ j) = \mathrm{logit}^{-1}\!\left(\frac{\theta_i - \theta_j}{\sigma}\right)
\]     
The normal and logistic links are nearly identical up to a linear scaling, which does not affect the model fit~\citep{camilli1994teacher}. The constant is therefore absorbed into $\sigma$, which retains its interpretation as the discriminable dispersion. A small $\sigma$ yields near deterministic outcomes while a large $\sigma$ drives win probability towards $0.5$, producing noisy outcomes.\\

Every pair must be sampled infinitely often to satisfy Assumption 5b. Some pairs can be permanently starved under pure greedy pair selections such as KG and entropy, since these select the most informative pairs at each step. This violates our assumption 5b. We introduce $\varepsilon$-greedy exploration to address this issue. A pair is chosen uniformly at random with probability $\varepsilon$ (exploration), and with probability $(1- \varepsilon)$ the most informative pair is chosen (exploitation) at each comparison. This ensures that every pair is eventually sampled, satisfying assumption 5b. The selection is purely greedy when $\varepsilon = 0$, and exploration increases as $\varepsilon$ increases.\\

To validate convergence behaviour, we perform a simulation study with $N \in {10, 15}$ items, $K \in {5,10,20,30}$ comparisons per item, and $\sigma \in {0.0001,1.25,2.5,5.0,10.0}$. Further, we consider these exploration levels $\varepsilon \in {0.0,0.05,0.10,0.20}$ and run each configuration over 50 seeds. We focus only on the greedy pair selectors (KG-Entropy, KG-Tau, KG-SSR, entropy) as only these starve pairs and violate assumption 5b. We measure convergence through normalised Kendall’s tau distance and pair coverage through the fraction of unique pairs sampled.\\

\begin{figure}[H]
    \centering
    \includegraphics[width=\textwidth, height=12cm, keepaspectratio]{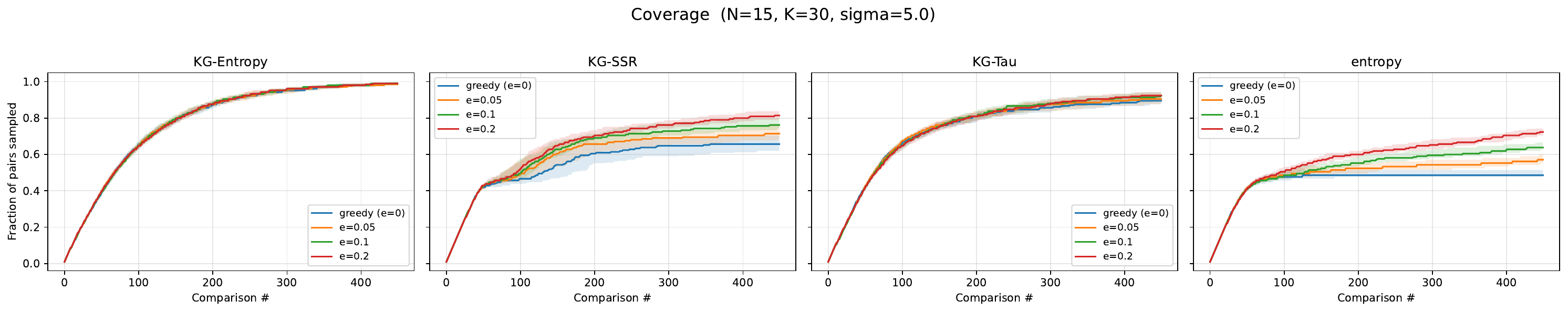}
    \caption{Pair coverage, the fraction of unique pairs sampled, against the number of comparisons, for four greedy selectors ($N=15$, $K=30$, $\sigma=5.0$, $50$ seeds). Each panel shows the coverage across exploration levels $\varepsilon \in \{0, 0.05, 0.10, 0.20\}$. Coverage for the remaining noise levels ($\sigma \in \{0.0001, 1.25, 2.5, 10.0\}$) is given in Appendix~\ref{app:convergence-results}.}
    \label{fig:coverage-5.0}
\end{figure}

The fraction of unique pairs sampled against the number of comparisons across exploration levels is shown in Figure~\ref{fig:coverage-5.0}. The pair coverage of entropy and KG-SSR are 49\% and 66\% respectively under purely greedy selection $(\varepsilon = 0.0)$. This indicates entropy and KG-SSR starve a considerable number of pairs and violate assumption 5b. With $\varepsilon$-greedy exploration, the coverage of entropy increases to 72\% and KG-SSR coverage increases to 81\% at $\varepsilon = 0.20$. In contrast, KG-Entropy and KG-Tau satisfy assumption 5b even without exploration as they cover 99\% of pairs at $\varepsilon = 0.0$.\\

\begin{figure}[H]
    \centering
    \includegraphics[width=\textwidth, height=10cm, keepaspectratio]{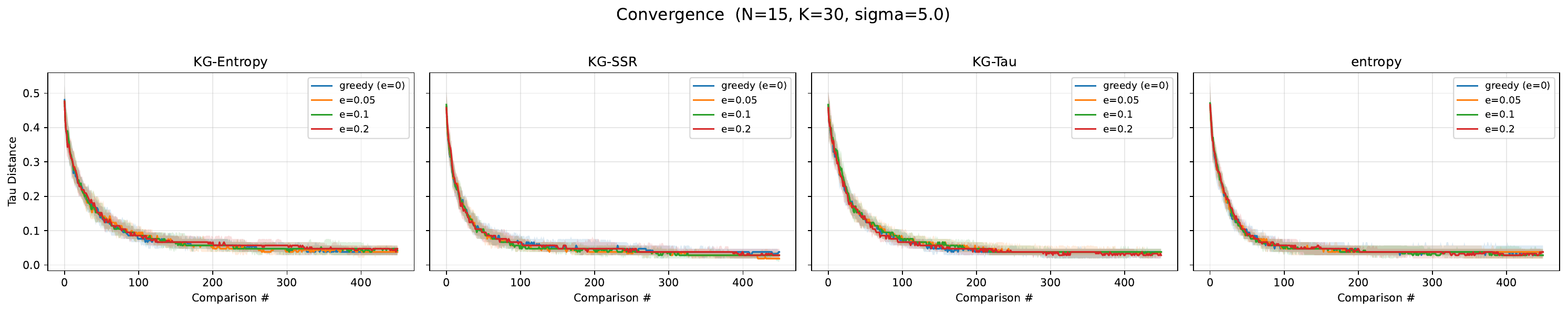}
    \caption{Convergence plots, Normalised Kendall's $\tau$ distance against the number of comparisons, for four greedy selectors ($N=15$, $K=30$, $\sigma=5.0$, $50$ seeds). Each panel shows the median $\tau$ distance across exploration levels $\varepsilon \in \{0, 0.05, 0.10, 0.20\}$, with shaded bands indicating the interquartile range. Convergence for the remaining noise levels ($\sigma \in \{0.0001, 1.25, 2.5, 10.0\}$) is given in Appendix~\ref{app:convergence-results}.}
    \label{fig:convergence-5.0}
\end{figure}

Figure~\ref{fig:convergence-5.0} illustrates the tau distance against comparisons. Pure greedy selections starve pairs but they still converge on these problems. The final tau distance of the greedy selections $(\varepsilon = 0.0)$ is comparable to, or slightly better than, that of $\varepsilon$-greedy. This indicates that the sampled pairs are sufficient to recover the ranking. The theoretical guarantee of the convergence theorem does not apply to greedy selection, as it violates assumption 5b. $\varepsilon$-greedy exploration satisfies assumption 5b and thereby provides the theoretical guarantee without harming empirical performance.

Overall, this experiment validates our third assumption and assumption 5b. BBTM converges to the true ranking while empirically confirming the convergence results under the theorem’s assumptions.

\subsection{Rank Distribution Recovery}
In addition to the rank accuracy, we evaluate how well the BBTM recovers the full rank distribution. We first generate the target rank distribution using true generative parameters. Using the posterior samples, we can then generate the estimated BBTM rank distribution, and the Jensen-Shannon divergence (JSD) helps us to compare these two distributions. The JSD value ranges between $0$ (identical distributions) and $1$ (completely disjoint distributions)~\citep{lin1991divergence} when using base 2. In this analysis we consider the JSD value of the worst-matched item as the performance measure. We then perform the Wilcoxon signed rank test as we did in Section~\ref{win_methd}.

\begin{figure}[H]
    \centering
    \includegraphics[width=\textwidth, height=10cm, keepaspectratio]{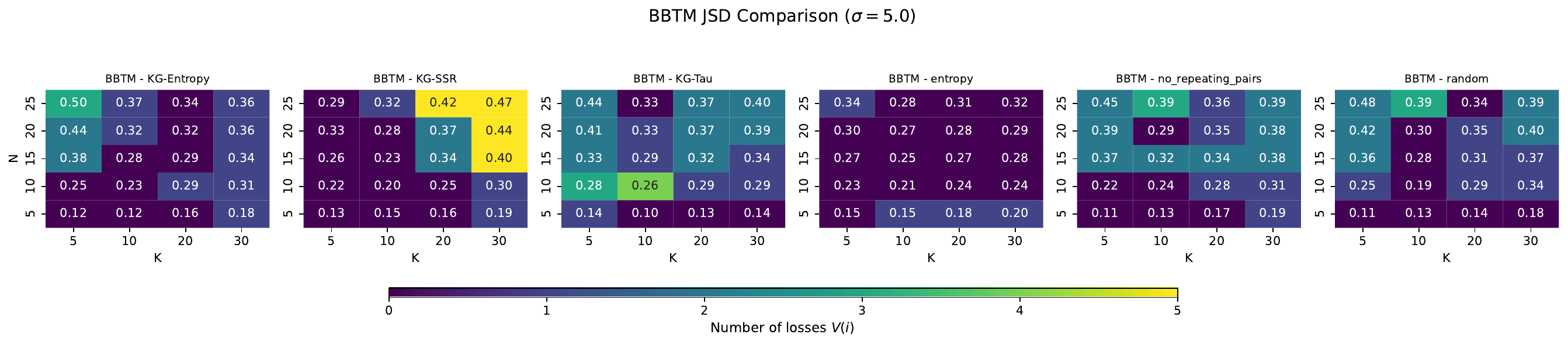}
    \caption{BBTM rank-distribution recovery at $\sigma = 5.0$. Each panel corresponds to one pair-selection method, evaluated over the $N \times K$ grid. The colour scale shows the number of losses $V(i)$ under the Wilcoxon signed-rank test, with the darkest cells indicating a method not beaten by any other ($V(i)=0$). The number in white reports the median worst-matched JSD over the repeats, with lower values indicating better recovery of the target rank distributions. JSD heatmaps for the remaining noise levels are given in Appendix~\ref{app:heatmap_jsd-results}.}
    \label{fig:heatmap_jsd-sigma-50}
\end{figure}

Figure~\ref{fig:heatmap_jsd-sigma-50} shows a representative pattern for $\sigma=5.0$ while Appendix~\ref{app:heatmap_jsd-results} shows the patterns for other noise levels. Entropy pair selection outperforms all other pair selections in both rank accuracy and rank distribution recovery at the noise level $\sigma=5.0$. Entropy's advantage is clearest at the two extreme noise levels. At $\sigma=0.0001$ and $\sigma=10.0$, entropy is near perfect with no losses. At moderate noise levels $\sigma=1.25, 2.5$, KG-SSR and entropy compete closely.\\

A notable finding emerges at $\sigma = 10.0$, where the two metrics diverge. In the $\tau$ distance, KG-SSR overtakes entropy, but in the JSD, entropy is clearly better. That is, at high noise KG-SSR recovers the point ranking ($\tau$) better, whereas entropy recovers the full posterior distribution (JSD) better. The best selection strategy, therefore, depends on the metric of interest. Entropy directly targets uncertainty, which naturally calibrates the posterior distribution, while KG-SSR optimises separation, favouring the ordering but not the shape of the distribution.\\

\begin{figure}[H]
    \centering
    \includegraphics[width=\textwidth]{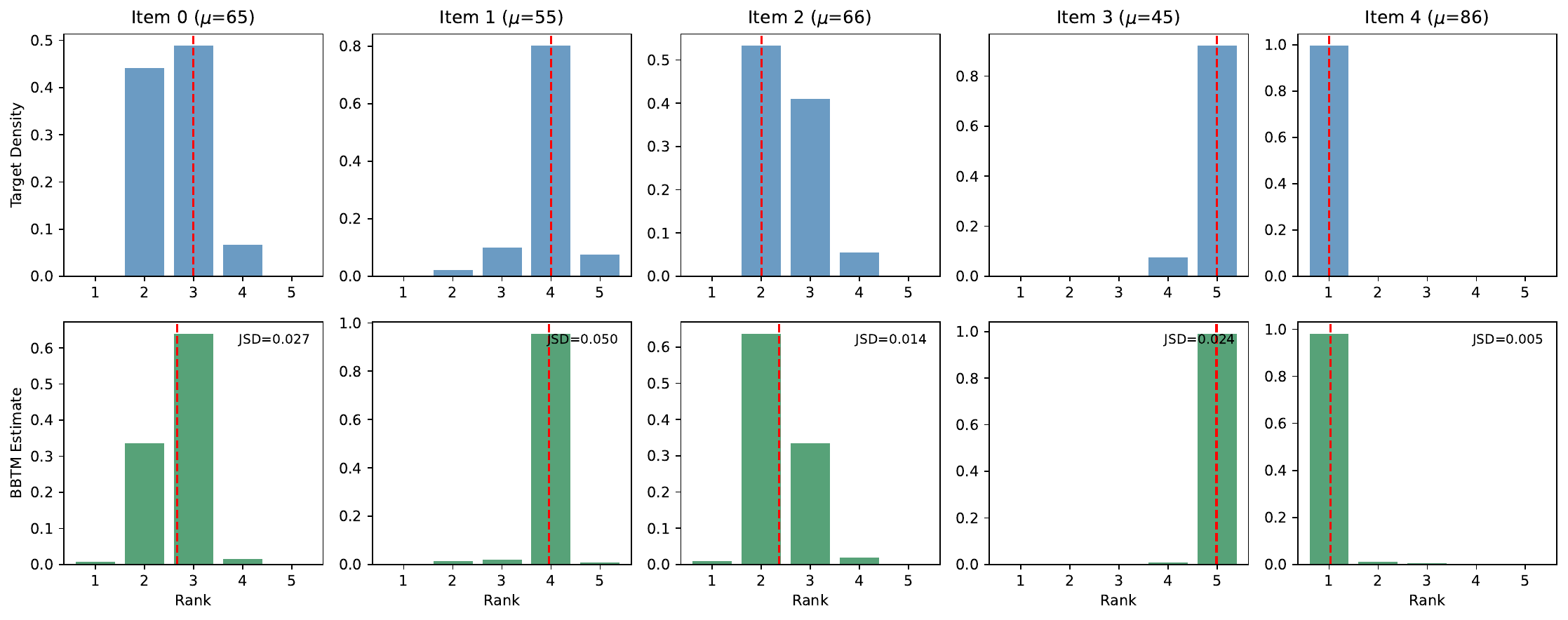}
    \caption{A comparison between target rank densities and the estimated rank densities from the BBTM after $N \times K=5\times10 = 50$ pairwise comparisons. The top row shows the target rank distributions, whilst the bottom row shows the posterior rank densities. The expected ranks of items marked by the red dashed line and the per-item JSD value is given.}
    \label{fig:MLR-50}
\end{figure}

A representative configuration for $N=5$, $K=10$ by BBTM at $\sigma=5.0$ is shown in Figure~\ref{fig:MLR-50}. The quality of rank recovery clearly depends on the degree of item separation. For instance, Item 4 ($\mu = 86$) and Item 3 ($\mu=45$) are well separated as they are the clear best and worst, respectively. BBTM recovers them perfectly.\\   

Item 2 ($\mu=66$) and item 0 ($\mu=65$) are separated by only one mark. Because of this, their ranks are highly ambiguous, and hence their target distributions are spread. BBTM, however, correctly reproduces their spread without false confidence ($\mathrm{JSD} = 0.014, 0.027$). BBTM not only produces the point ranks but also calibrates the rank uncertainty well. The model is confident if items are well separated and uncertain when they are close. This uncertainty quantification is the main advantage of the Bayesian approach.

\subsection{Posterior Separation Reliability (PSR): Empirical Behaviour and Validity}
\begin{figure}[H]
\centering

\begin{subfigure}{0.32\textwidth}
    \centering
    \includegraphics[width=\linewidth]{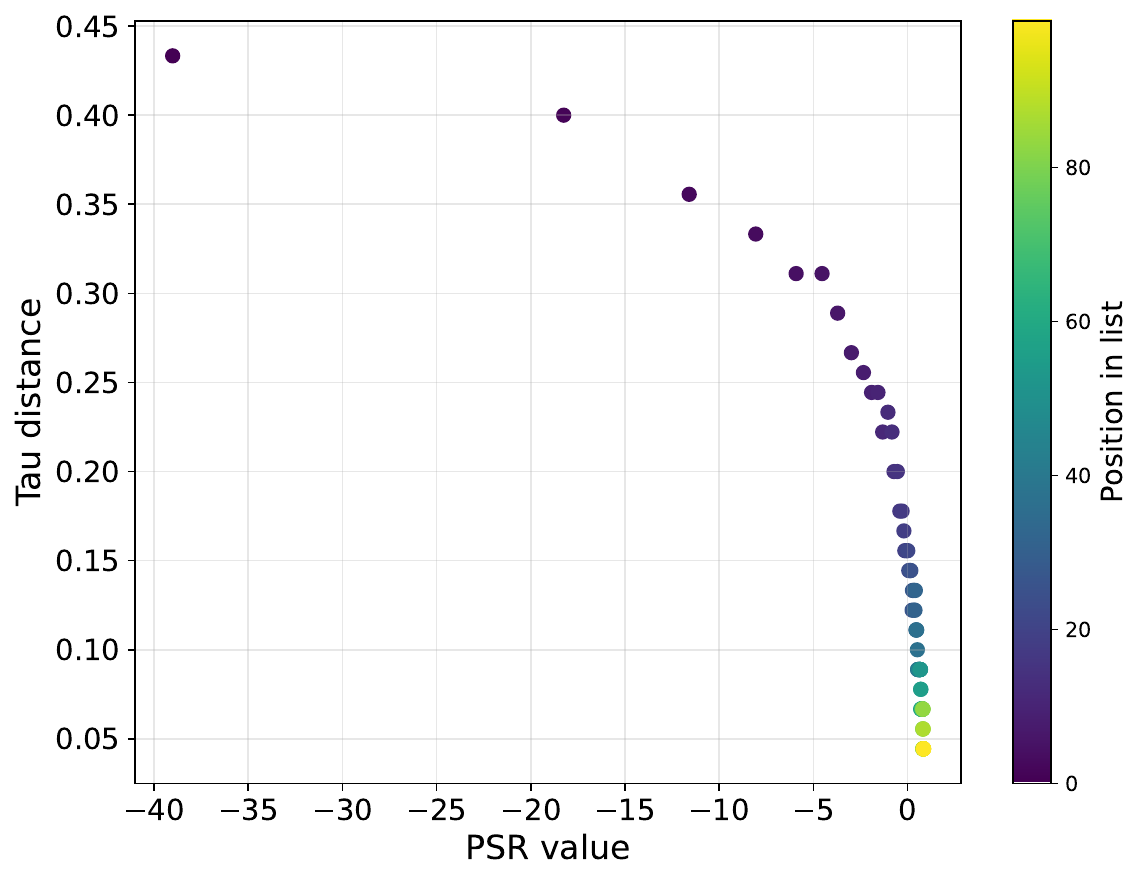}
    \caption{PSR \& Tau value plot}
    \label{fig:1a}
\end{subfigure}
\hfill
\begin{subfigure}{0.32\textwidth}
    \centering
    \includegraphics[width=\linewidth]{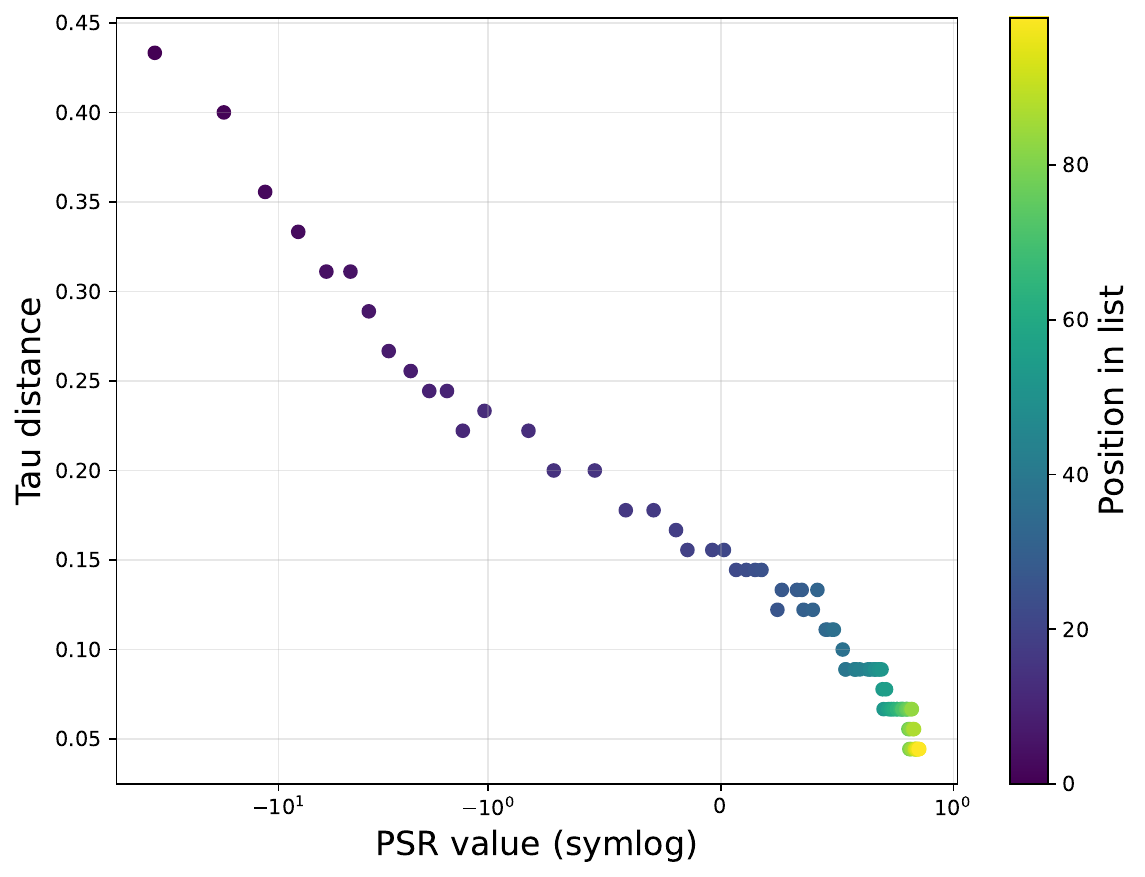}
    \caption{PSR Symlog Scale}
    \label{fig:1b}
\end{subfigure}
\hfill
\begin{subfigure}{0.32\textwidth}
    \centering
    \includegraphics[width=\linewidth]{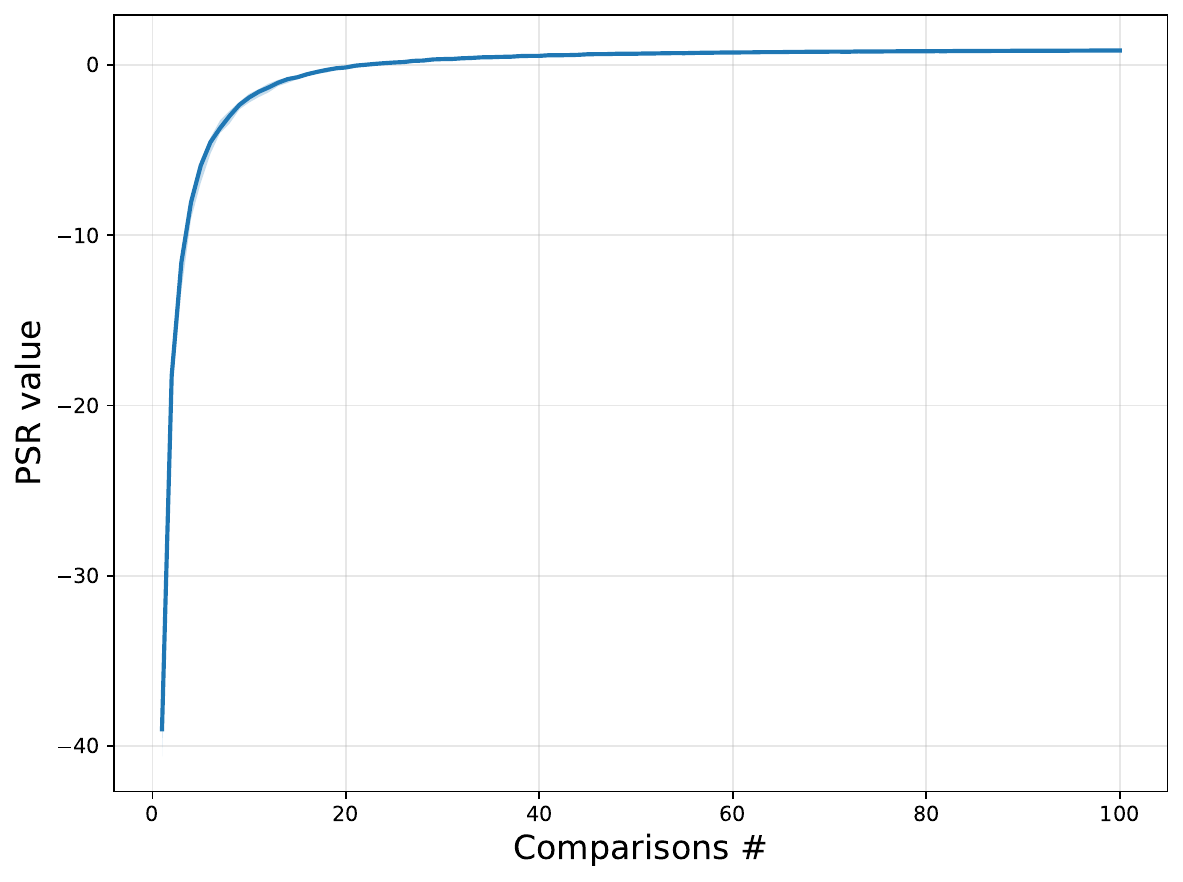}
    \caption{PSR value}
    \label{fig:1c}
\end{subfigure}

\caption{PSR as a Bayesian reliability measure in BBTM. Results shown for KG-Entropy with N = 10 items, K = 10, $\sigma$ = 5.0 (a) Median PSR versus Median normalised Kendall's Tau distance across 50 runs (colour indicates the comparison index). (b) Symlog transformation of the PSR values. (c) Median PSR values over comparisons.}
\label{fig:comparison}
\end{figure}

\Cref{fig:comparison} illustrates the behaviour of the median PSR value over 50 repetitions under KG-Entropy for $N = 10$ items, $K = 10$ comparisons per item, and $\sigma = 5.0$ for each item. \Cref{fig:1a} demonstrates the relationship between PSR value and ranking accuracy. It can be seen that the PSR value increases and the normalised tau value decreases towards zero as the number of comparisons increases. This shows that PSR is aligned with the ground truth rank order. \Cref{fig:1b} shows that, at the start, PSR can take large negative values when posterior uncertainty dominates the between-item spread in the posterior mean strengths. This can be seen clearly because of the Symlog scale. \Cref{fig:1c} confirms this interpretation by showing that PSR increases monotonically over comparisons. This is because the posterior distributions transition from uncertainty-dominated estimates to strong posterior separation as additional evidence accumulates. Therefore, these results support PSR as a reliable Bayesian reliability indicator for BBTM. It tracks the extent to which inferred latent strengths are separated relative to the posterior uncertainty. Furthermore, its growth is aligned with improved rank recovery as comparisons accumulate.\\

Since KG-Entropy neither optimises separation directly nor attains the best ranking accuracy, we used KG-Entropy to illustrate this scenario. Note that the PSR and $\tau$ relationship reflects a general property rather than an artefact of the selection method. The PSR is independent of how pairs are selected as it is computed solely from the posterior. Appendix~\ref{app:psr-results} shows similar illustrations for entropy and KG-SSR. Those illustrations clearly follow the same pattern.\\

This reliability measure can be implemented for any Bayesian implementation for CJ, with or without Bradley-Terry likelihood, and with either continuous or discrete posterior distributions. It is easy to calculate the variance of posterior means and the average of posterior variances in latent strengths for the continuous case; however, the calculation of these values for discrete cases is not straightforward.

\subsection{Robustness of the model}
\begin{figure}[H]
    \centering
    \includegraphics[width=\textwidth, height=8cm, keepaspectratio]{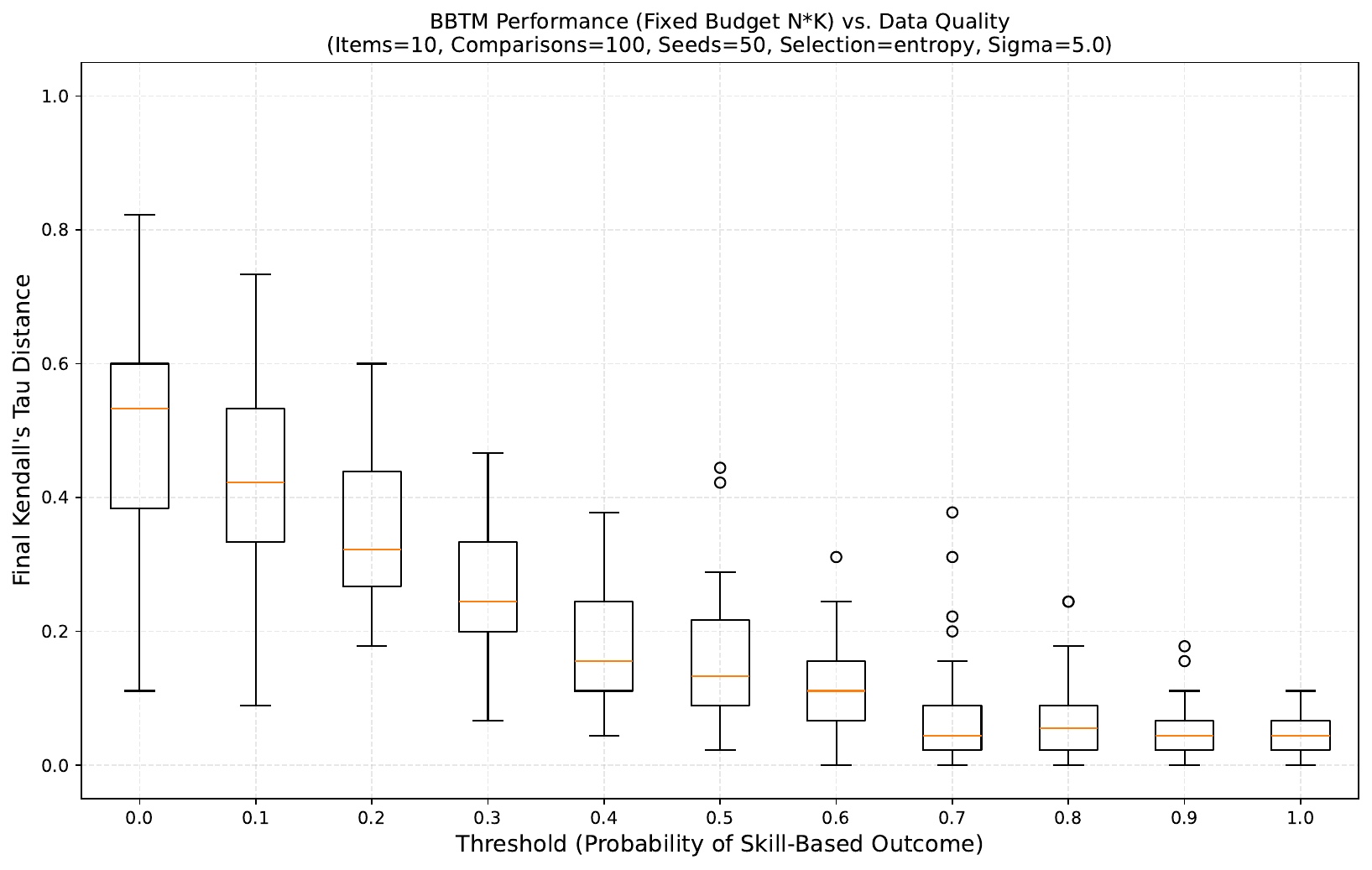}
    \caption{BBTM robustness to the true skill-based and coin toss outcomes for the entropy pair selection at $\sigma=5.0$ for $N=10$, $K=10$, and $50$ seeds. The $X$-axis shows the threshold level ($\alpha$), which is the proportion of true skill-based outcomes, and $(1-\alpha)$ is the proportion of coin toss outcomes. The distribution of the final normalised $\tau$ value of 50 seeds for a fixed budget ($N \times K$) is shown in each box plot at each $\alpha$ level. Results for the noise levels $\sigma = 1.25, 5.0,$ and $10.0$ for entropy and KG-SSR pair selections are given in Appendix~\ref{app:robustness-results}.}
    \label{fig:robustness_bbtm-sigma-5.0}
\end{figure}

It is evident that the BBTM entropy pair selection performs well for moderate noise levels.  Figure~\ref{fig:robustness_bbtm-sigma-5.0} illustrates the behaviour of BBTM for entropy pair selection at $\sigma=5.0$ when we change the proportion of true skill-based and random coin toss outcomes for a fixed budget. As we increase the true skill-based outcomes, the $\tau$ distance decreases, which means the rank accuracy increases. At $\alpha=0.0$, the median $\tau = 0.53$ implies chance-level ranking. Random coin toss outcomes are occasionally correct but do not provide systematic skill information. Notably, $\tau$ settles at chance rather than exceeding it. This indicates the model degrades gracefully rather than overfitting to the noise.\\       

The diminishing returns pattern is the most important finding of this experiment. The $\tau$ distance reduces steeply and then plateaus as informed outcomes increase. According to Figure~\ref{fig:robustness_bbtm-sigma-5.0}, median $\tau$ at $\alpha = 0.7$ and $\alpha=1.0$ is almost the same (0.044). This indicates that 70\% of genuine outcomes (with 30\% random coin toss outcomes) are enough to get a near-optimal ranking. We believe this might be cost-effective, as it can reduce costly human judgments without significantly reducing ranking accuracy.\\

According to the Table~\ref{tab:robustness-summary}, the tolerance depends on the noise level ($\sigma$) and the pair selection (also see Appendix~\ref{app:robustness-results}). At a low noise level ($\sigma=1.25$), the model can tolerate a higher proportion of random coin-toss outcomes (up to 40\% at the $\tau \leq 0.05$ target) without affecting the ranking accuracy. At a higher noise level ($\sigma=10.0$), the model requires a higher proportion of skill-based outcomes to achieve an optimal rank. However, the model could not achieve $\tau \leq 0.05$ target at this noise level. This is because the noise itself limits the ranking accuracy. The entropy pair selector reaches the target faster than KG-SSR in most settings, and hence entropy needs relatively fewer informed outcomes to recover the near-optimal ranking.\\

\begin{table}[H]
\centering
\caption{Summary of the BBTM robustness results across noise levels and selection strategies. At $\alpha = 0$ (all outcomes random), the median $\tau \approx 0.53$ for all configurations. This is no better than ranking the items at random. The floor is the median $\tau$ at $\alpha = 1$ (all outcomes skill-based), giving the best accuracy at each $\sigma$. The threshold $\alpha^{*}$ is the smallest proportion of skill-based outcomes required to reach the accuracy targets of $\tau \leq 0.10$ and $\tau \leq 0.05$. The tolerated random level $(1 - \alpha^{*})$ is the proportion of genuine judgements that can be replaced by random outcomes while still achieving this accuracy. All values are medians over $50$ seeds at a fixed budget of $N \times K = 100$ comparisons ($N = 10$, $K = 10$).}
\label{tab:robustness-summary}
\begin{tabular}{c l c c c c c}
\toprule
& & & \multicolumn{2}{c}{$\tau \leq 0.10$} & \multicolumn{2}{c}{$\tau \leq 0.05$} \\
\cmidrule(lr){4-5} \cmidrule(lr){6-7}
$\sigma$ & Selection & Floor $\tau$ & $\alpha^{*}$ & \makecell{Tolerated \\ random level} & $\alpha^{*}$ & \makecell{Tolerated \\ random level} \\
\midrule
\multirow{2}{*}{$1.25$} & entropy & $0.00$ & $0.5$ & $50\%$ & $0.6$ & $40\%$ \\
                        & KG-SSR  & $0.00$ & $0.5$ & $50\%$ & $0.7$ & $30\%$ \\
\midrule
\multirow{2}{*}{$5.0$}  & entropy & $0.04$ & $0.7$ & $30\%$ & $0.7$ & $30\%$ \\
                        & KG-SSR  & $0.04$ & $0.8$ & $20\%$ & $0.9$ & $10\%$ \\
\midrule
\multirow{2}{*}{$10.0$} & entropy & $0.09$ & $0.8$ & $20\%$ & $-$ & $-$ \\
                        & KG-SSR  & $0.07$ & $1.0$ & $0\%$  & $-$ & $-$ \\
\bottomrule
\end{tabular}
\end{table}
    
\newpage

\section{Conclusions}\label{Conclusion}

\newpage

\bibliographystyle{apalike}
\bibliography{reference}

\newpage
\begin{appendices}

\section{Additional Convergence plots}
\label{app:conv_plots}
This appendix presents the convergence plots for noise levels $\sigma = 1.25$ and $\sigma = 10.0$.
\begin{figure}[H]
    \centering
    \includegraphics[width=\textwidth]{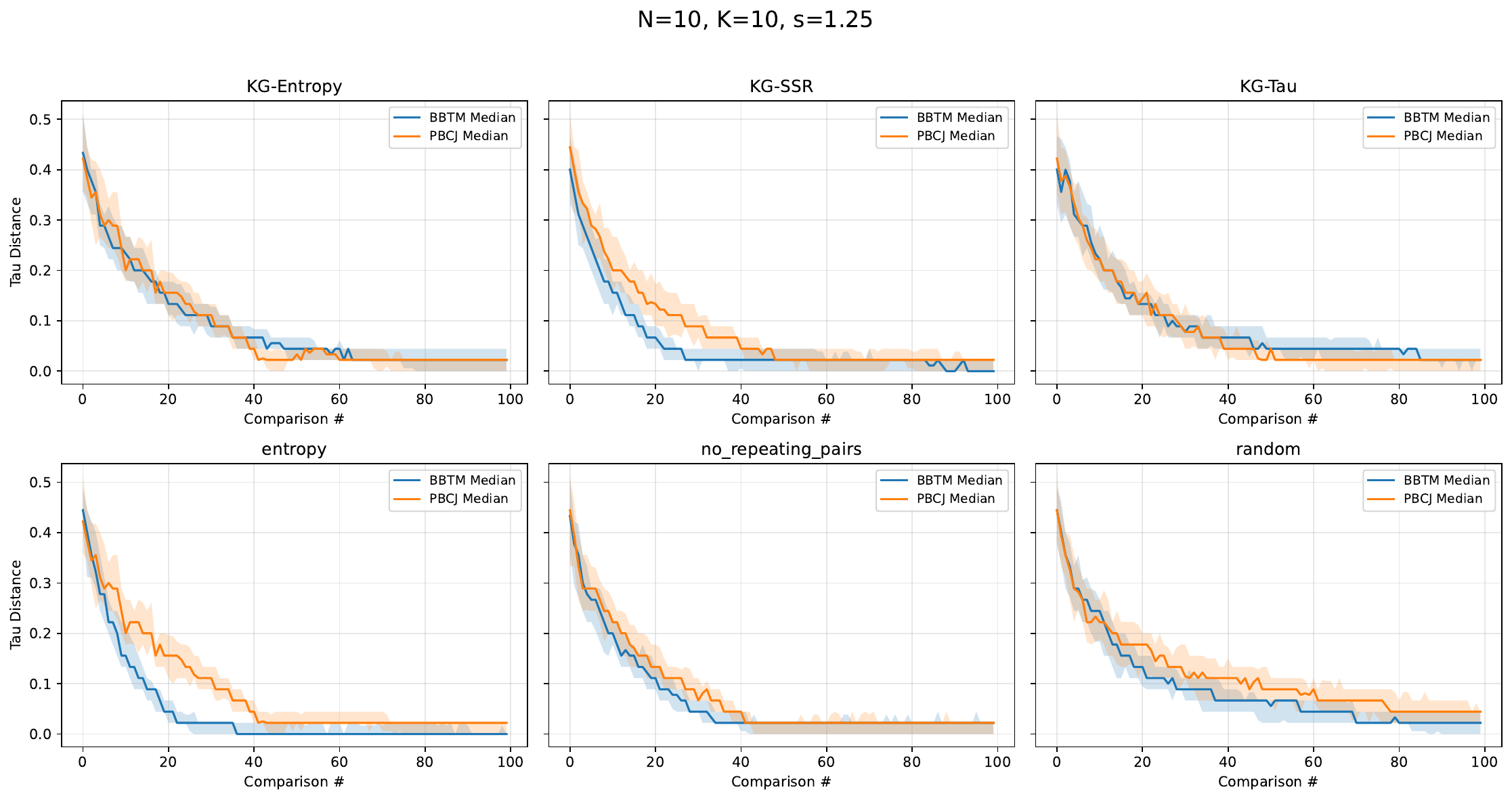}
    \caption{Model convergence plot for $\sigma = 1.25$.}
    \label{fig:app-conv-sigma-125}
\end{figure}

\begin{figure}[H]
    \centering
    \includegraphics[width=\textwidth]{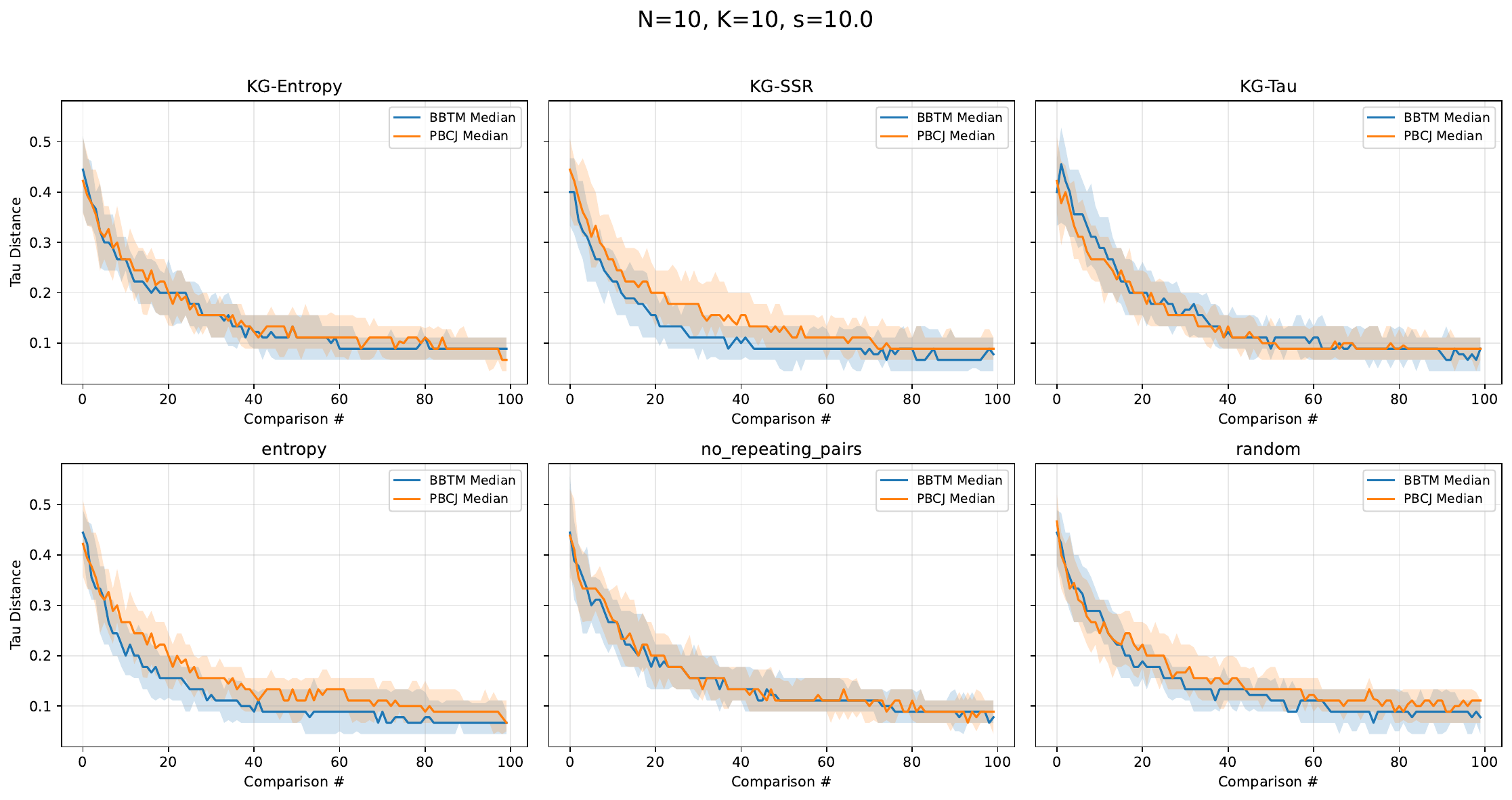}
    \caption{Model convergence plot for $\sigma = 10.0$.}
    \label{fig:app-conv-sigma-10}
\end{figure}

\section{Additional Heatmap Results}
\label{app:heatmap-results}
This appendix presents the full set of heatmap results for each noise level $\sigma$.
\begin{figure}[H]
    \centering
    \includegraphics[width=\textwidth]{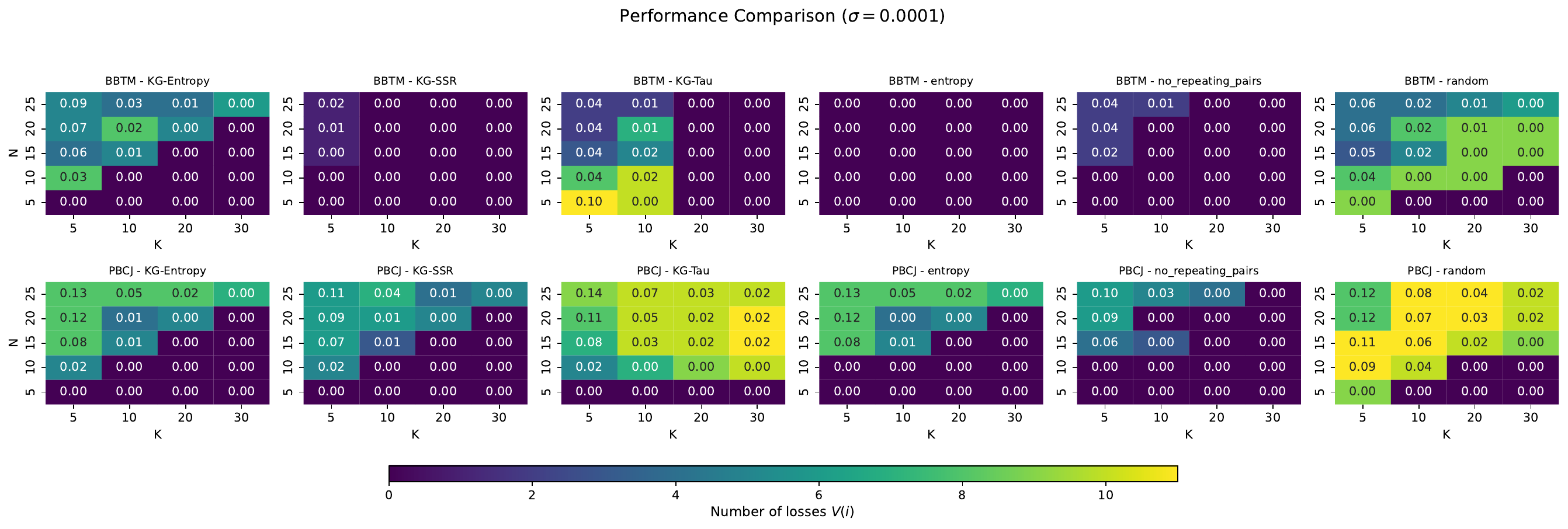}
    \caption{Performance comparison heatmap for $\sigma = 0.0001$.}
    \label{fig:app-heatmap-sigma-0001}
\end{figure}

\begin{figure}[H]
    \centering
    \includegraphics[width=\textwidth]{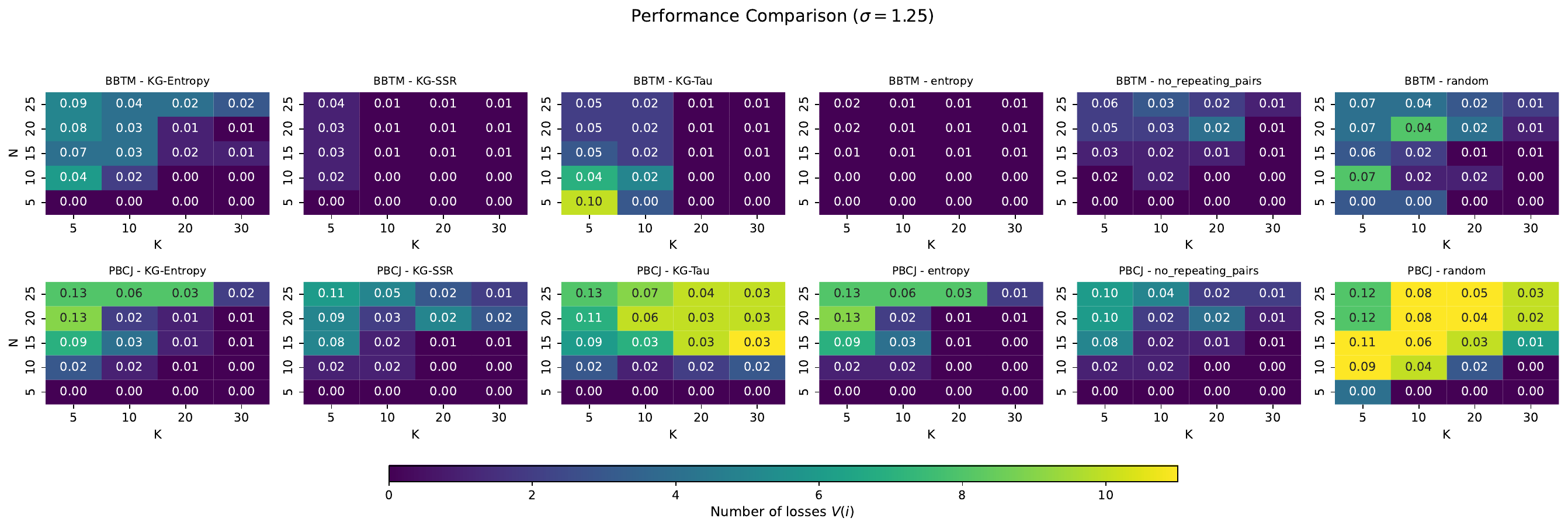}
    \caption{Performance comparison heatmap for $\sigma = 1.25$.}
    \label{fig:app-heatmap-sigma-125}
\end{figure}

\begin{figure}[H]
    \centering
    \includegraphics[width=\textwidth]{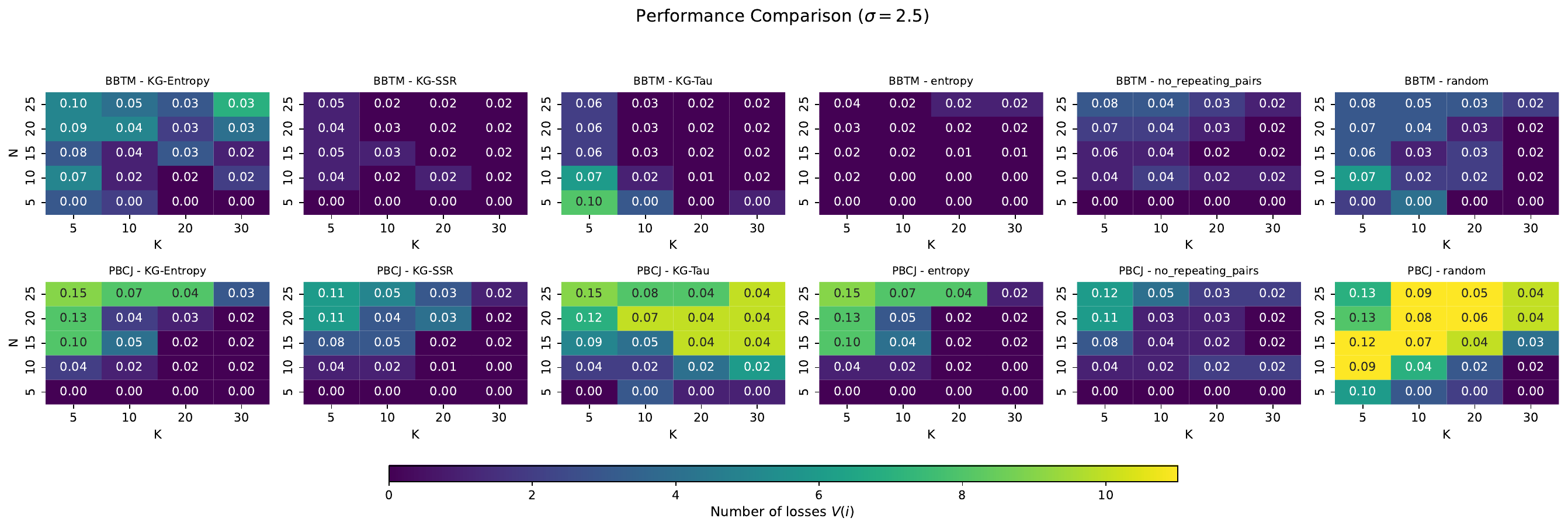}
    \caption{Performance comparison heatmap for $\sigma = 2.5$.}
    \label{fig:app-heatmap-sigma-25}
\end{figure}

\begin{figure}[H]
    \centering
    \includegraphics[width=\textwidth]{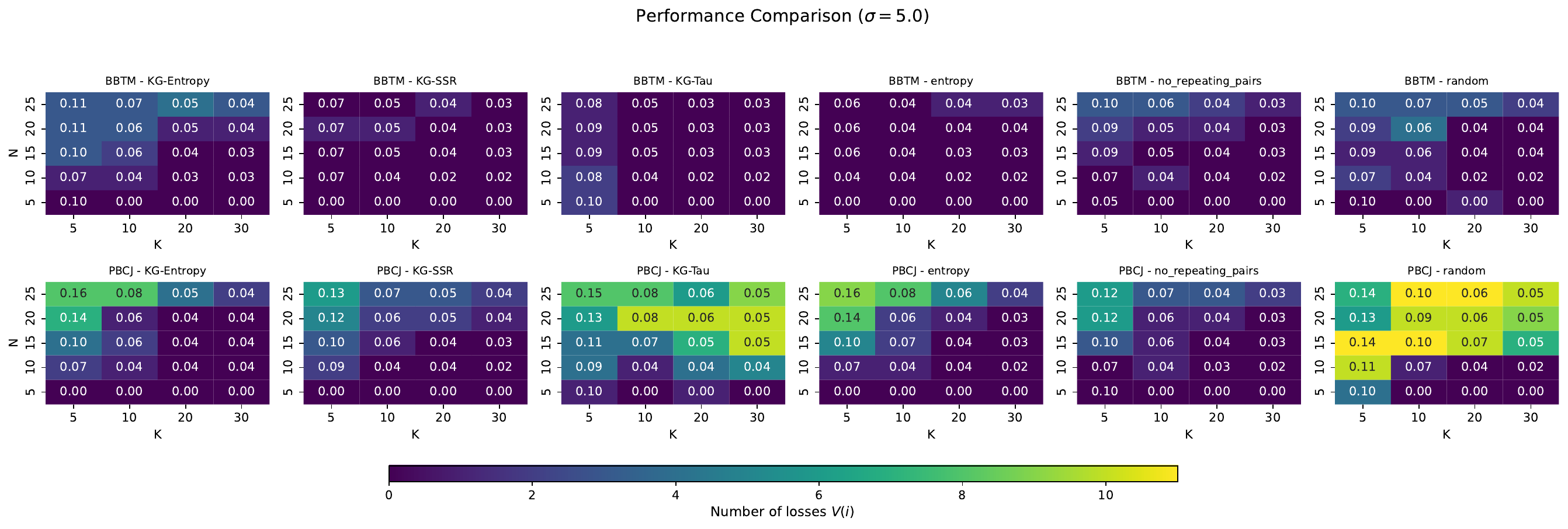}
    \caption{Performance comparison heatmap for $\sigma = 5.0$.}
    \label{fig:app-heatmap-sigma-50}
\end{figure}

\begin{figure}[H]
    \centering
    \includegraphics[width=\textwidth]{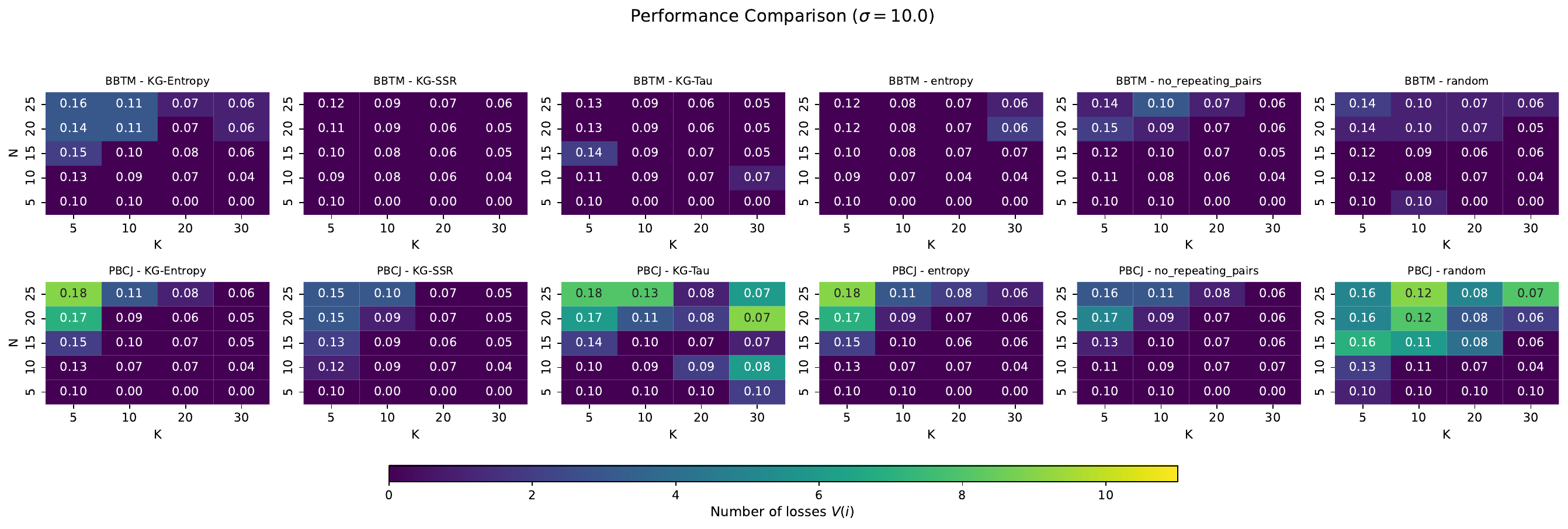}
    \caption{Performance comparison heatmap for $\sigma = 10.0$.}
    \label{fig:app-heatmap-sigma-100}
\end{figure}

\section{Additional JSD Heatmap Results}
\label{app:heatmap_jsd-results}
This appendix presents the full set of heatmap results for each noise level $\sigma$.
\begin{figure}[H]
    \centering
    \includegraphics[width=\textwidth]{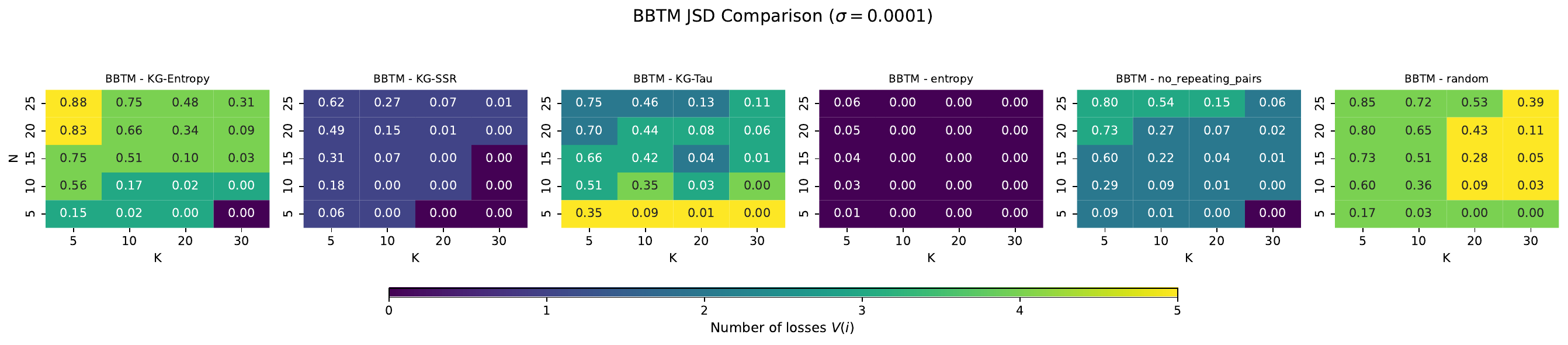}
    \caption{Performance comparison heatmap for $\sigma = 0.0001$.}
    \label{fig:app-heatmap_jsd-sigma-0001}
\end{figure}

\begin{figure}[H]
    \centering
    \includegraphics[width=\textwidth]{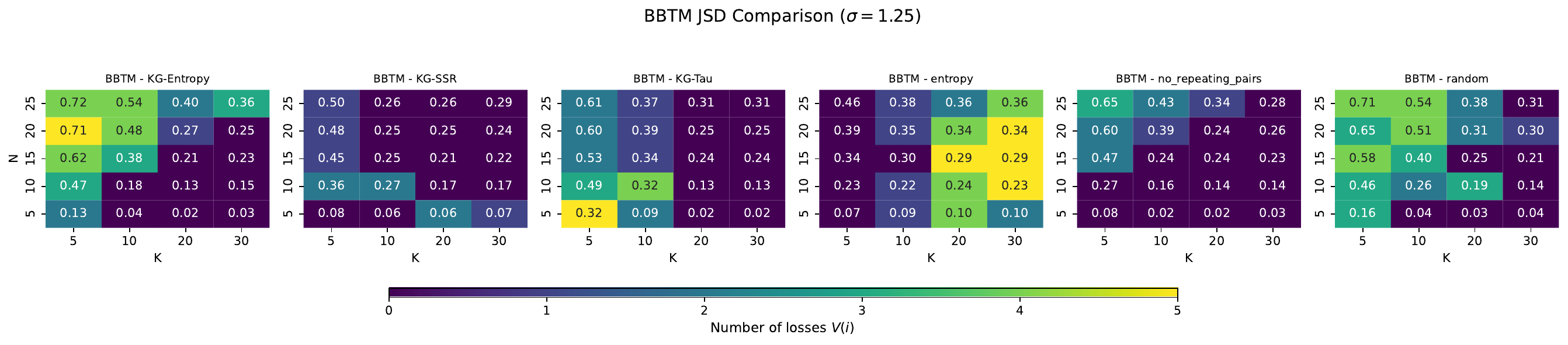}
    \caption{Performance comparison heatmap for $\sigma = 1.25$.}
    \label{fig:app-heatmap_jsd-sigma-125}
\end{figure}

\begin{figure}[H]
    \centering
    \includegraphics[width=\textwidth]{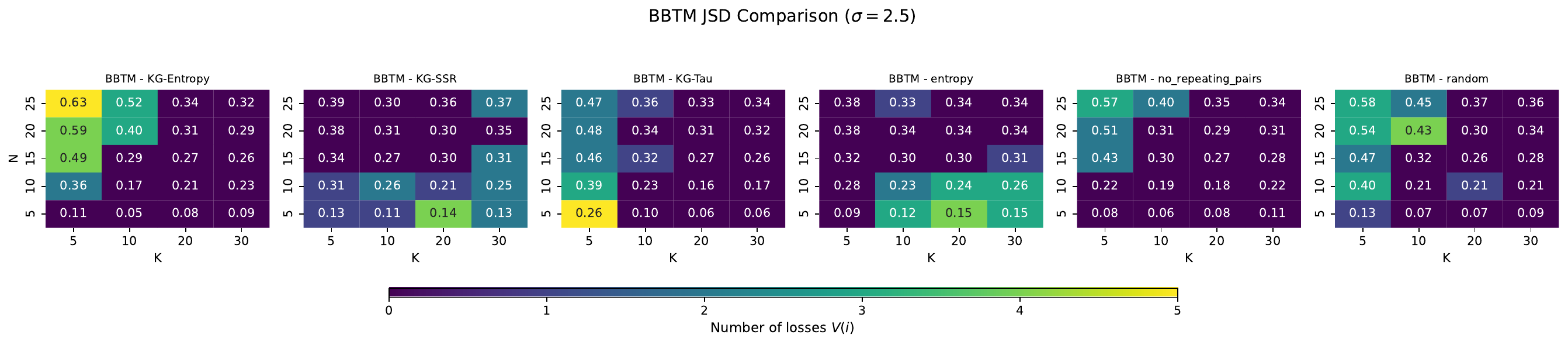}
    \caption{Performance comparison heatmap for $\sigma = 2.5$.}
    \label{fig:app-heatmap_jsd-sigma-25}
\end{figure}

\begin{figure}[H]
    \centering
    \includegraphics[width=\textwidth]{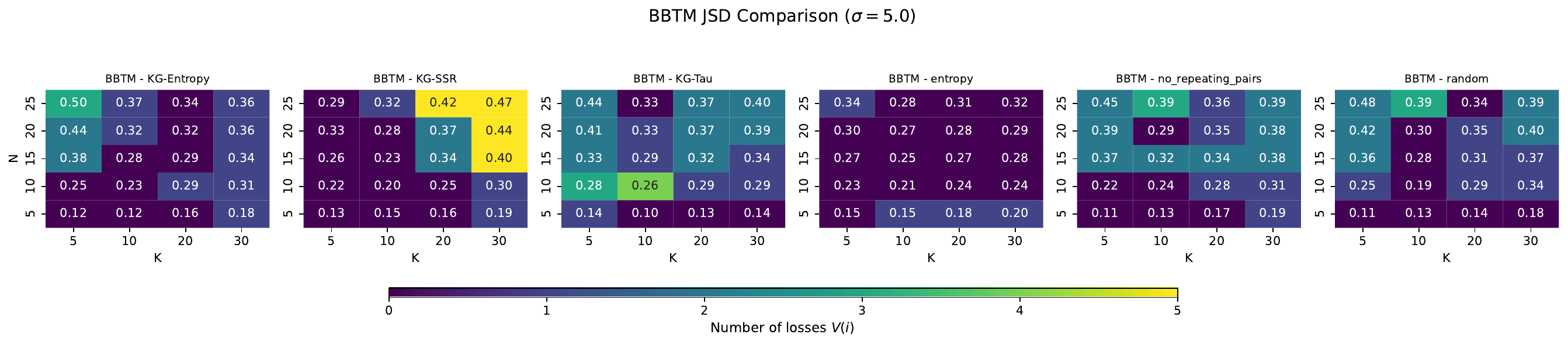}
    \caption{Performance comparison heatmap for $\sigma = 5.0$.}
    \label{fig:app-heatmap_jsd-sigma-50}
\end{figure}

\begin{figure}[H]
    \centering
    \includegraphics[width=\textwidth]{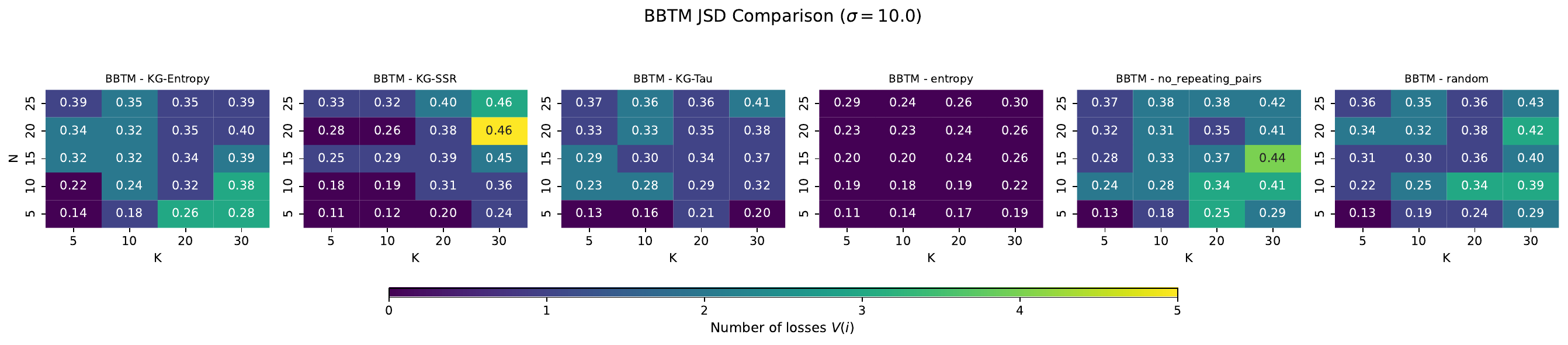}
    \caption{Performance comparison heatmap for $\sigma = 10.0$.}
    \label{fig:app-heatmap_jsd-sigma-100}
\end{figure}

\section{Additional PSR Plots}
\label{app:psr-results}

This appendix presents the PSR plots for entropy and KG-SSR, for the same 
configuration as KG-Entropy ($N=10$, $K=10$, $\sigma=5.0$).

\begin{figure}[H]
\centering
\begin{subfigure}{0.32\textwidth}
    \centering
    \includegraphics[width=\linewidth]{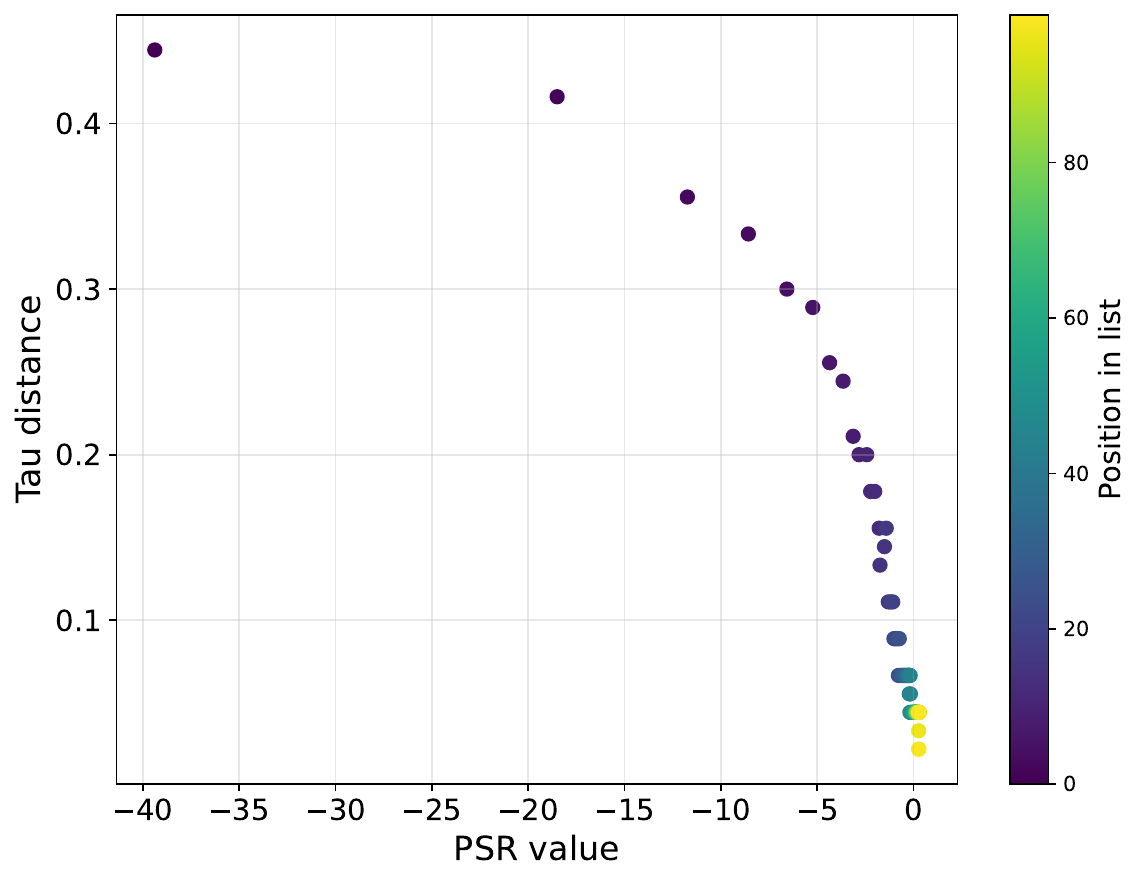}
    \caption{PSR \& $\tau$ value plot}
    \label{fig:psr-ent-a}
\end{subfigure}
\hfill
\begin{subfigure}{0.32\textwidth}
    \centering
    \includegraphics[width=\linewidth]{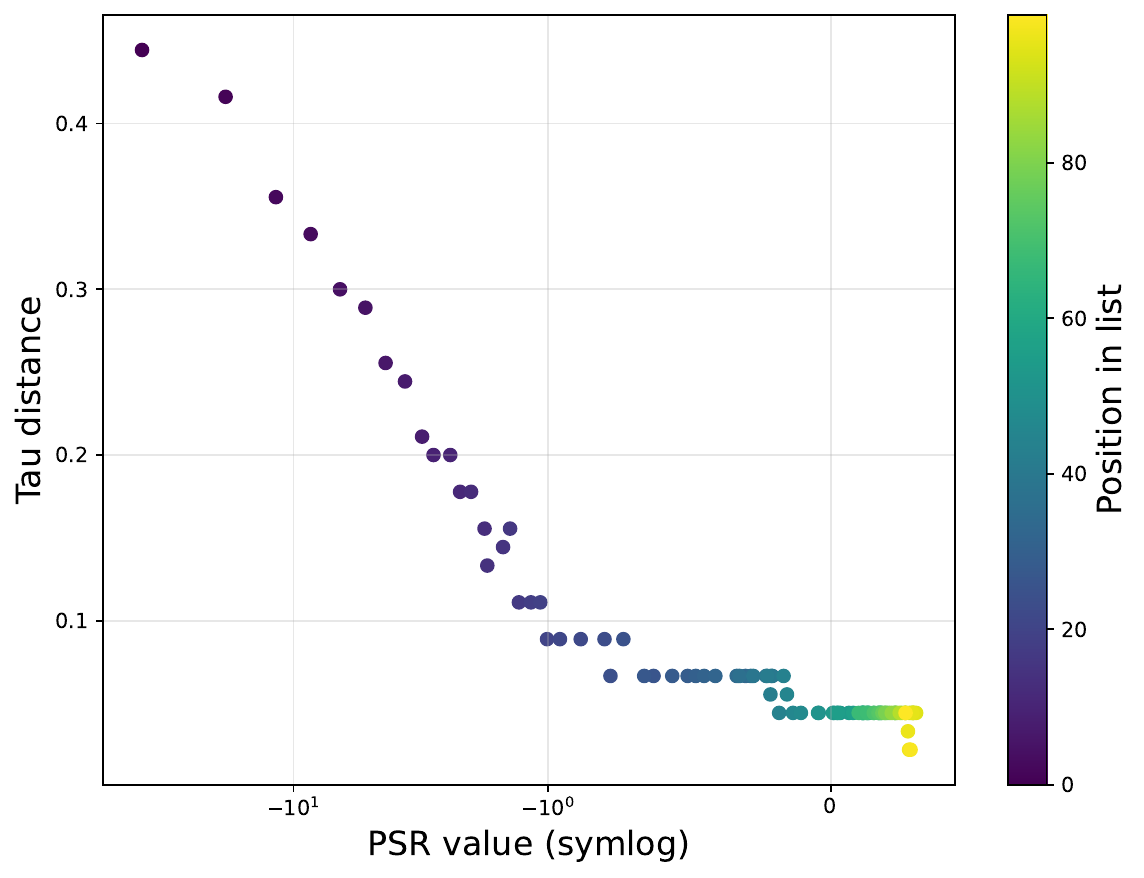}
    \caption{PSR symlog scale}
    \label{fig:psr-ent-b}
\end{subfigure}
\hfill
\begin{subfigure}{0.32\textwidth}
    \centering
    \includegraphics[width=\linewidth]{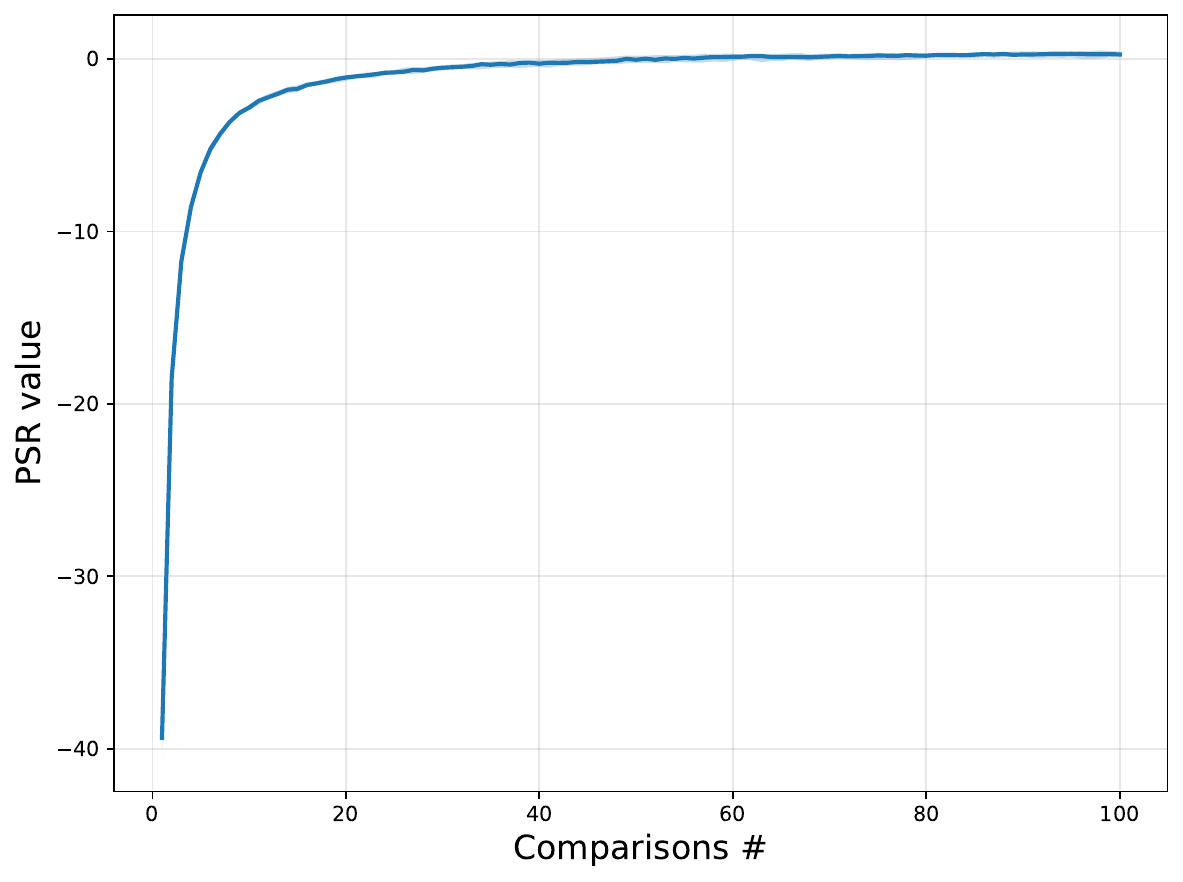}
    \caption{PSR value}
    \label{fig:psr-ent-c}
\end{subfigure}
\caption{PSR behaviour for entropy selection ($N=10$, $K=10$, $\sigma=5.0$).}
\label{fig:psr-entropy}
\end{figure}

\begin{figure}[H]
\centering
\begin{subfigure}{0.32\textwidth}
    \centering
    \includegraphics[width=\linewidth]{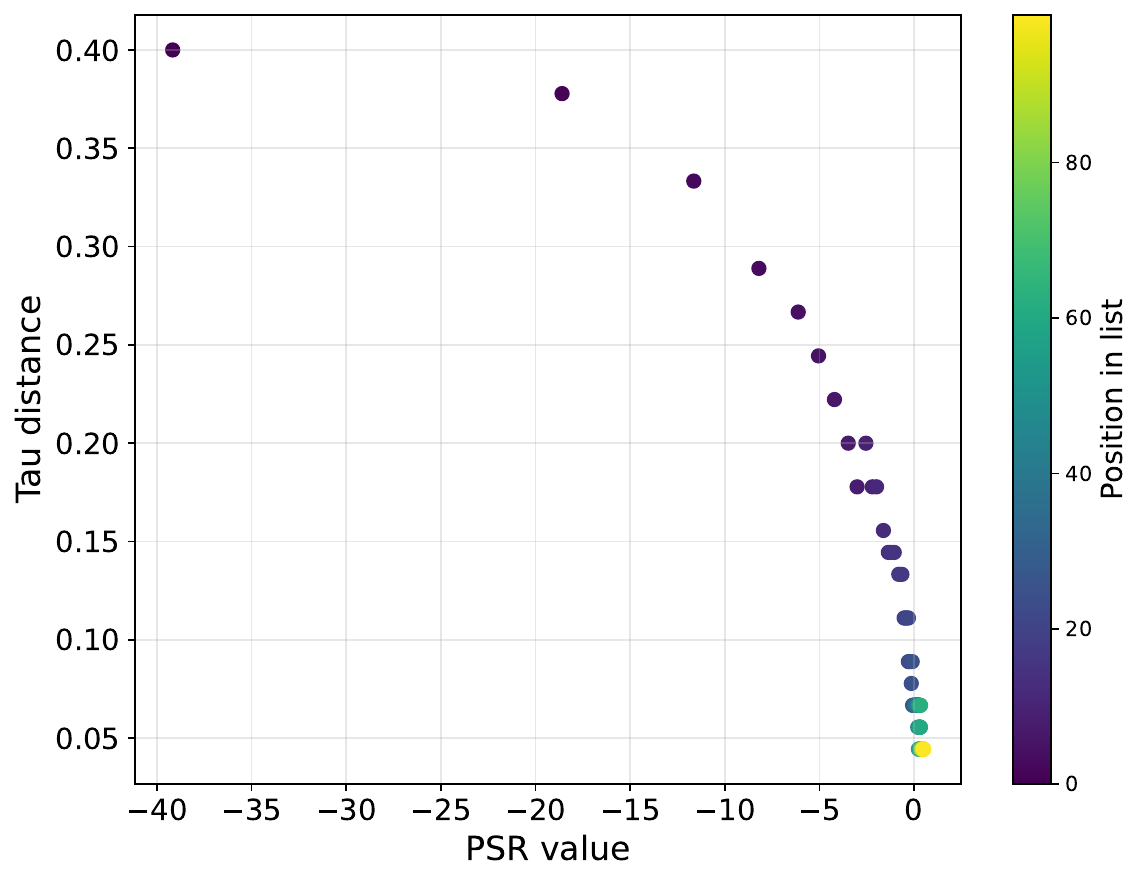}
    \caption{PSR \& $\tau$ value plot}
    \label{fig:psr-kgssr-a}
\end{subfigure}
\hfill
\begin{subfigure}{0.32\textwidth}
    \centering
    \includegraphics[width=\linewidth]{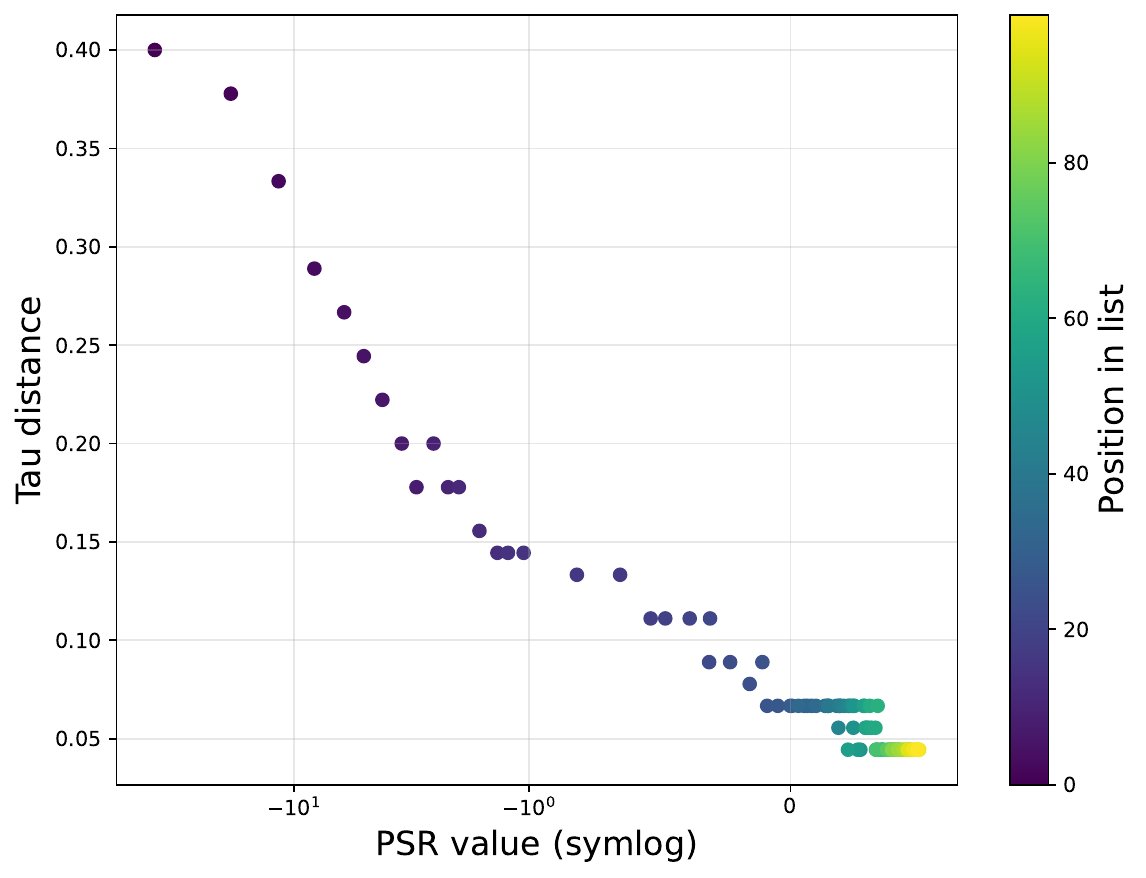}
    \caption{PSR symlog scale}
    \label{fig:psr-kgssr-b}
\end{subfigure}
\hfill
\begin{subfigure}{0.32\textwidth}
    \centering
    \includegraphics[width=\linewidth]{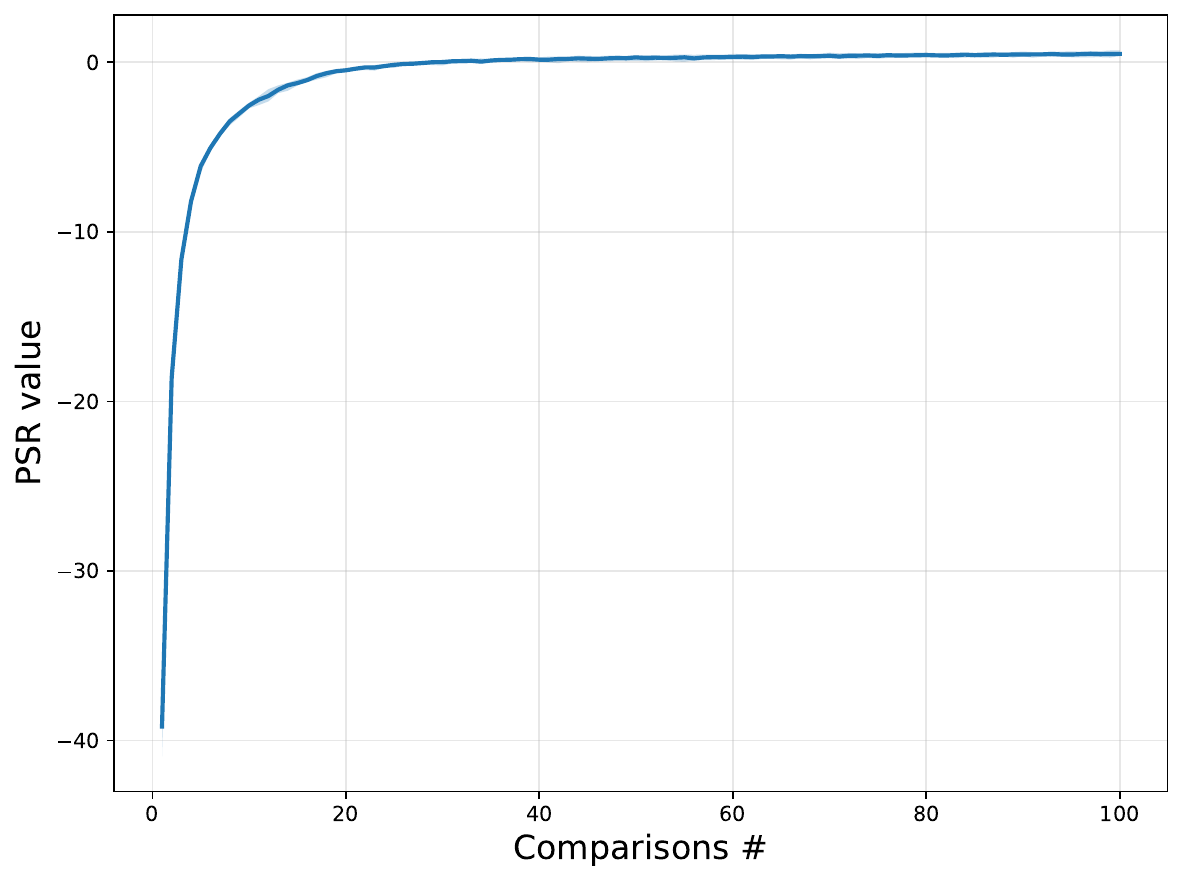}
    \caption{PSR value}
    \label{fig:psr-kgssr-c}
\end{subfigure}
\caption{PSR behaviour for KG-SSR selection ($N=10$, $K=10$, $\sigma=5.0$).}
\label{fig:psr-kgssr}
\end{figure}

\section{Model robustness plots for $\sigma = 1.25, 5.0, 10.0$ for entropy and KG-SSR pair selections}
\label{app:robustness-results}

This appendix includes the BBTM robustness plots for entropy and KG-SSR pair selections, for $N=10$, $K=10$, and $\sigma=1.25, 5.0, 10.0$.

\begin{figure}[H]
\centering
\begin{subfigure}{0.495\textwidth}
    \centering
    \includegraphics[width=\linewidth]{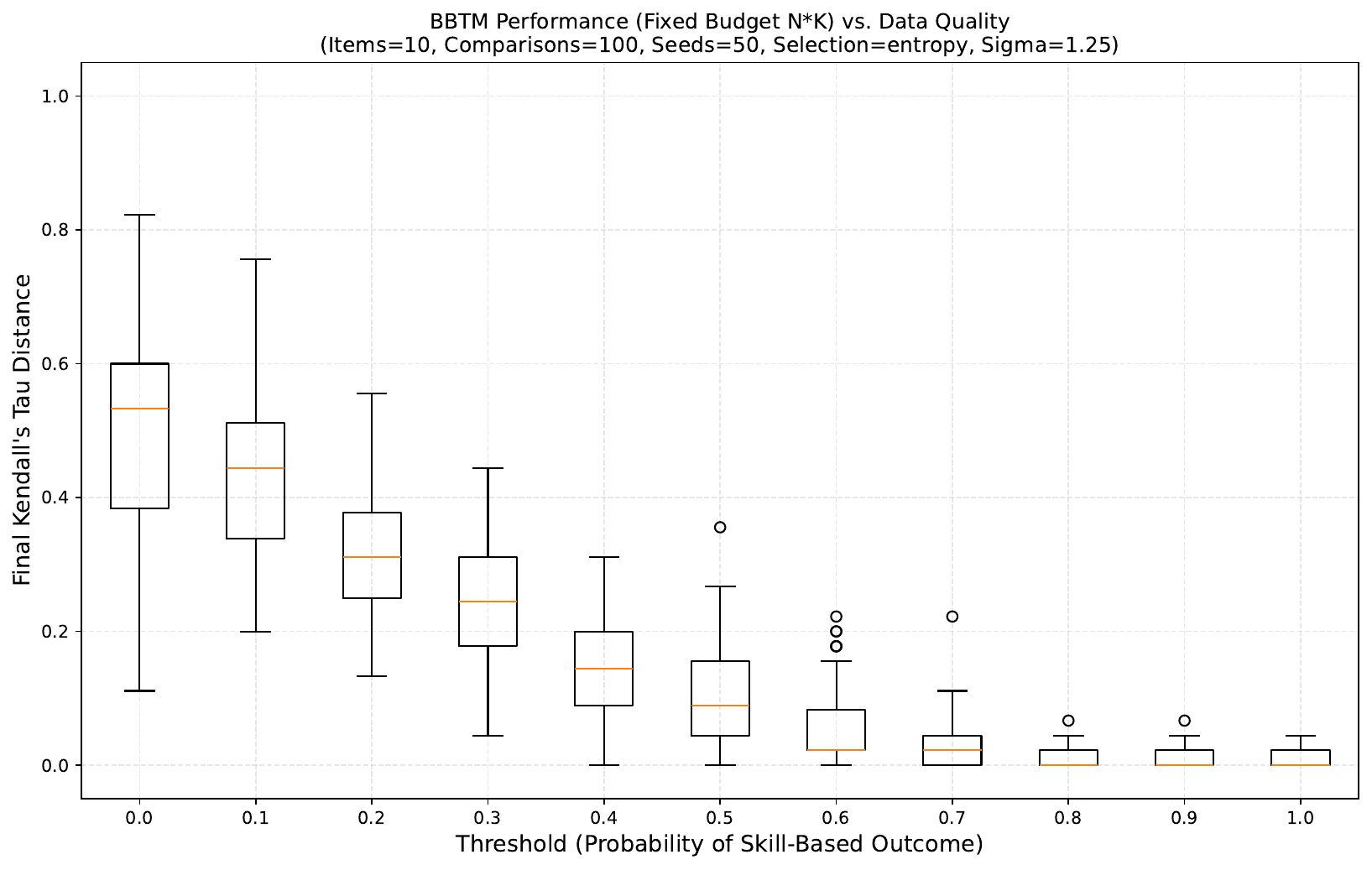}
    \caption{entropy}
    \label{fig:entropy pair selection}
\end{subfigure}
\hfill
\begin{subfigure}{0.495\textwidth}
    \centering
    \includegraphics[width=\linewidth]{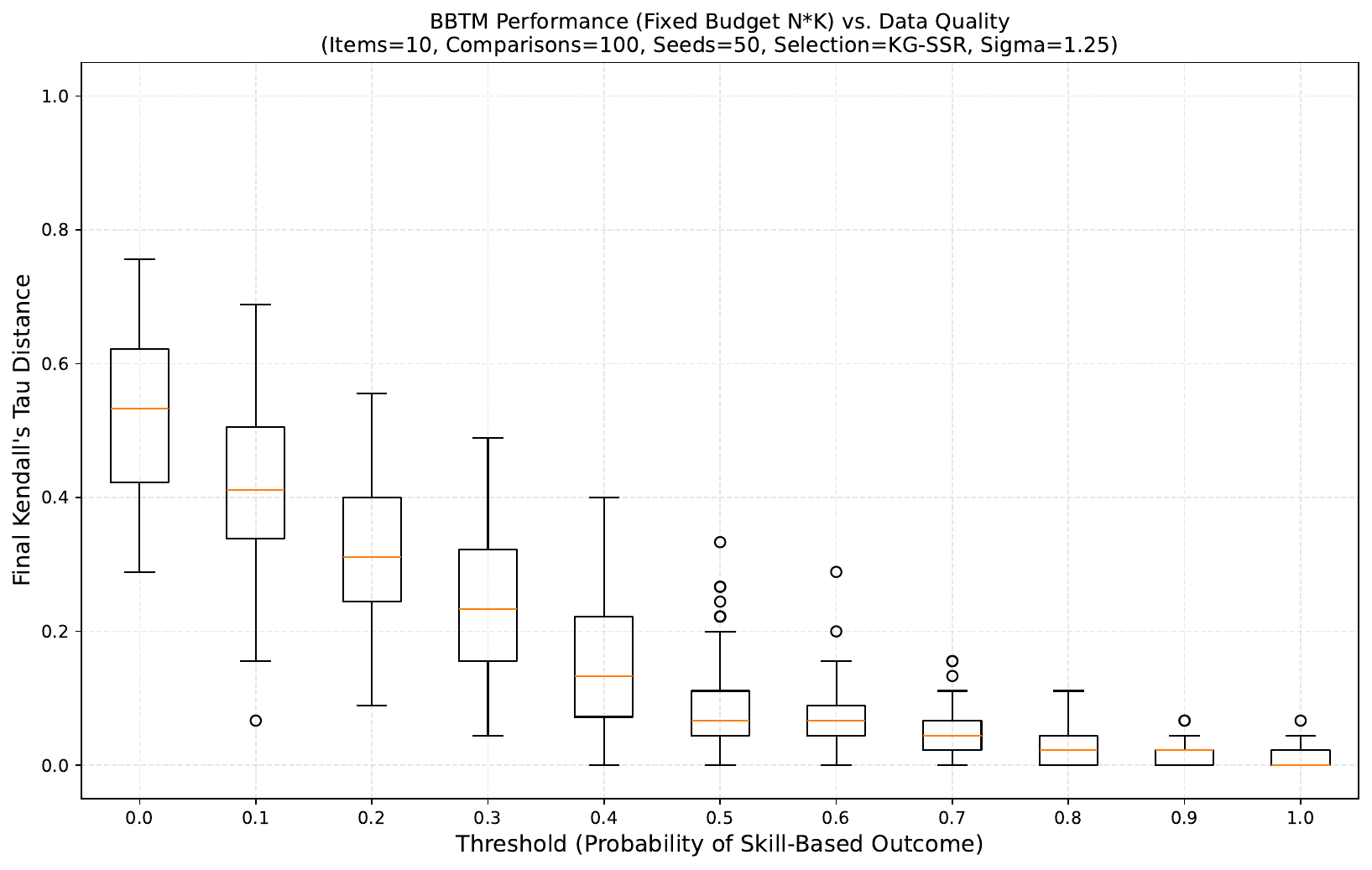}
    \caption{KG-SSR}
    \label{fig:KG-SSR pair selection}
\end{subfigure}

\caption{Robustness plots for $N=10$, $K=10$, $\sigma=1.25$}
\label{fig:psr-entropy}
\end{figure}

\begin{figure}[H]
\centering
\begin{subfigure}{0.495\textwidth}
    \centering
    \includegraphics[width=\linewidth]{robustness_plots/robustness_entropy_sigma_5.0.pdf}
    \caption{entropy}
    \label{fig:entropy pair selection}
\end{subfigure}
\hfill
\begin{subfigure}{0.495\textwidth}
    \centering
    \includegraphics[width=\linewidth]{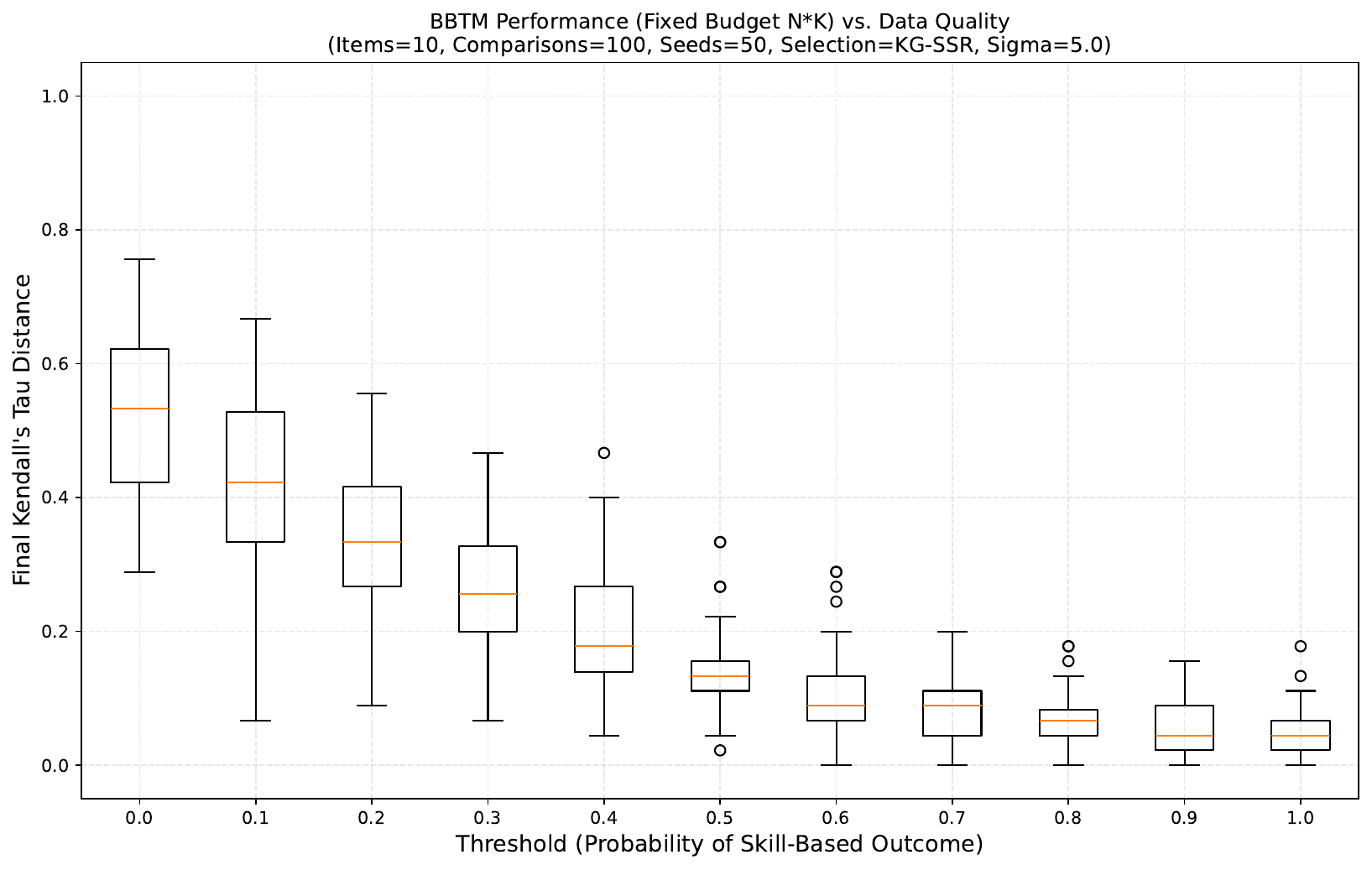}
    \caption{KG-SSR}
    \label{fig:KG-SSR pair selection}
\end{subfigure}

\caption{Robustness plots for $N=10$, $K=10$, $\sigma=5.0$}
\label{fig:psr-entropy}
\end{figure}

\begin{figure}[H]
\centering
\begin{subfigure}{0.495\textwidth}
    \centering
    \includegraphics[width=\linewidth]{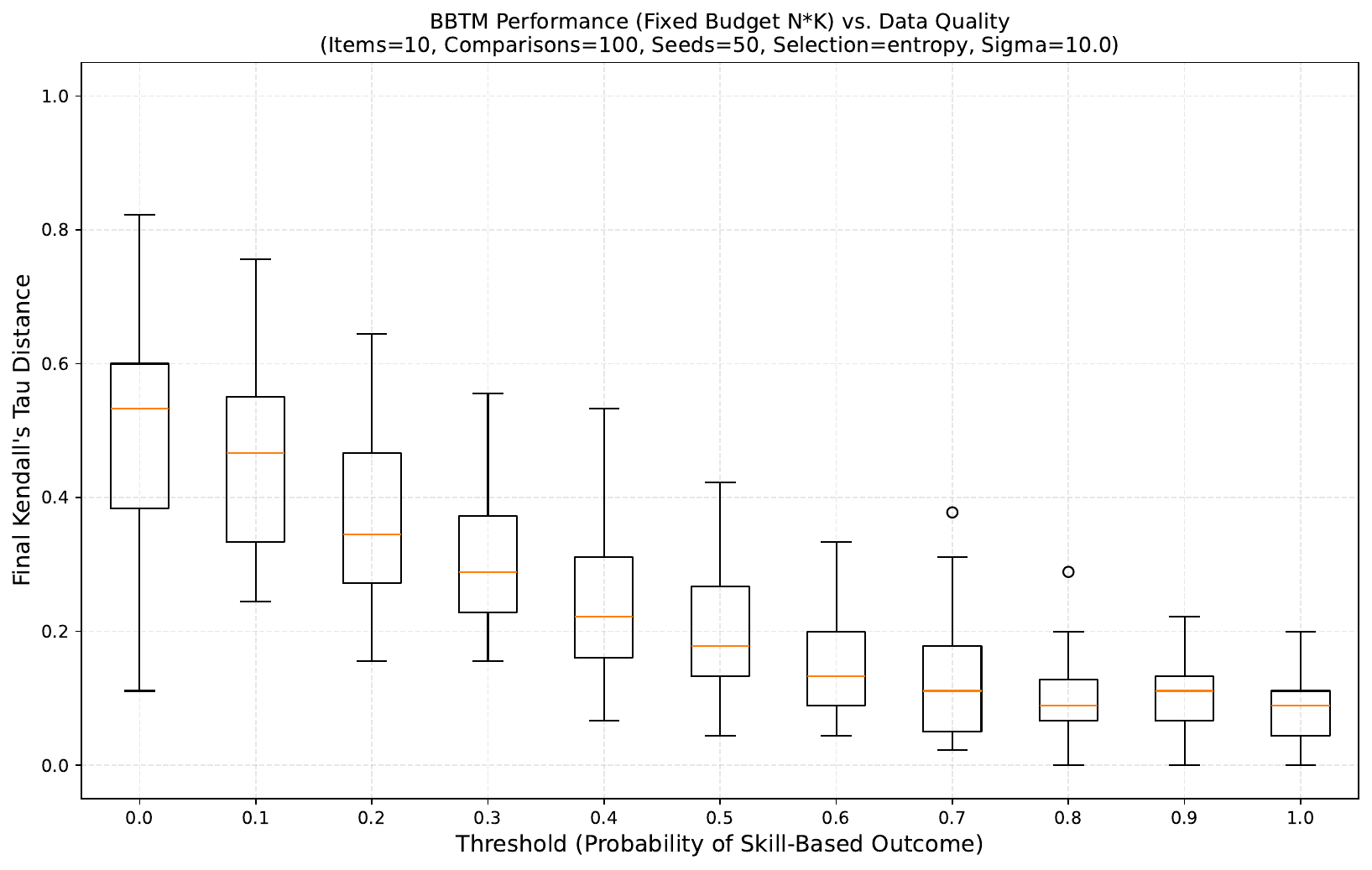}
    \caption{entropy}
    \label{fig:entropy pair selection}
\end{subfigure}
\hfill
\begin{subfigure}{0.495\textwidth}
    \centering
    \includegraphics[width=\linewidth]{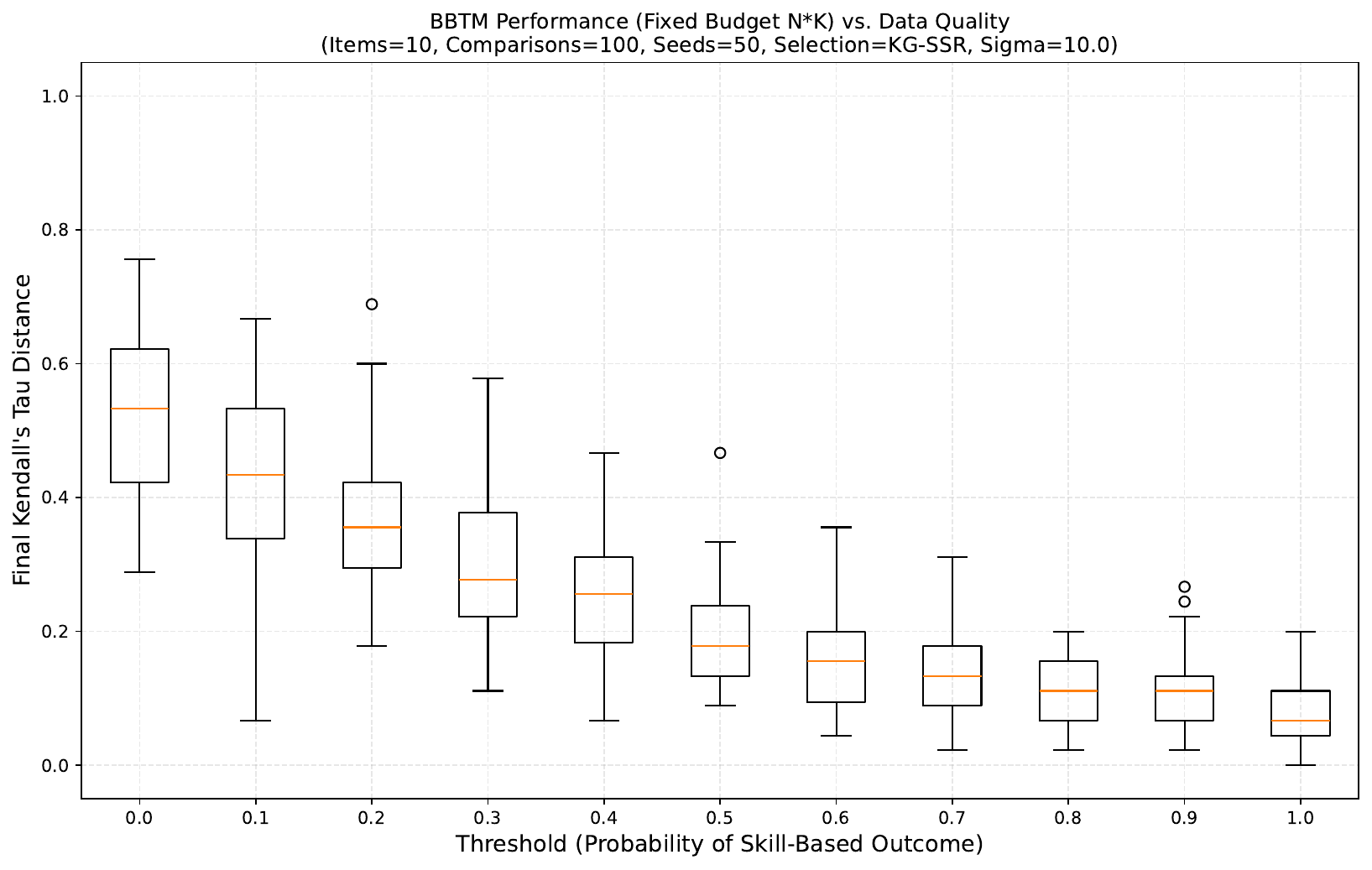}
    \caption{KG-SSR}
    \label{fig:KG-SSR pair selection}
\end{subfigure}

\caption{Robustness plots for $N=10$, $K=10$, $\sigma=10.0$}
\label{fig:psr-entropy}
\end{figure}

\section{Additional plots for empirical validation of the convergence theorem for $\sigma = 0.0001, 1.25, 2.50, 10.0$}
\label{app:convergence-results}
This appendix presents coverage and convergence plots remaining noise ($\sigma$) levels.
\begin{figure}[H]
    \centering
    \includegraphics[width=\textwidth]{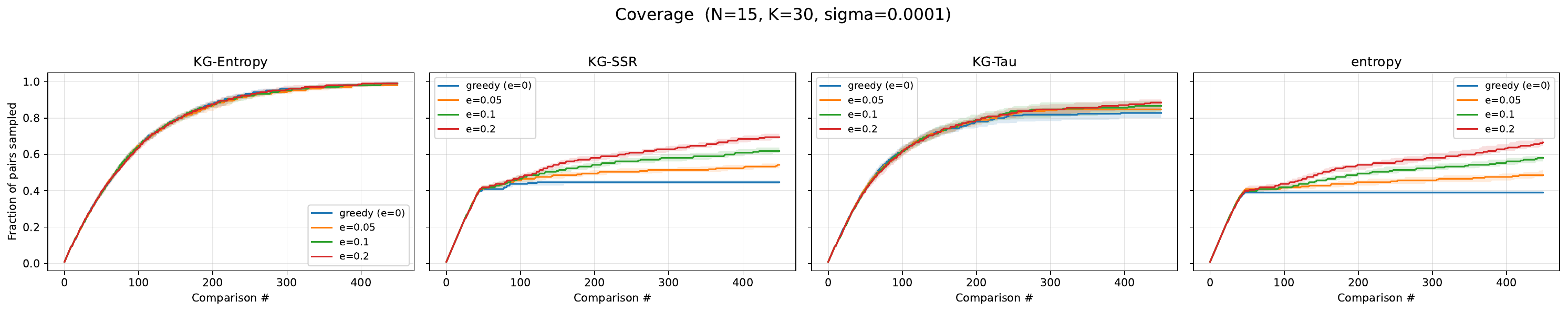}
    \caption{Pair coverage for $\sigma = 0.0001$.}
    \label{fig:app-coverage-sigma-0001}
\end{figure}

\begin{figure}[H]
    \centering
    \includegraphics[width=\textwidth]{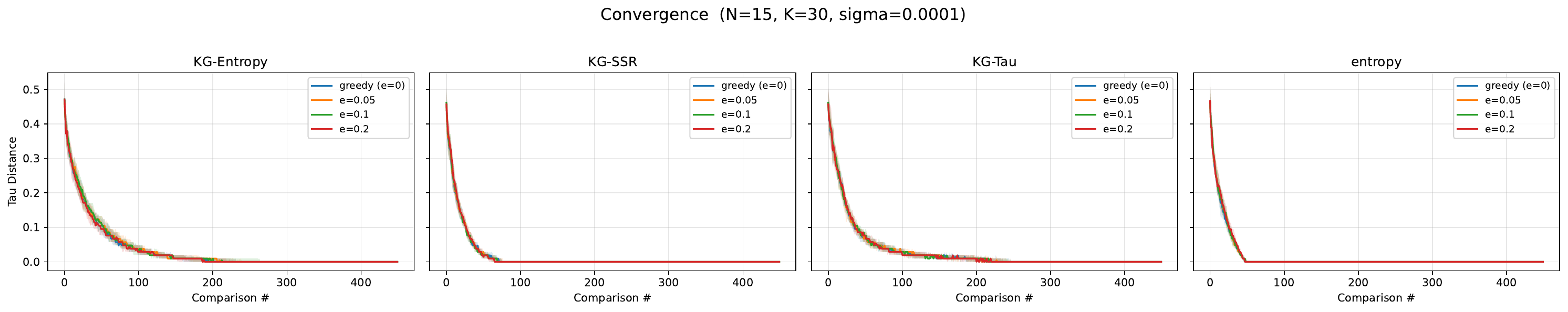}
    \caption{Convergence for $\sigma = 0.0001$.}
    \label{fig:app-convergence-sigma-0001}
\end{figure}

\begin{figure}[H]
    \centering
    \includegraphics[width=\textwidth]{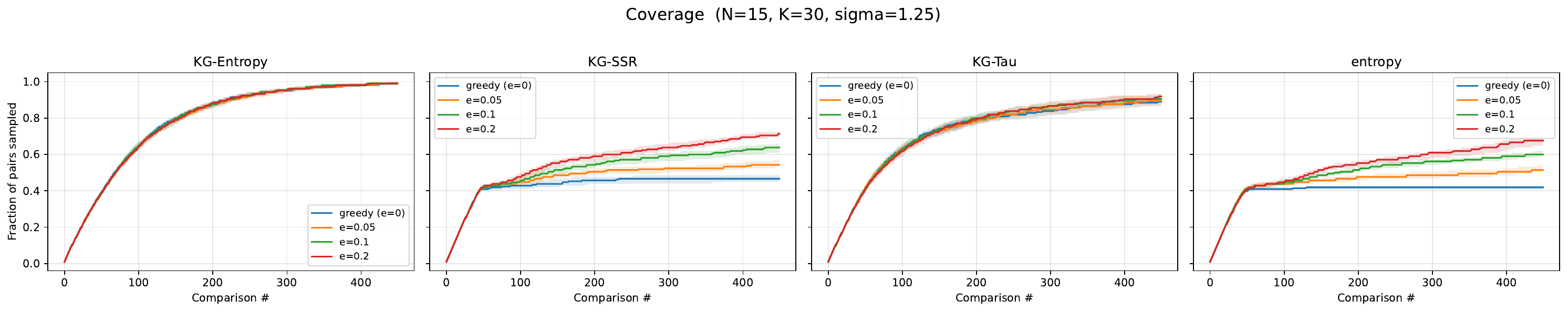}
    \caption{Pair coverage for $\sigma = 1.25$.}
    \label{fig:app-coverage-sigma-125}
\end{figure}

\begin{figure}[H]
    \centering
    \includegraphics[width=\textwidth]{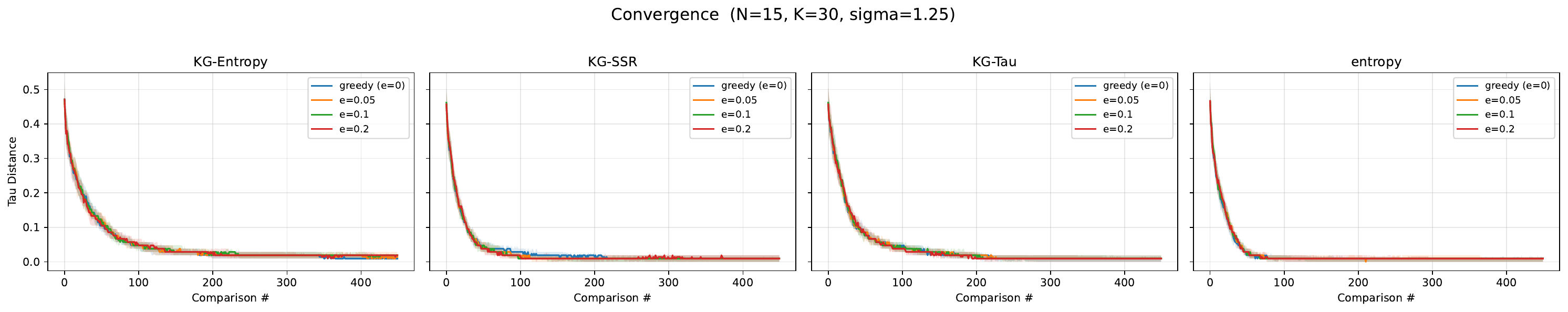}
    \caption{Convergence for $\sigma = 1.25$.}
    \label{fig:app-convergence-sigma-125}
\end{figure}

\begin{figure}[H]
    \centering
    \includegraphics[width=\textwidth]{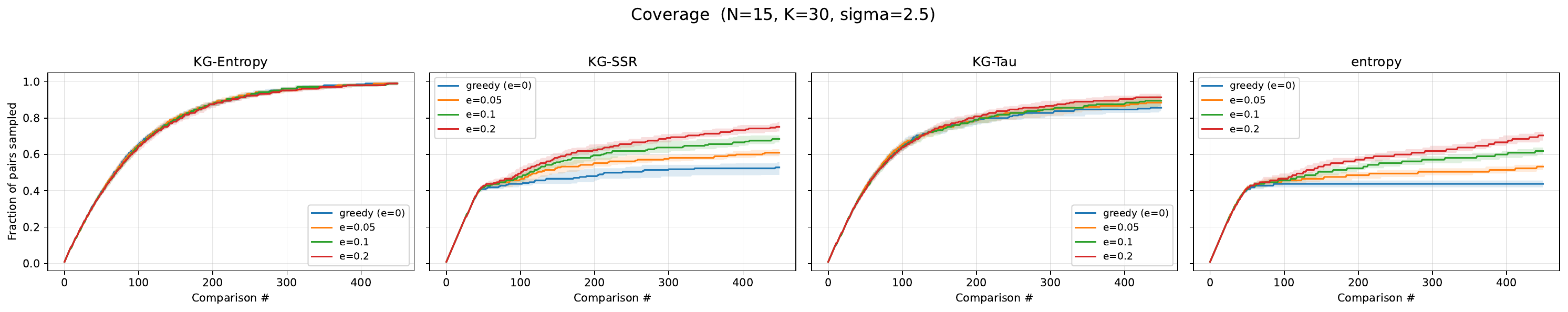}
    \caption{Pair coverage for $\sigma = 2.5$.}
    \label{fig:app-coverage-sigma-25}
\end{figure}

\begin{figure}[H]
    \centering
    \includegraphics[width=\textwidth]{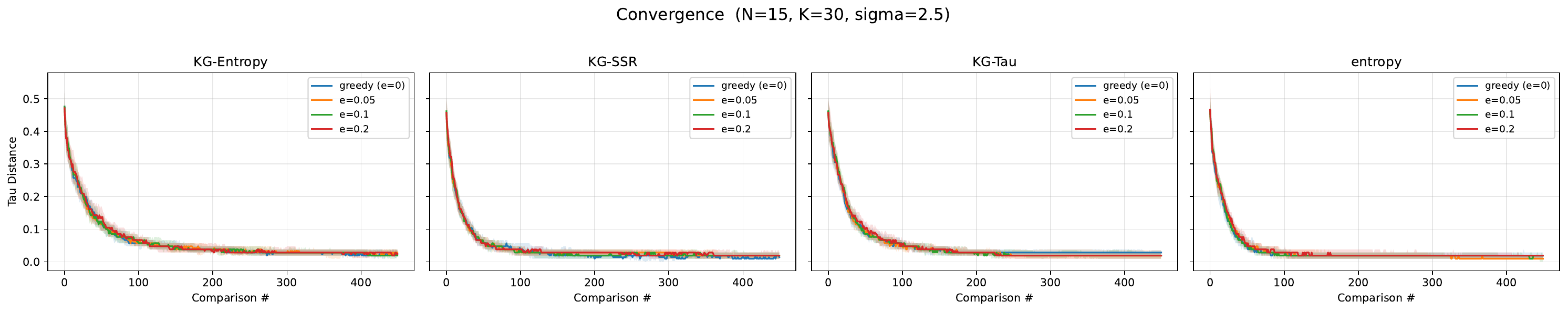}
    \caption{Convergence for $\sigma = 2.5$.}
    \label{fig:app-convergence-sigma-25}
\end{figure}

\begin{figure}[H]
    \centering
    \includegraphics[width=\textwidth]{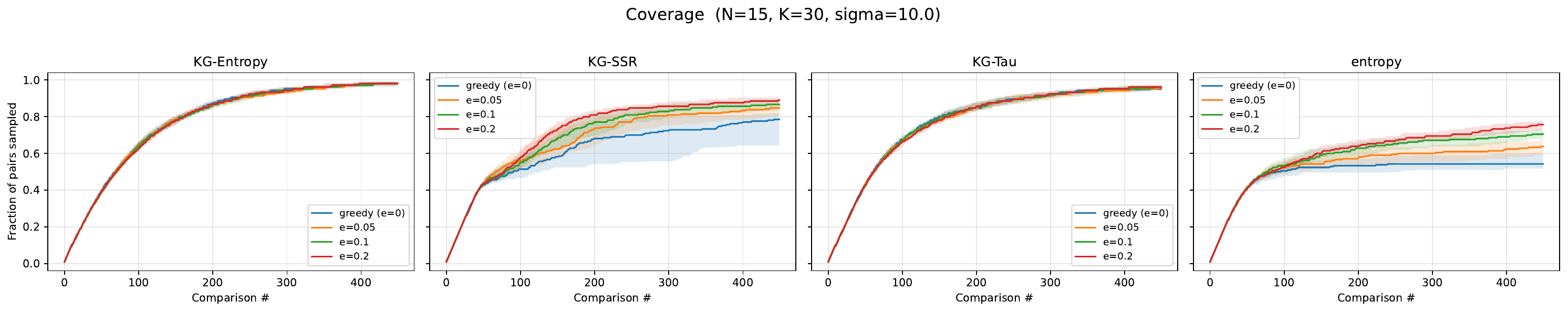}
    \caption{Pair coverage for $\sigma = 10.0$.}
    \label{fig:app-coverage-sigma-10}
\end{figure}

\begin{figure}[H]
    \centering
    \includegraphics[width=\textwidth]{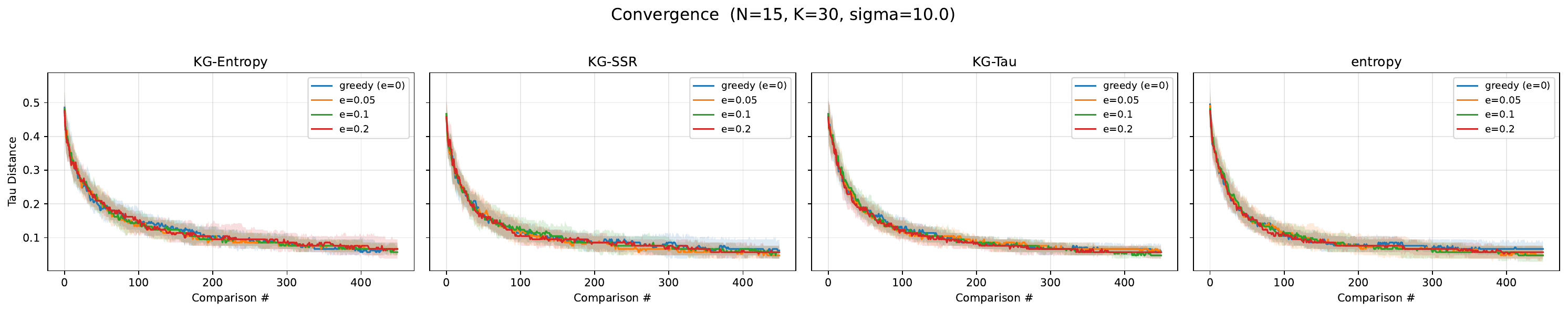}
    \caption{Convergence for $\sigma = 10.0$.}
    \label{fig:app-convergence-sigma-10}
\end{figure}

\end{appendices}

\end{document}